\documentclass[preprint,5p,times,twocolumn,authoryear]{elsarticle}

\usepackage{amssymb}

\usepackage{amsmath} 
\usepackage{booktabs} 
\usepackage{makecell} 
\usepackage{multirow} 
\usepackage{subcaption} 
\usepackage{mwe} 
\usepackage{bm}
\usepackage{xcolor}
\usepackage{array}

\newcolumntype{L}[1]{>{\raggedright\let\newline\\\arraybackslash\hspace{0pt}}m{#1}}
\newcolumntype{C}[1]{>{\centering\let\newline\\\arraybackslash\hspace{0pt}}m{#1}}
\newcolumntype{R}[1]{>{\raggedleft\let\newline\\\arraybackslash\hspace{0pt}}m{#1}}

\usepackage{pifont}
\usepackage[hidelinks]{hyperref}

\newcommand{\xmark}{\ding{55}}%
\newcommand{\cmark}{\ding{51}}%

\journal{Neural Networks}

\begin{document}

\begin{frontmatter}




\title{OnDev-LCT: On-Device Lightweight Convolutional Transformers\\towards Federated Learning}



\author[label1]{Chu Myaet Thwal}
\ead{chumyaet@khu.ac.kr}
\author[label2]{Minh N. H. Nguyen}
\ead{nhnminh@vku.udn.vn}
\author[label1]{Ye Lin Tun}
\ead{yelintun@khu.ac.kr}
\author[label1]{Seong Tae Kim}
\ead{st.kim@khu.ac.kr}
\author[label3]{My T. Thai}
\ead{mythai@cise.ufl.edu}
\author[label1]{Choong Seon Hong\corref{cor1}}
\ead{cshong@khu.ac.kr}

\affiliation[label1]{organization={Department of Computer Science and Engineering, Kyung Hee University},
            city={Yongin-si},
            state={Gyeonggi-do 17104},
            country={South Korea}}
            
\affiliation[label2]{organization={Vietnam - Korea University of Information and Communication Technology},
            city={Danang},
            country={Vietnam}} 

\affiliation[label3]{organization={Department of Computer and Information Science and Engineering, University of Florida},
            city={Gainesville},
            state={Florida 32611},
            country={USA}}

\cortext[cor1]{Corresponding author}


\begin{abstract}\label{sec:abstract}
Federated learning (FL) has emerged as a promising approach to collaboratively train machine learning models across multiple edge devices while preserving privacy.
The success of FL hinges on the efficiency of participating models and their ability to handle the unique challenges of distributed learning.
While several variants of Vision Transformer (ViT) have shown great potential as alternatives to modern convolutional neural networks (CNNs) for centralized training, the unprecedented size and higher computational demands hinder their deployment on resource-constrained edge devices, challenging their widespread application in FL.
Since client devices in FL typically have limited computing resources and communication bandwidth, models intended for such devices must strike a balance between model size, computational efficiency, and the ability to adapt to the diverse and non-IID data distributions encountered in FL.
To address these challenges, we propose \emph{OnDev-LCT}: \emph{L}ightweight \emph{C}onvolutional \emph{T}ransformers for \emph{On}-\emph{Dev}ice vision tasks with limited training data and resources.
Our models incorporate image-specific inductive biases through the LCT tokenizer by leveraging efficient depthwise separable convolutions in residual linear bottleneck blocks to extract local features, while the multi-head self-attention (MHSA) mechanism in the LCT encoder implicitly facilitates capturing global representations of images.
Extensive experiments on benchmark image datasets indicate that our models outperform existing lightweight vision models while having fewer parameters and lower computational demands, making them suitable for FL scenarios with data heterogeneity and communication bottlenecks.
\end{abstract}





\end{frontmatter}




\section{Introduction}\label{sec:introduction}

Recent trends in deep learning enable artificial intelligence to make significant strides, even surpassing humans in many tasks.
When it comes to image understanding, convolutional neural networks (CNNs) have become the de facto among numerous types of deep neural architectures.
Along with the tremendous size of publicly available image datasets, i.e., ImageNet~\citep{deng2009imagenet} and Microsoft COCO~\citep{lin2014coco}, deep CNNs like ResNet~\citep{he2016resnet} and its variants have been dominating the field of computer vision for decades.
Evidently, lightweight CNNs, such as MobileNets~\citep{howard2017mobilenetv1, sandler2018mobilenetv2, howard2019mobilenetv3} and their extensions~\citep{han2020ghostnet, li2021micronet}, have achieved state-of-the-art performance for many on-device vision tasks since image-specific inductive biases~\citep{lenc2015understanding, mitchell2017spatial} allow them to learn visual representations and recognize complex patterns with fewer parameters.
CNNs, however, are spatially local~\citep{zuo2015convolutional}, making them hard to capture global representations, i.e., contextual dependencies between different image regions.

\begin{figure}[t]
	\centering
	\includegraphics[width=0.85\columnwidth]{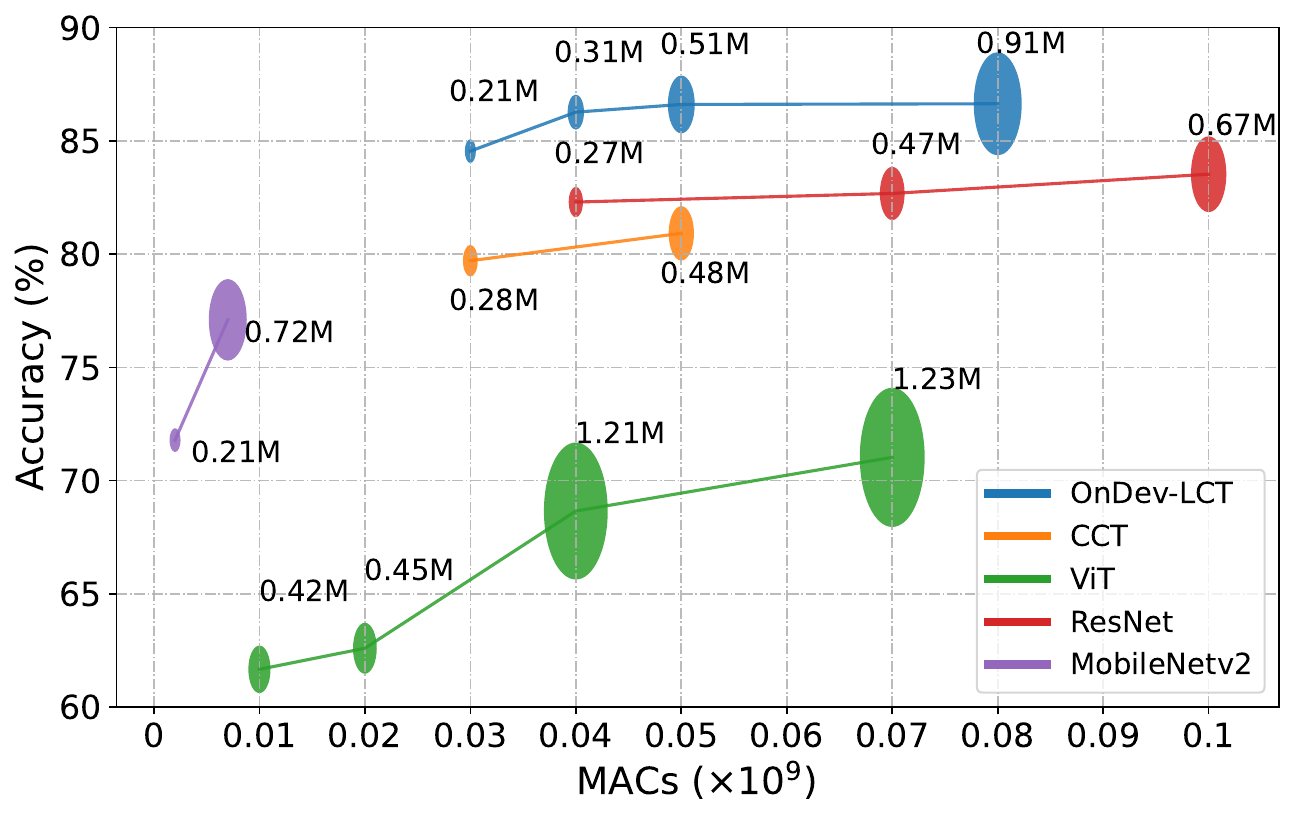}
	\caption{\textbf{Performance comparison on CIFAR-10 dataset in the centralized scenario.} The comparison is performed among small model variants, i.e., \#Params~$<$~1.3M, while constraining the computational budget within $0.1 \times 10^9$~MACs. Each bubble's area is proportional to the model size.}
	\label{fig:comparison}
\end{figure}

Parallel to the widespread use of CNNs in computer vision, transformer architectures~\citep{vaswani2017attention}, which exploit the multi-head self-attention mechanism, have emerged to become an integral part of natural language processing (NLP).
In response to the dominant success of transformers in large-scale language modeling, attention-based architectures have recently piqued a growing interest in computer vision.
Attention-embedded CNN models~\citep{wang2017residual, hu2018senet, bello2019aacn}, transformer-based networks~\citep{parmar2018image, dosovitskiy2020image, touvron2021training}, and most recently, hybrid architectures of convolutional transformers~\citep{wu2021cvt, hassani2021escaping, mehta2021mobilevit, graham2021levit, chen2022mobile} have shown their potential in a wide range of vision tasks.
Vision Transformer (ViT)~\citep{dosovitskiy2020image}, adapting the encoder design of NLP transformer~\citep{vaswani2017attention} with minimum modifications, shows its superior performance over cutting-edge CNNs in several benchmark tasks.
Hence, ViT becomes more prevalent in computer vision for its ability to learn global representations through attention mechanisms.

Conversely, the general trend towards improving the performance of transformers is to increase the number of parameters, which demands a vast amount of computational resources.
Thus, the unprecedented size of ViT hinders its applications on resource-constrained devices.
Moreover, ViT requires pre-training on large-scale image datasets, i.e., ImageNet-21K or JFT-300M, to learn generalizable visual representations since it lacks some inductive biases inherent in CNNs~\citep{dosovitskiy2020image}, such as translation equivariance and locality.
Recent studies focus on combining the strengths of CNNs with transformers and have become acknowledged for their high-performing hybrid architectures of convolutional transformers~\citep{wu2021cvt, hassani2021escaping, mehta2021mobilevit, graham2021levit, chen2022mobile}.
However, these hybrid models still have a huge number of parameters or require extensive data augmentation techniques, demanding higher computational resources for training to perform at the same level as CNNs in low-data regimes.

On the other hand, for many domains including science and medical research~\citep{varoquaux2022machine, ghassemi2020review}, it is costly to get a large quantity of annotated data with the size of ImageNet~\citep{deng2009imagenet} for training.
Due to the privacy-sensitive nature of user data in such domains, collecting and centralizing the training data also turn out to be challenging.
Federated learning (FL) is a distributed model training strategy that guarantees user privacy without the need for data collection.
FL enables multiple client devices to collaboratively train their local models without exposing the sensitive data of each client.
The vanilla FL algorithm called federated averaging (FedAvg)~\citep{mcmahan2017fedavg} learns a shared global model by iteratively averaging the local model updates of participating clients.
A successful FL implementation can yield promising results for on-device applications, especially in certain domains where data privacy and confidentiality are crucial.
Nonetheless, FL introduces several core challenges, including communication bottlenecks and data heterogeneity~\citep{li2020flchallenges}, which need to be considered before it can be applied in real-world distributed systems.
Moreover, since client devices can only offer a limited amount of computing resources, models intended for those devices should be lightweight and capable of on-device applications.
As a result, it demands an open research question: \emph{``Whether a powerful lightweight vision transformer can be efficiently designed for resource-constrained devices?"}.

In light of this demand, we propose \emph{OnDev-LCT}: \emph{On}-\emph{Dev}ice \emph{L}ightweight \emph{C}onvolutional \emph{T}ransformers for vision tasks, inspired by the idea of employing an early convolutional stem in transformers.
We introduce image-specific inductive priors to our models by incorporating a convolutional tokenizer before the transformer encoder.
Specifically, we leverage efficient \emph{depthwise separable convolutions} in our LCT tokenizer for extracting local features from images.
The \emph{multi-head self-attention (MHSA) mechanism} in our LCT encoder implicitly enables our models to learn global representations of images.
Hence, benefiting from both convolutions and attention mechanisms, we design our OnDev-LCTs for on-device vision tasks with limited training data and resources.
We analyze the performance of our models on various image classification benchmarks in comparison with popular lightweight vision models in the centralized scenario.
We further explore our OnDev-LCT models in FL scenarios to empower client devices for vision tasks while coping with data heterogeneity, resource constraints, and user privacy~\citep{li2020flchallenges}.
Since data augmentations during local training can incur additional computation overhead for resource-constrained FL devices, we compare our models with the baselines in FL scenarios without applying data augmentation techniques.
Our empirical analysis indicates that OnDev-LCTs significantly outperform the other baseline models in both centralized and FL scenarios under various data heterogeneity settings.
Additionally, we provide detailed discussions based on our observations from various performance analyses to demonstrate the efficiency and effectiveness of our OnDev-LCTs.
Figure~\ref{fig:comparison} depicts the image classification performance on the CIFAR-10~\citep{krizhevsky2009cifar} dataset for different model families with variants of comparable sizes, i.e., \#Params~$<$~1.3M (millions) and multiply-accumulate or MACs~$<$~0.1G (billions), in the centralized scenario.
Compared to the other baselines, we can observe that our OnDev-LCTs can achieve better accuracy with fewer parameters and lower computational demands.


\section{Related Works}\label{sec:related_works}

In this section, we highlight the evolution of prior research efforts in computer vision.
Then, we give an overview of how FL can be applied for on-device vision applications.

\paragraph{Convolutional neural networks}\label{para:cnns}
Since AlexNet's~\citep{krizhevsky2012alexnet} record-breaking achievement on ImageNet~\citep{deng2009imagenet}, CNN architectures have gained a great deal of traction in computer vision.
As the general trend towards better accuracy is to develop deeper and wider networks~\citep{li2021survey}, deep CNNs~\citep{he2016resnet, simonyan2014vggnet, szegedy2015googlenet} that can extract complex features from images have become popular for vision applications.
The residual attention networks proposed by~\citet{wang2017residual} build attention-aware feature maps to provide different representations of focused image patches as they embed the power of attention mechanisms in CNNs for better classification.
However, these improvements in accuracy do not always result in more efficient networks in terms of computation, speed, and size~\citep{khan2020survey}.
Real-world applications, such as medical imaging~\citep{varoquaux2022machine}, self-driving cars~\citep{parekh2022review}, robotics~\citep{roy2021machine, neuman2022tiny}, and augmented reality (AR)~\citep{mutis2020challenges} require edge devices to run vision tasks promptly under limited resources.
In response to these challenges, several lightweight CNNs~\citep{howard2017mobilenetv1, sandler2018mobilenetv2, howard2019mobilenetv3, han2020ghostnet, li2021micronet, hu2018senet, iandola2016squeezenet, park2020simple} have been developed as their flexible and simple-to-train nature may easily replace deep CNN backbones while minimizing the network size and improving latency.
Yet, CNNs have one major drawback: as they are spatially local, it makes them hard to extract complex global interactions among the spatial features of images on a broader range~\citep{zuo2015convolutional, lecun2015deep}.

\paragraph{Transformers in vision}\label{para:vit}
\citet{dosovitskiy2020image}~introduced Vision Transformer (ViT) as the first notable example of using a pure transformer backbone for computer vision tasks.
However, lacking some inductive priors makes it difficult for ViT to enhance its generalization ability for out-of-distribution data samples when training data is insufficient.
Hence, transformers are data-hungry, demanding extremely large pre-training datasets, such as ImageNet-21K or JFT-300M, to compete with their convolutional counterparts~\citep{khan2021transformers}.
This has led to an explosion in model and dataset sizes, limiting the use of ViT among practitioners in low-data regimes.
In the context of lightweight and efficient vision transformers, parallel lines of research have emerged, exploring different strategies to accomplish high-performance on-device vision tasks.
Researchers have delved into techniques such as knowledge distillation~\citep{touvron2021training}, quantization~\citep{li2022qvit, lin2021fqvit}, pruning~\citep{zhu2021vtp, fang2023depgraph}, and architectural modifications~\citep{mehta2021mobilevit, graham2021levit} to enhance the efficiency of transformers on vision tasks.
In an effort to improve upon ViT, the Data-efficient image Transformer (DeiT)~\citep{touvron2021training} introduced a transformer-specific distillation strategy that does not require a large quantity of training data.
For a given parameter constraint, DeiT still requires huge computational resources, with its performance slightly below that of  EfficientNets~\citep{tan2019efficientnet} with equivalent sizes.
In addition, extensive data augmentation techniques and powerful teacher models are prerequisites for the best performance of DeiT models.
Nonetheless, models for real-world vision applications should be \emph{lightweight and efficient} so that they can be deftly run on resource-constrained devices~\citep{menghani2023efficient}.

As one of the most powerful model compression approaches, quantization has gained considerable attention due to its potential in reducing the computation cost and memory overhead while striving to maintain the performance of vision transformers.
Quantization involves converting the full-precision or floating-point values (i.e., 32 bits) of weights and activations to integers with lower bit-width, thus accelerating the processing of vision transformers in resource-constrained settings.
To this point, several works have been studied for adapting vision transformers to two distinct families of quantization methods: quantization-aware training (QAT)~\citep{li2022qvit, liu2023ofqvit} and post-training quantization (PTQ)~\citep{liu2021vtptq, lin2021fqvit, yuan2021ptq4vit, ding2022apqvit}, depending on the trade-off between achieving better accuracy and simpler deployment.
With an intuitive idea of compressing the model without significant performance sacrifice, pruning~\citep{blalock2020state, FrankleD0C21} has also proved to be highly effective and practical for accelerating vision transformers to varying degrees.
Researchers have explored various pruning techniques aimed at removing specific components of vision transformers, such as attention heads or individual weights, to create more streamlined variants suitable for resource-constrained devices.
For instance, unstructured pruning~\citep{dong2017learning, lecun1989obd, hassibi1993optimal, sanh2020movement} targets individual weights, providing more flexibility in model compression, but it often requires special hardware accelerators or software for model acceleration~\citep{han2015deep_compression}.
Conversely, structured pruning~\citep{zhu2021vtp, yu2023x, yu2022width, yang2023global, fang2023depgraph} has gained great interest due to its hardware-friendly nature and ability to reduce inference overhead by removing entire components of vision transformers, such as multi-head self-attention layers, token embeddings, or neurons.
Nonetheless, the choice of the most suitable pruning technique for a particular task depends on several factors including the target model size, the desired performance level, and the available computational resources, in order to strike a balance between reducing the model size and retaining essential representations and prediction capabilities.

\paragraph{Convolutional transformers}\label{para:conv_vit}
In the context of architectural modifications, the Convolutional vision Transformer (CvT)~\citep{wu2021cvt} introduced convolutions into ViT for improving its performance and efficiency on vision tasks.
A convolutional token embedding is added prior to each hierarchical stage of CvT transformers, along with a convolutional projection layer, which replaces the position-wise projection of multi-head self-attention in the transformer encoders.
\citet{hassani2021escaping}~dispelled the myth of ``data-hungry vision transformers" by presenting ViT-Lite with smaller patch sizing for patch-based image tokenization of the original ViT.
The same authors introduced a sequential pooling method (SeqPool) in Compact Vision Transformer (CVT)~\citep{hassani2021escaping} to eliminate the need for a classification token from vision transformers.
Coupling the strengths of CNNs and ViT, Compact Convolutional Transformer (CCT)~\citep{hassani2021escaping} and MobileViT~\citep{mehta2021mobilevit} have been proposed as lightweight hybrid models to improve the flexibility of transformers in vision tasks.
CCT replaced the patch-based tokenization of ViT with convolutional tokenization that preserves the local spatial relationships between convolutional patches.
MobileViT introduced a lightweight ViT model by replacing the local processing in convolutions with transformers to learn global representations of images.
Thus, in addition to the aspect of ViT in capturing long-range feature dependencies provided by the attention mechanism, convolution and pooling operations of CNNs come up with the translation equivariance and locality, allowing these hybrid models to work better on smaller datasets.
To this end, these models desire high-level data augmentation techniques to justify their optimal performance in vision tasks.
Likewise, large computational resources are still essential for them to beat mobile-friendly CNN models.

\begin{figure}[t]
	\centering
	\includegraphics[width=0.85\columnwidth]{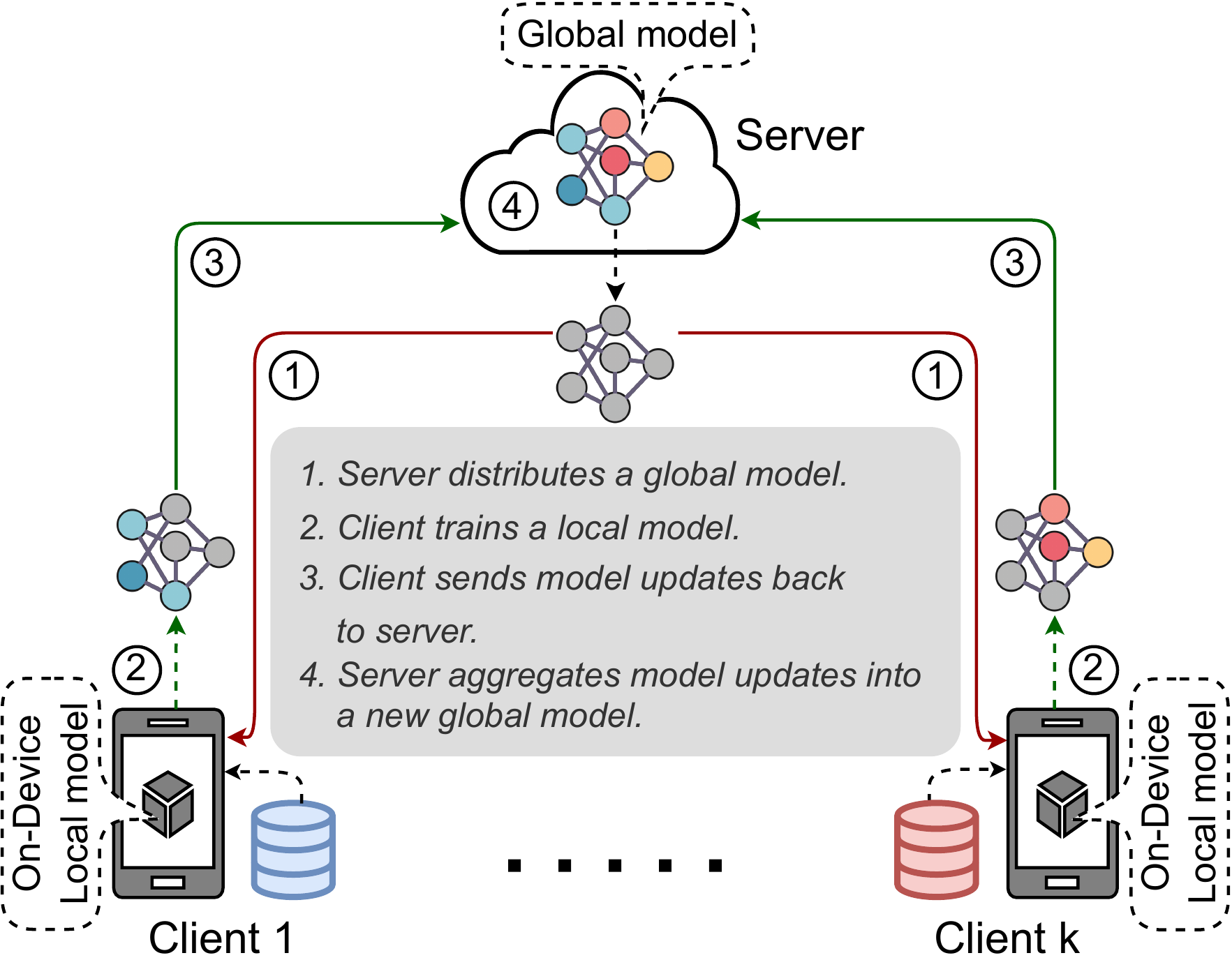}
	\caption{\textbf{The vanilla federated averaging (FedAvg) framework.} A typical four-step procedure is iterated until convergence.}
	\label{fig:fedavg}
\end{figure}

Recently,~\citet{graham2021levit} proposed a hybrid neural network named LeViT for faster inference of image classification by designing a multi-stage transformer in ConvNet's clothing.
Convolutional blocks of LeViT are built as pyramids for decreasing the resolution of feature maps with an increasing number of channels.
The authors replaced the ViT's positional embedding with a learned, per-head translation-invariant attention bias and redesigned attention-MLP blocks to improve the network capacity.
LeViT models are trained following the DeiT's distillation-driven training with two classification heads; thus, they require a powerful pre-trained teacher model, i.e., RegNetY-16GF, and strong computational resources, i.e., 32 GPUs, to perform 1000 training epochs in 3 to 5 days.
Mobile-Former proposed by~\citet{chen2022mobile} parallelizes MobileNet~\citep{howard2019mobilenetv3} and transformer with a bidirectional cross-attention bridge in between.
In contrast to the computationally efficient parallel design, Mobile-Former introduces additional parameters due to its parameter-heavy classification head.
The EdgeNeXt~\citep{maaz2022edgenext}, with its innovative split depthwise transpose attention (SDTA) encoder, demonstrates remarkable potential for mobile devices through its stage-wise design of hybrid architecture.
EdgeNeXt variants surpass previous hybrid models in both accuracy and model size while significantly reducing the computational overhead.

\citet{mehta2022mobilevitv2} proposed MobileViTv2, an improved version of the original MobileViT, to alleviate the main efficiency bottleneck caused by the multi-headed self-attention.
The authors introduced a separable self-attention method with linear complexity, making MobileViTv2 suitable for deployment on resource-constrained devices.
\citet{jeevan2022vix} adapted linear attention mechanisms to Vision Transformer (ViT)~\citep{dosovitskiy2020image}, resulting in the creation of Vision X-formers (ViXs), where X $\in$ \{Performer~(P), Linformer~(L), Nystr\"omformer~(N)\} \citep{choromanski2020performer, wang2020linformer, xiong2021nystromformer}.
This adaptation led to a remarkable reduction of up to seven times in GPU memory requirements.
Additionally, the authors introduced inductive priors by replacing the initial linear embedding layer with convolutional layers in ViXs, resulting in a substantial improvement in classification accuracy without increasing the model size.
Drawing our inspiration from the key concepts presented in those prior works, we aim to develop high-performing, lightweight convolutional transformers for vision applications on devices with limited resources in low-data regimes.

\begin{figure*}[t]
	\centering
	\includegraphics[width=0.85\linewidth]{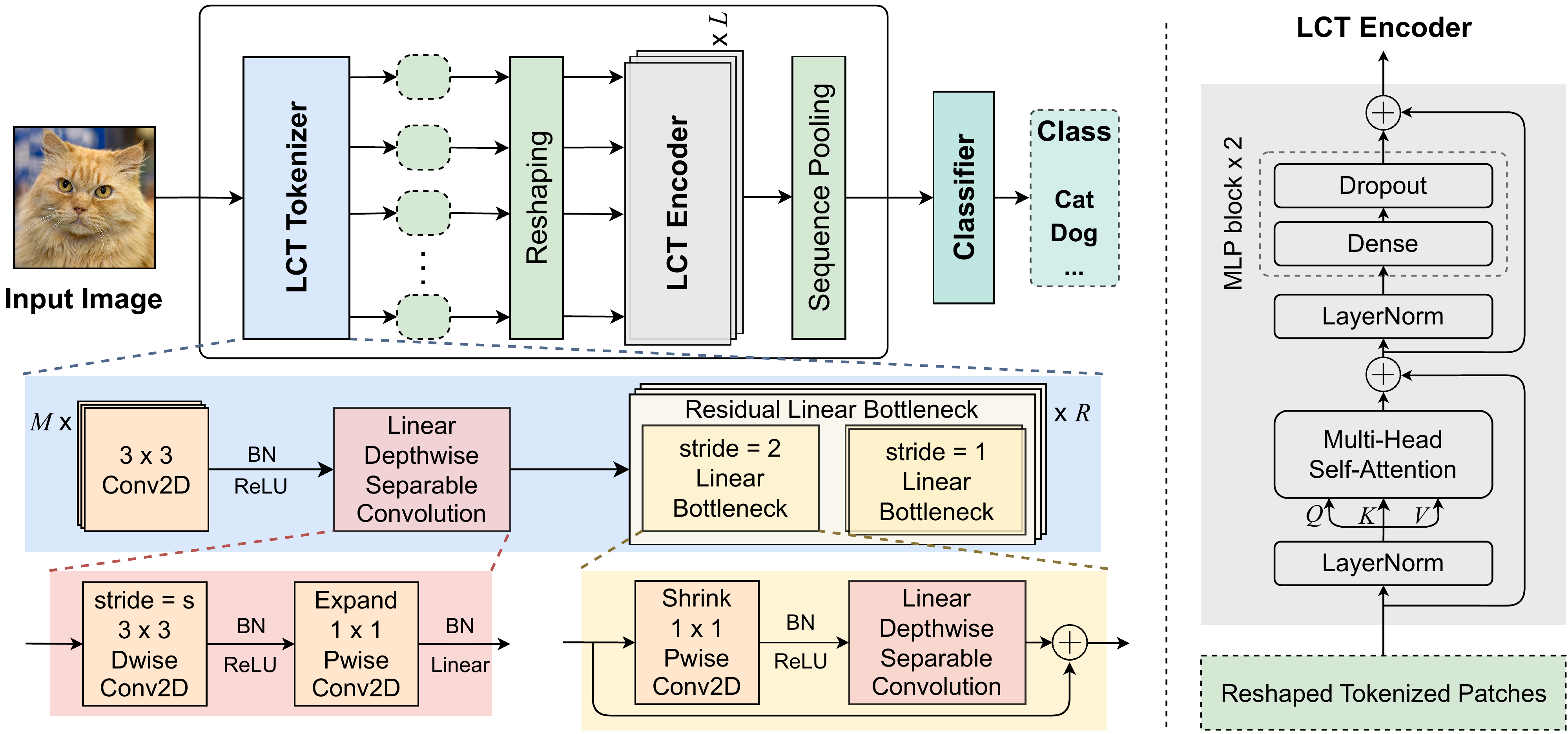}
	\caption{\textbf{An overview of On-Device Lightweight Convolutional Transformer (OnDev-LCT).} The middle row depicts the architecture of an LCT tokenizer, visualizing its core components in the bottom row. The structure of an LCT encoder is illustrated on the right.}
	\label{fig:system_model}
\end{figure*}

\paragraph{Federated learning}\label{para:fl}
Federated averaging (FedAvg)~\citep{mcmahan2017fedavg} is the earliest and most widely used FL baseline that executes an iterative model averaging process via a typical four-step procedure, as illustrated in Figure~\ref{fig:fedavg}.
However, FedAvg suffers from performance deterioration when dealing with distribution shifts in training data across multiple clients.
Various works have been proposed in an attempt to enhance the performance of FedAvg for tackling the data heterogeneity by focusing on either two aspects: \emph{improvements in local training}, i.e., Step 2 of Figure~\ref{fig:fedavg}~\citep{li2020fedprox, karimireddy2020scaffold, li2021moon} or \emph{improvements in server aggregation}, i.e., Step 4 of Figure~\ref{fig:fedavg}~\citep{wang2020fedma, hsu2019fedavgm, wang2020fednova}.
A recent study on ViT-FL~\citep{qu2022rethinking} presents a new perspective on utilizing ViT in federated vision models.
Their empirical analysis demonstrates that ``transformers are highly robust to distribution shifts in data".
However, the performance of ViT-FL heavily relies on the pre-training dataset, or ViT requires a huge amount of training data to outperform CNNs when trained from scratch.
Often, it is hard to readily obtain a well-pre-trained ViT model for various domain-specific tasks in a practical distributed system.
Moreover, the unprecedented size of ViT (e.g., $\approx6$M for ViT-Tiny or $\approx22$M for ViT-S~\citep{qu2022rethinking, touvron2022three}) may introduce hurdles to communication bottlenecks when exchanging the model updates for aggregation.
Hence, in this study, we aim to tackle those challenges by deploying our OnDev-LCTs in FL scenarios under various data heterogeneity settings.


\section{Preliminaries}\label{sec:preliminaries}

In this section, we describe the preliminaries and motivations behind the development of our OnDev-LCT models.

\paragraph{Patch-based tokenization}\label{para:vit_token}
In the standard ViT model~\citep{dosovitskiy2020image}, a $2D$ input image $x$ is split into multiple non-overlapping fixed-size patches:
\begin{equation*}
    x \in \mathbb{R}^{H \times W \times C} \Rightarrow x_p \in \mathbb{R}^{N \times (P^2C)},
\end{equation*}
where $C$ is the number of channels, $N = \frac{HW}{P^2}$ is the number of patches with $(H,W)$ and $(P,P)$ for the dimensions of original image and patches, respectively.
Image patches are flattened into a sequence of $1D$ vectors (aka tokens), which are then linearly projected into $D$-dimensional space: $[x_p^1\text{E}; x_p^2\text{E}; \dots; x_p^N\text{E}]$, where $\text{E} \in \mathbb{R}^{(P^2C) \times D}$ is a projection vector.
For the purpose of classification, an extra learnable $x_{class}$ token is prepended to the patch embeddings.
Since this ViT's patch-based tokenization step ignores \emph{spatial inductive biases}, information along the boundary regions of the input image is lost~\citep{khan2021transformers}.
Thus, a positional vector $\text{E}_{pos} \in \mathbb{R}^{(N+1) \times D}$ is encoded to the sequence, resulting in an input embedding $z_0$ to the transformer encoder as:
\begin{equation*}
    z_0 = [x_{class}; x_p^1\text{E}; x_p^2\text{E}; \dots; x_p^N\text{E}] + \text{E}_{pos}.
\end{equation*}
Although this positional encoding makes it easier to learn complex features through visual representations, ViT still requires a large scale of training data to generalize well.

\paragraph{Convolutional tokenization}\label{para:conv_token}
A convolutional tokenization block introduced by CCT~\citep{hassani2021escaping} consists of a single standard convolution with a ReLU~\citep{agarap2018relu} activation followed by a MaxPool operation.
An input image $x$ is tokenized into convolutional patches and reshaped into an input embedding $z_0$ to the transformer encoder as:
\begin{equation*}
    z_0 = \mathrm{MaxPool}(\mathrm{ReLU}(\mathrm{Conv2D}(x))).
\end{equation*}
This convolutional tokenization step makes CCT flexible enough to remove the positional encoding as well by injecting inductive biases into the model.
However, as the MaxPool operation reduces the spatial resolution of the input image, all the relevant information cannot be preserved well~\citep{sabour2017dynamic, xinyi2018capsule}.
Moreover, having multiple convolutional tokenization blocks may speed up the loss of all spatial information, which in turn hurts the model performance.

\paragraph{Transformer encoder}\label{para:vit_encoder}
In the original ViT~\citep{dosovitskiy2020image}, each encoder, i.e., $l = 1, 2, \dots, L$, is stacked on top of one another, sending its output $z_l$ to the subsequent encoder.
Each encoder contains two main components, such as a multi-head self-attention (MHSA) block and a 2-layer multi-layer perceptron (MLP) block, with a LayerNorm (LN) and residual connections in between.
The last encoder outputs a sequence $z_L$ of image representations.

\paragraph{Sequence pooling}\label{para:seq_pool}
An attention-based sequence pooling (SeqPool) in CVT~\citep{hassani2021escaping} maps the last encoder's output sequence to the classifier by pooling the relevant information across different parts of the sequence and eliminates the need for an extra learnable class token.


\section{OnDev-LCT Architecture}\label{sec:ondev-lct}

In this section, we present the architecture of our OnDev-LCT by explaining key components in detail.
It includes three main parts: \emph{LCT tokenizer}, \emph{LCT encoder}, and a \emph{classifier}.
An overview of our OnDev-LCT is illustrated in Figure~\ref{fig:system_model}.

\paragraph{LCT tokenizer}\label{para:lct_tokenizer}
Inspired by the idea of leveraging a convolutional stem for image tokenization, we introduce \emph{spatial inductive priors} into our models by replacing the entire patch-based tokenization of ViT~\citep{dosovitskiy2020image} with an LCT tokenizer.
Given an input image $x$, our LCT tokenizer encodes it into convolutional patches before forwarding it to the LCT encoder.
Our tokenizer starts with $M$ convolution operations:
\begin{equation}
    x^{m} = \mathrm{ReLU}(\mathrm{BN}(\mathrm{Conv2D}(x))), \quad m = 1, 2, \dots, M
\end{equation}
where Conv2D represents a standard $3 \times 3$ convolution followed by a BatchNorm (BN) with ReLU~\citep{agarap2018relu} activation.
After the final convolution operation, the output $x^{M}$ is sent to a block of \emph{Linear Depthwise Separable Convolution}:
\begin{equation}
    x'_0 = \mathrm{LinearDWS}(x^{M}),
\end{equation}
which is composed of a layer of $3 \times 3$ depthwise convolution followed by a layer of $1 \times 1$ pointwise convolution:
\begin{equation}
    \begin{split}
        x'_{0,dw} &= \mathrm{ReLU}(\mathrm{BN}(\mathrm{DepthwiseConv2D}(x^{M}))), \\
        x'_{0} &= \mathrm{Linear}(\mathrm{BN}(\mathrm{PointwiseConv2D}(x'_{0,dw}))).
    \end{split}
\end{equation}
DepthwiseConv2D extracts features through spatial filtering per input channel, while PointwiseConv2D is used to project channels output from the depthwise layer by a linear combination of those channels.
Each convolution operation is followed by BatchNorm with a ReLU for depthwise convolution and a Linear activation for pointwise convolution.
The output feature map is forwarded through $R$ \emph{Residual Linear Bottleneck} blocks:
\begin{equation}
        x'_r = \mathrm{LinearBottleneck}(x'_{r-1}) + x'_{r-1}, \quad r = 1, 2, \dots, R
\end{equation}
Each \emph{Bottleneck} block first shrinks the feature map via a $1 \times 1$ convolution.
DWS convolution is then applied to remap the output to the same dimension as the input feature map:
\begin{equation}
    \begin{split}
        x'_{r,shrink} &= \mathrm{ReLU}(\mathrm{BN}(\mathrm{PointwiseConv2D}(x'_{r-1}))), \\
        x'_{r} &= \mathrm{LinearDWS}(x'_{r,shrink}).
    \end{split}
\end{equation}

Here, instead of using several MaxPool operations, we implement DWS convolutions at the heart of our OnDev-LCT to maintain the spatial information of the input image while building up better representations.
The amount of computations can also be drastically reduced as DWS breaks the standard convolution operation, i.e., channel-wise and spatial-wise computations, into two separate layers.
Moreover, the \emph{Bottleneck} structure reduces the number of parameters and computation costs while preventing information loss of using non-linear functions, such as ReLU~\citep{agarap2018relu}).
The idea of residual blocks is to make our model as thin as possible by increasing the depth with fewer parameters while improving the ability to effectively train a lightweight vision transformer.
At the end of the LCT tokenization step, encoded patches are reshaped to be processed through $L$ blocks of LCT encoder:
\begin{equation}
    x'_R \in \mathbb{R}^{H_R \times W_R \times C_R} \Rightarrow z_0 \in \mathbb{R}^{HW \times D},
\end{equation}
where $C_R=D$ is the dimension of the input embedding $z_0$.

\begin{figure*}[t]
     \centering
     \begin{subfigure}[b]{0.163\textwidth}
         \centering
         \includegraphics[width=1.025\textwidth]{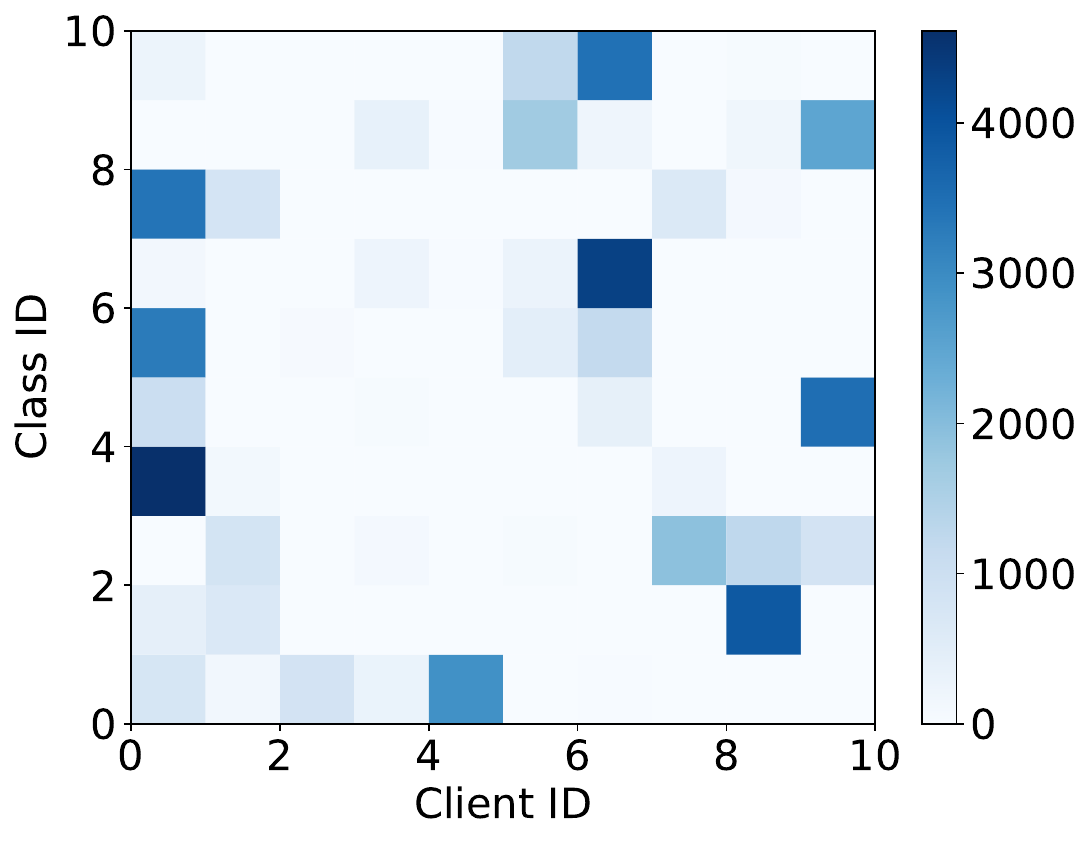}
         \caption{For 10 clients ($\beta=0.1$)}
         \label{fig:cifar-10_0.1_10}
     \end{subfigure}
     \hfill
     \begin{subfigure}[b]{0.163\textwidth}
         \centering
         \includegraphics[width=1.025\textwidth]{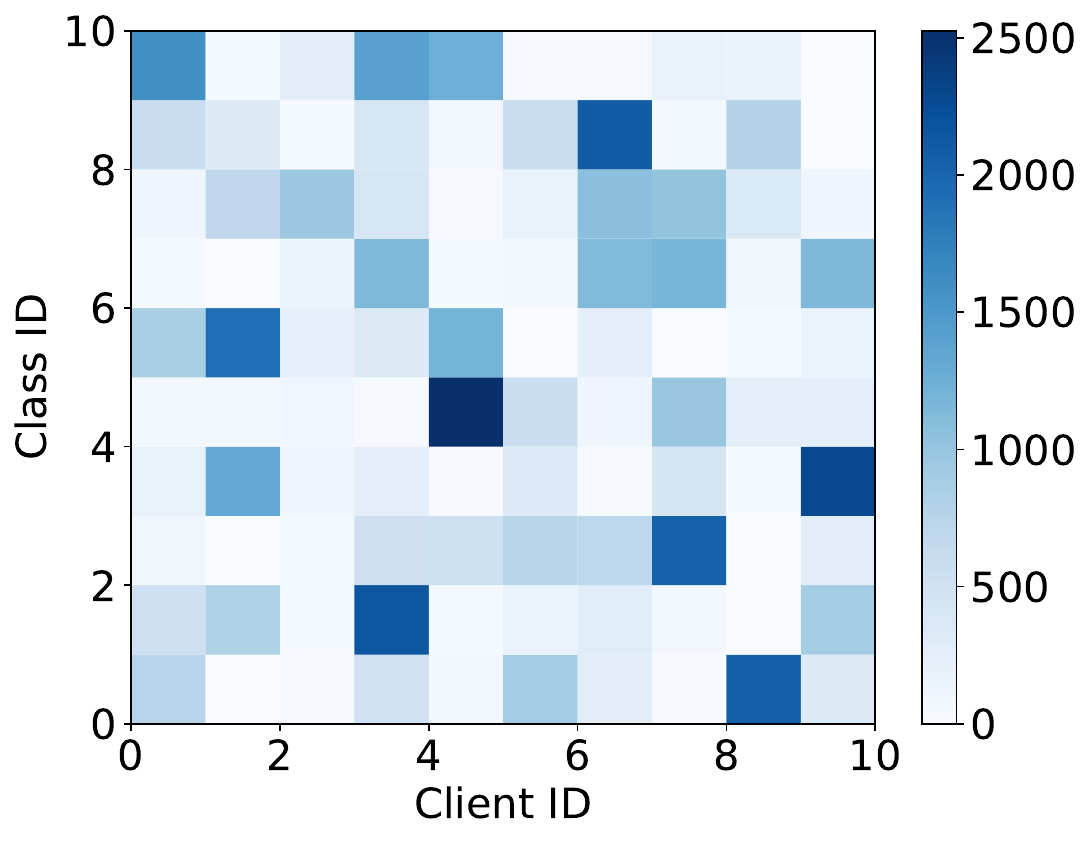}
         \caption{For 10 clients ($\beta=0.5$)}
         \label{fig:cifar-10_0.5_10}
     \end{subfigure}
     \hfill
     \begin{subfigure}[b]{0.163\textwidth}
         \centering
         \includegraphics[width=1.025\textwidth]{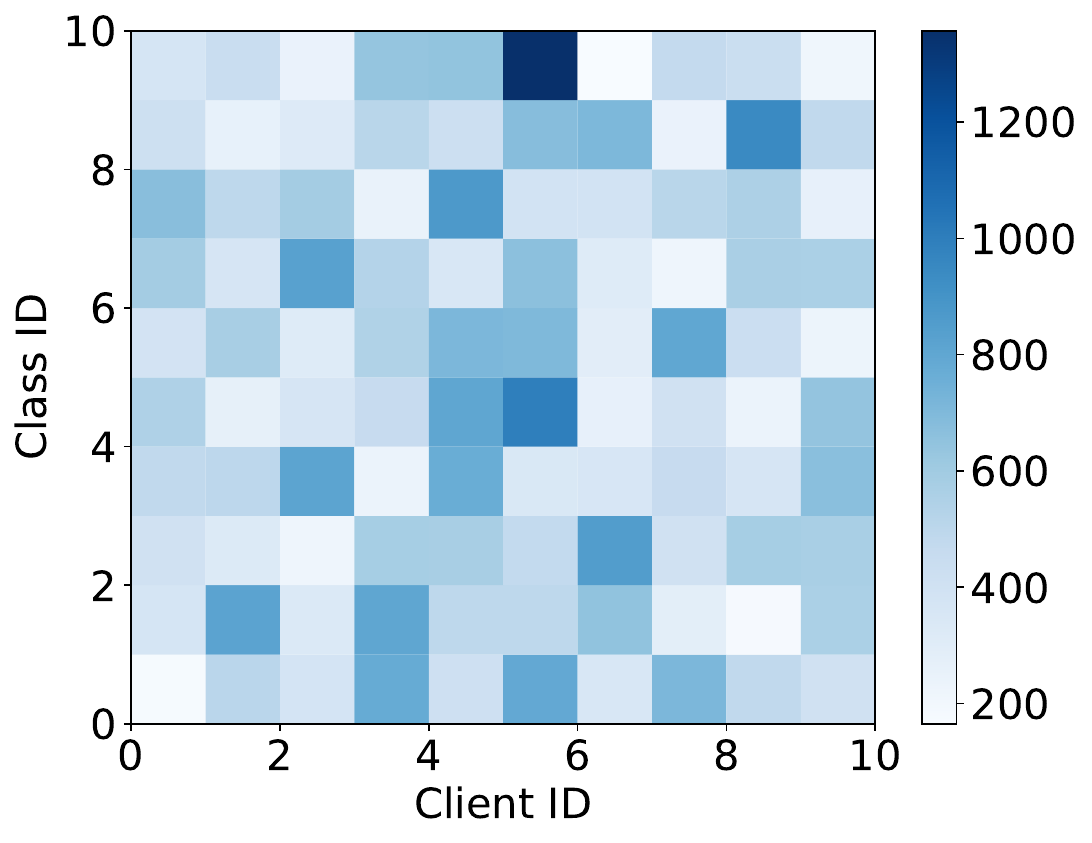}
         \caption{For 10 clients ($\beta=5$)}
         \label{fig:cifar-10_5_10}
     \end{subfigure}
     \begin{subfigure}[b]{0.163\textwidth}
         \centering
         \includegraphics[width=1.025\textwidth]{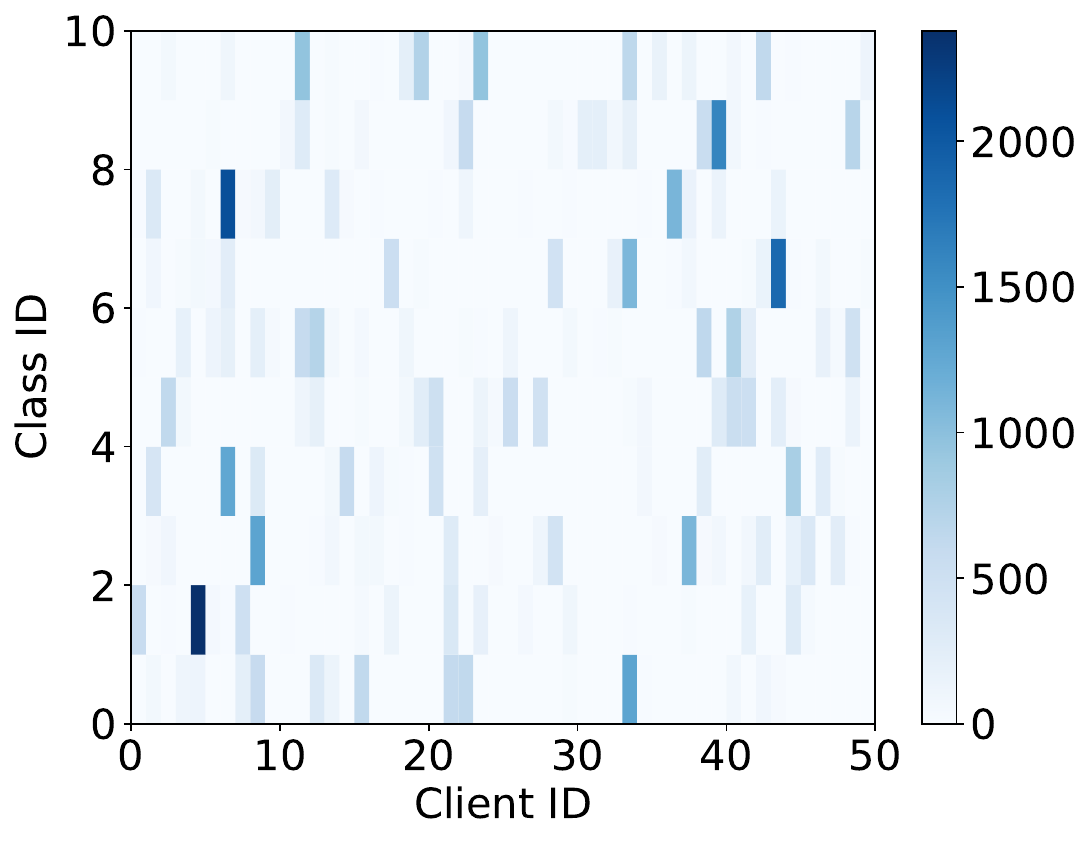}
         \caption{For 50 clients ($\beta=0.1$)}
         \label{fig:cifar-10_0.1_50}
     \end{subfigure}
     \hfill
     \begin{subfigure}[b]{0.163\textwidth}
         \centering
         \includegraphics[width=\textwidth]{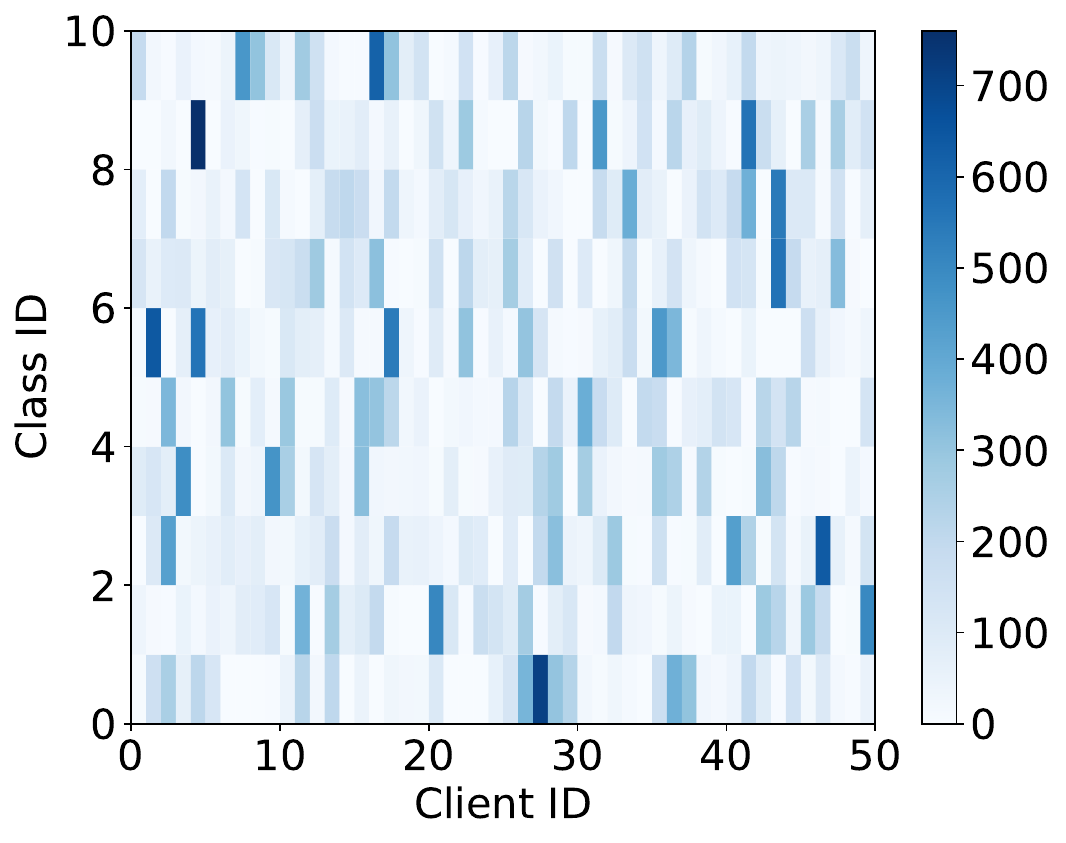}
         \caption{For 50 clients ($\beta=0.5$)}
         \label{fig:cifar-10_0.5_50}
     \end{subfigure}
     \hfill
     \begin{subfigure}[b]{0.163\textwidth}
         \centering
         \includegraphics[width=\textwidth]{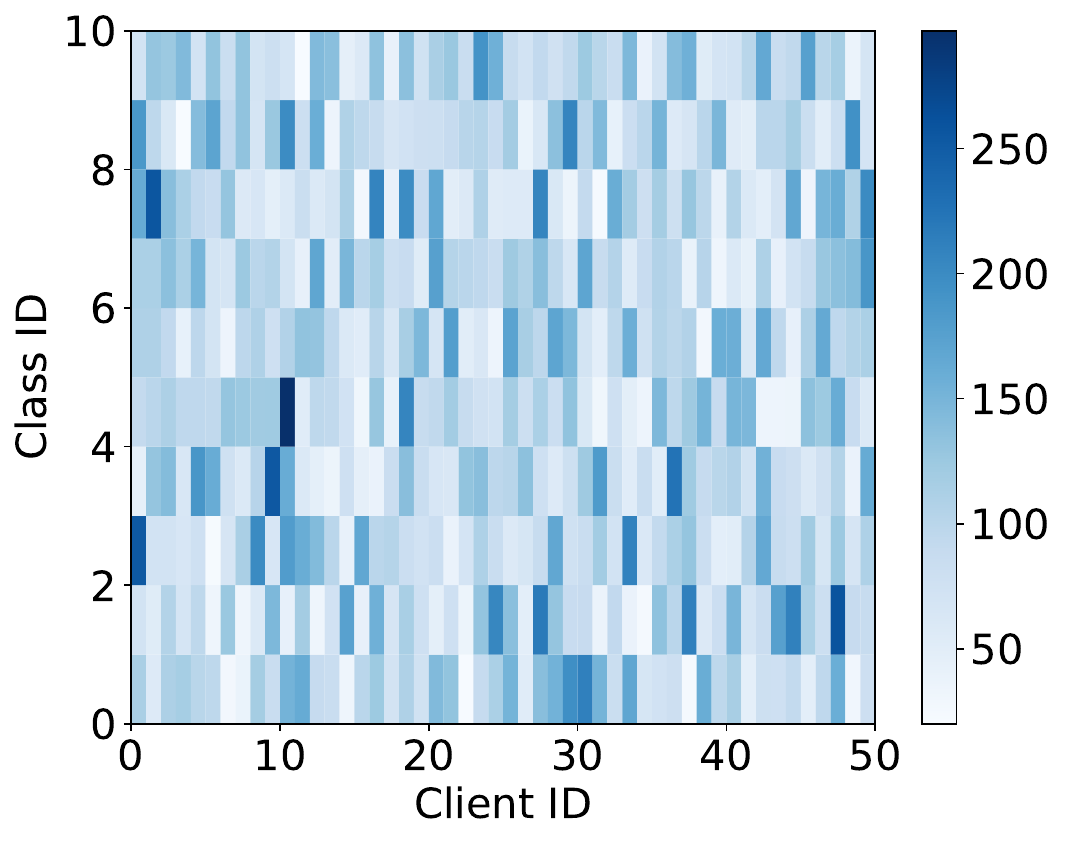}
         \caption{For 50 clients ($\beta=5$)}
         \label{fig:cifar-10_5_50}
     \end{subfigure}
        \caption{\textbf{Detailed data partitions on CIFAR-10 dataset using Dirichlet distribution.} Smaller $\beta$ corresponds to higher data heterogeneity. Best viewed in color.}
        \label{fig:cifar-10}
\end{figure*}

\paragraph{ViT-based LCT encoder}\label{para:lct_encoder}
We construct our LCT encoder by following the encoder design of the original ViT~\citep{dosovitskiy2020image}, which comprises MHSA and MLP blocks:
\begin{align}
    z'_l &= \mathrm{MHSA}(\mathrm{LN}(z_{l-1})) + z_{l-1}, \quad l = 1, 2, \dots, L \\
    z_l &= \mathrm{MLP}(\mathrm{LN}(z'_l)) + z'_l, \quad l = 1, 2, \dots, L
\end{align}
The MHSA block in the encoder receives the input embedding $z_0$ in the form of three parameters \textit{Query} ($Q$), \textit{Key} ($K$), and \textit{Value} ($V$).
Self-attention (SA) begins by computing the dot products of the \textit{Query} with all \textit{Keys} and applying a softmax function to obtain the attention weights on the \textit{Values}.
Multi-head enables the multiple calculations of self-attention in parallel with $h$ independent attention heads:
\begin{equation}
    \mathrm{MHSA}(Q,K,V) = \mathrm{Concat}(\mathrm{SA_1}, \mathrm{SA_2}, \dots, \mathrm{SA_h}).
\end{equation}

With the multi-head self-attention~(MHSA), our model can capture long-range positional information as each head can pay attention to different parts of the input embedding and capture different contextual information by calculating the associations between image patches in a unique way.
The MLP block contains two layers with GELU~\citep{hendrycks2016gelu} activation and dropouts.
LayerNorm (LN) is applied before each component of the encoder to prevent its values from changing abruptly, improving the training performance.
A residual connection is also applied to route the information between the components.
Then, the final output sequence $z_L$ is forwarded to the sequence pooling (SeqPool) layer.

\paragraph{Classifier}\label{para:classifier}
We apply SeqPool~\citep{hassani2021escaping} over the output sequence $z_L$ to eliminate the use of learnable class token:
\begin{equation}
    z = \sigma(\mathrm{FC}(\mathrm{LN}(z_L))) \times z_L,
\end{equation}
where LayerNorm (LN) is applied before passing the sequence to the fully connected (FC) layer, then attention weights are added by a softmax function $\sigma$.
As the output sequence contains relevant information across different parts of the input embedding, preserving this information via attention weights can improve the model performance with no additional parameters compared to the class token and thus significantly reduce the computation overhead.
Eventually, the output $z$ is sent to a single-layer classifier:
\begin{equation}
    y = \mathrm{FC}(z),
\end{equation}
where $y$ is the final classification output of our OnDev-LCT.


\section{Experiments}\label{sec:experiments}

In this section, we analyze the performance of our OnDev-LCTs over existing lightweight vision models by conducting extensive experiments on multiple image datasets in both centralized and federated learning scenarios.

\paragraph{Datasets}\label{para:datasets}
We conduct extensive experiments on six image classification datasets: CIFAR-10, CIFAR-100~\citep{krizhevsky2009cifar}, MNIST~\citep{deng2012mnist}, Fashion-MNIST~\citep{xiao2017fashion}, EMNIST-Balanced~\citep{cohen2017emnist}, and FEMNIST~\citep{caldas2018leaf}.
Since our study focuses on edge devices with limited resources for tackling vision tasks in low-data regimes, we do not use additional pre-training data for all experiments.
CIFAR-10 and CIFAR-100~\citep{krizhevsky2009cifar} datasets consist of 50,000 training samples and 10,000 test samples of $32 \times 32$ RGB images in 10 and 100 classes, respectively.
MNIST~\citep{deng2012mnist} and Fashion-MNIST~\citep{xiao2017fashion} datasets contain 60,000 training samples and 10,000 test samples of $28 \times 28$ single-channel images in 10 classes.
EMNIST-Balanced~\citep{cohen2017emnist} contains 112,800 training samples and 18,800 test samples of $28 \times 28$ single-channel handwritten character images in  47 balanced classes.
We use the original training and test sets of each dataset for the centralized scenario while dividing the training sets into 10 and 50 client data partitions for the FL scenarios.
To control the level of data heterogeneity for each partition, we adjust the concentration parameter $\beta$ of the Dirichlet distribution~\citep{ferguson1973bayesian} from $\{0.1, 0.5, 5\}$, where smaller $\beta$ corresponds to a higher degree of data heterogeneity.
Using this partitioning strategy, each FL client may have relatively few or no data samples in certain classes.
Figure~\ref{fig:cifar-10} visualizes the data distribution of 10 and 50 FL clients for the CIFAR-10 dataset.
Additionally, we evaluate the effectiveness of our OnDev-LCTs in realistic non-IID settings by using the Federated Extended MNIST (FEMNIST) dataset provided by the LEAF benchmark~\citep{caldas2018leaf}.
The FEMNIST dataset is commonly used in the FL literature, as it is designed to replicate practical FL scenarios for multi-class (i.e., 62 classes) image classification by naturally partitioning the total of 805,263 data samples in Extended MNIST (EMNIST)~\citep{cohen2017emnist} dataset into 3,550 clients based on the writer of the digit or character.
To align with prior FL works~\citep{caldas2018leaf, wu2021fedcg, chen2022pfl, shi2022mg}, we sub-sample 5\% of the total data (approximately 180 clients) and split it into 80\% for training and 20\% for testing per client.

\begin{table}[t]
\centering
\resizebox{0.9\columnwidth}{!}{%
\begin{tabular}{l|ccccc|c} 
\hline
\multicolumn{1}{c|}{\textbf{Model}} & \textbf{\#Trans} & \textbf{\#Convs} & \textbf{Patch} & \textbf{Pos\_Emb} & \textbf{SeqPool} & \textbf{\#Params}  \\ 
\hline
ViT-Lite-1/16              & 1        & \xmark         & $16\times16$  & \cmark  & \xmark    & 0.25M      \\
ViT-Lite-2/16             & 2        & \xmark         & $16\times16$  & \cmark  & \xmark    & 0.27M      \\
ViT-Lite-1/8               & 1        & \xmark         & $8\times8$    & \cmark  & \xmark    & 0.42M      \\
ViT-Lite-2/8               & 2        & \xmark         & $8\times8$    & \cmark  & \xmark    & 0.45M      \\
ViT-Lite-1/4               & 1        & \xmark         & $4\times4$    & \cmark  & \xmark    & 1.21M      \\
ViT-Lite-2/4               & 2        & \xmark         & $4\times4$    & \cmark  & \xmark    & 1.23M      \\ 
\hline
CCT-2/2                    & 2        & 2        & \xmark   & \cmark  & \cmark      & 0.28M      \\
CCT-4/2                    & 4        & 2        & \xmark   & \cmark  & \cmark      & 0.48M      \\ 
\hline
OnDev-LCT-1/1              & 1        & 1        & \xmark   & \xmark   & \cmark     & 0.21M      \\
OnDev-LCT-2/1              & 2        & 1        & \xmark   & \xmark   & \cmark     & 0.31M      \\
OnDev-LCT-4/1              & 4        & 1        & \xmark   & \xmark   & \cmark     & 0.51M      \\
OnDev-LCT-8/1              & 8        & 1        & \xmark   & \xmark   & \cmark     & 0.91M      \\ 
\hline
OnDev-LCT-1/3              & 1        & 3        & \xmark   & \xmark   & \cmark     & 0.25M      \\
OnDev-LCT-2/3              & 2        & 3        & \xmark   & \xmark   & \cmark     & 0.35M      \\
OnDev-LCT-4/3              & 4        & 3        & \xmark   & \xmark   & \cmark     & 0.55M      \\
OnDev-LCT-8/3              & 8        & 3        & \xmark   & \xmark   & \cmark     & 0.95M      \\
\hline
\end{tabular}
}%
\caption{\textbf{Detailed architecture of transformer-based model variants.} We follow the implementation of~\citet{hassani2021escaping} for ViT-Lite variants.}
\label{tab:model_variants}
\end{table}

\paragraph{Implementation details}\label{para:implementation}
We implement our experiments using the TensorFlow~\citep{abadi2015tensorflow} framework.
All experiments are conducted on a single NVIDIA RTX 3080 GPU with 10 GB memory.
We consider two client sampling strategies for experiments in FL scenarios (except for the FEMNIST dataset): (1) we sample 10 clients where all the clients participate in every FL round, (2) we sample 50 clients and randomly choose 10 clients to participate in each FL round.
For CIFAR-10 and CIFAR-100~\citep{krizhevsky2009cifar}, we train each model for 200 epochs in centralized experiments and conduct 60 communication rounds for the FL scenario with 10 clients and 100 rounds for the FL scenario with 50 clients, with each client training for 5 local epochs.
For other datasets, except for FEMNIST, we train each model for 50 epochs in centralized experiments and conduct 30 communication rounds for the FL scenario with 10 clients and 60 rounds for the FL scenario with 50 clients, with each client training for 3 local epochs.
For FL experiments with the FEMNIST dataset, we conduct 120 communication rounds, with each client training for 3 local epochs per round.
During the communication rounds, we randomly select $C$ clients to participate in training, considering three different settings of  $C \in \{5, 10, 2 \sim 10\}$, which simulate common scenarios where clients may be offline or have limited connectivity to participate in every round. 
The setting 2 $\sim$ 10 represents the number of clients selected in each round, varying from 2 clients to 10 clients.
We conduct 5 separate runs for each centralized experiment and report the best accuracy values.
For FL, reported values are the best median values from the last 5 FL rounds of 3 separate runs.

\paragraph{Model variants and baselines}\label{para:models}
We explore several variants of OnDev-LCT by tuning the number of LCT encoders and the number of standard convolutions in the LCT tokenizer.
For instance, ``OnDev-LCT-2/3" specifies a variant with 2 LCT encoders and 3 standard convolutions.
We compare the performance of our OnDev-LCTs with existing lightweight vision models as baselines.
For pure CNN baselines, we implement small-sized variants of ResNet~\citep{he2016resnet} and MobileNetv2~\citep{sandler2018mobilenetv2}, which are specially designed for CIFAR, denoted by ``ResNet-$r$" where $r$ is the number of layers and ``MobileNetv2/$\alpha$" where $\alpha$ is the width multiplier, respectively.
Following~\citep{hassani2021escaping}, we design several ``ViT-Lite" model variants with pure backbones of the vision transformer~\citep{dosovitskiy2020image} by varying numbers of transformer layers and patch sizes.
We also implement small ``CCT" variants~\citep{hassani2021escaping} (only with $3 \times 3$ convolution kernels), denoted with the number of transformer encoders followed by the number of convolutional blocks.
For all the transformer-based models, we set the embedding dimension as 128 and use 4 attention heads.
Table~\ref{tab:model_variants} shows the detailed architecture of transformer-based model variants.

\begin{table}[t]
\centering
\resizebox{0.9\columnwidth}{!}{%
\begin{tabular}{l|cccccc|c} 
\hline
\multicolumn{1}{c|}{\textbf{Model Family}} & \begin{tabular}[c]{@{}c@{}}\textbf{CIFAR}\\\textbf{10}\end{tabular} & \begin{tabular}[c]{@{}c@{}}\textbf{CIFAR}\\\textbf{100}\end{tabular} & \textbf{MNIST} & \begin{tabular}[c]{@{}c@{}}\textbf{Fashion}\\\textbf{MNIST}\end{tabular} & \begin{tabular}[c]{@{}c@{}}\textbf{EMNIST}\\\textbf{Balanced}\end{tabular} & \textbf{FEMNIST} & \begin{tabular}[c]{@{}c@{}}\textbf{Optimizer}\\\textbf{Type}\end{tabular}  \\ 
\hline
ResNet                   & $5\mathrm{e}{-3}$     & $5\mathrm{e}{-3}$      & $5\mathrm{e}{-3}$     & $5\mathrm{e}{-3}$     & $5\mathrm{e}{-3}$  & $1\mathrm{e}{-3}$   & Adam   \\
MobileNetv2    & $5\mathrm{e}{-3}$     & $5\mathrm{e}{-3}$      & $3\mathrm{e}{-3}$     & $5\mathrm{e}{-3}$     & $5\mathrm{e}{-3}$   & $3\mathrm{e}{-3}$  & Adam   \\
ViT-Lite         & $1\mathrm{e}{-3}$     & $1\mathrm{e}{-3}$      & $1\mathrm{e}{-3}$     & $1\mathrm{e}{-3}$     & $1\mathrm{e}{-3}$   & $1\mathrm{e}{-3}$  & AdamW  \\
CCT               & $1\mathrm{e}{-3}$     & $1\mathrm{e}{-3}$      & $1\mathrm{e}{-3}$     & $1\mathrm{e}{-3}$     & $1\mathrm{e}{-3}$   & $1\mathrm{e}{-3}$  & AdamW  \\
OnDev-LCT                                    & $3\mathrm{e}{-3}$     & $5\mathrm{e}{-3}$      & $1\mathrm{e}{-3}$     & $3\mathrm{e}{-3}$   & $3\mathrm{e}{-3}$  & $1\mathrm{e}{-3}$     & Adam   \\
\hline
\end{tabular}
}%
\caption{\textbf{Learning rate values and optimizer types for each model family.} Models with AdamW use the default values of $\beta_1=0.9$, $\beta_2=0.999$, and weight decay = $1\mathrm{e}{-4}$.}
\label{tab:lr_optimizers}
\end{table}

\begin{table*}[t]
\centering
\resizebox{0.92\linewidth}{!}{%
    \begin{tabular}{|L{2.7cm}|ccccc|C{1.5cm}|C{1.5cm}|} 
    \hline
    \multicolumn{1}{|c|}{\textbf{Model}} & \makecell{\textbf{CIFAR-10}} & \makecell{\textbf{CIFAR-100}} & \textbf{MNIST} & \makecell{\textbf{Fashion}\\\textbf{MNIST}} & \makecell{\textbf{EMNIST}\\\textbf{Balanced}} & \multicolumn{1}{c|}{\textbf{\#Params}} & \textbf{MACs}  \\ 
    \hline
    
    \multicolumn{8}{|c|}{\textit{Convolutional Neural Networks (Designed for CIFAR)}}                                                     \\ 
    \hline
    
    ResNet-20            & 82.30                        & 42.62                         & 99.49                         & 92.45                         & \underline{88.38}         & 0.27M          & 0.04G         \\
    ResNet-32            & 82.68                        & 46.75                         & 99.50                         & \underline{92.82}             & 88.18                     & 0.47M          & 0.07G         \\
    ResNet-44            & \underline{83.53}            & \underline{48.79}             & \textbf{\underline{99.58}}    & 92.46                         & 88.07                     & 0.67M          & 0.10G         \\
    ResNet-56            & 83.03                        & 48.14                         & 99.50                         & 92.48                         & 88.14                     & 0.86M          & 0.13G         \\
    \hline
    
    MobileNetv2/0.2     & 71.79                         & 37.99                         & 99.06                         & 90.16                         & 85.42                     & 0.21M          & $<$0.01G         \\
    MobileNetv2/0.5     & 77.10                         & 41.71                         & 99.11                         & 89.70                         & \underline{87.66}         & 0.72M          & $<$0.01G         \\
    MobileNetv2/0.75    & 80.10                         & 44.37                         & 99.16                         & 90.13                         & 86.75                     & 1.39M          & $<$0.01G         \\
    MobileNetv2/1.0     & \underline{80.50}             & \underline{44.51}             & \underline{99.34}             & \underline{90.53}             & 87.49                     & 2.27M          & 0.01G         \\
    \hline
    
    \multicolumn{8}{|c|}{\textit{Lite Vision Transformers}}                                                                                     \\ 
    \hline
    
    
    ViT-Lite-1/16       & 53.71                         & 25.24                         & 97.66                         & 87.37                         & 82.02                     & 0.25M          & $<$0.01G        \\
    ViT-Lite-2/16       & 53.74                         & 25.21                         & 97.66                         & 87.67                         & 82.23                     & 0.27M          & $<$0.01G        \\
    
    ViT-Lite-1/8        & 61.67                         & 32.51                         & 98.04                         & 87.01                         & 83.35                     & 0.42M          & 0.01G        \\
    ViT-Lite-2/8        & 62.60                         & 33.46                         & 97.97                         & 87.30                         & 83.35                     & 0.45M          & 0.02G        \\
    
    ViT-Lite-1/4        & 68.66                         & 38.43                         & 98.63                         & 90.25                         & \underline{86.22}                     & 1.21M          & 0.04G        \\
    ViT-Lite-2/4        & \underline{71.03}             & \underline{41.98}             & \underline{98.65}             & \underline{90.70}             & 86.05                     & 1.23M          & 0.07G        \\
    \hline
    
    \multicolumn{8}{|c|}{\textit{Compact Convolutional Transformers}}                                                                       \\ 
    \hline
    
    CCT-2/2           & 79.71                       & 50.75                         & 99.17                         & 91.37                         & 88.93                     & 0.28M          & 0.03G          \\
    CCT-4/2           & \underline{80.92}           & \underline{53.23}             & \underline{99.20}             & \underline{91.73}             & \underline{89.41}         & 0.48M          & 0.05G          \\
    \hline
    
    \multicolumn{8}{|c|}{\textbf{\textit{On-Device Lightweight Convolutional Transformers (Ours)}}}                                                  \\ 
    \hline
    
    OnDev-LCT-1/1             & 84.55                       & 57.69                         & 99.46                         & 92.73                         & 89.52                     & 0.21M          & 0.03G      \\
    OnDev-LCT-2/1             & 86.27                       & 59.17                         & 99.39                         & 92.50                         & 89.18                     & 0.31M          & 0.04G      \\
    OnDev-LCT-4/1             & 86.61                       & 61.36                         & \underline{99.53}             & 92.70                         & 89.39                     & 0.51M          & 0.05G      \\
    OnDev-LCT-8/1             & 86.64                       & 62.62                         & 99.49                         & 92.59                         & 89.55                     & 0.91M          & 0.08G      \\
    \hline
    
    OnDev-LCT-1/3             & 85.73                       & 57.66                         & 99.41                         & 92.72                         & 88.96                     & 0.25M          & 0.07G     \\
    OnDev-LCT-2/3             & 86.04                       & 60.26                         & 99.46                         & \textbf{\underline{93.31}}    & \textbf{\underline{89.60}}& 0.35M          & 0.08G     \\
    OnDev-LCT-4/3             & 87.03                       & 61.95                         & 99.38                         & 92.60                         & 89.49                     & 0.55M          & 0.09G     \\
    OnDev-LCT-8/3             & \textbf{\underline{87.65}}  & \textbf{\underline{62.91}}    & 99.28                         & 92.82                         & 89.45                     & 0.95M          & 0.12G     \\
    \hline
    
    \end{tabular}%
    }%
    \caption{\textbf{Comparison on image classification performance in the centralized setting.} All reported accuracy (\%) values are the best of 5 runs. All experiments are conducted without applying data augmentation or learning rate schedulers. The best accuracy is marked in bold. The best entry of each model family is underlined.}
\label{tab:centralized_top1_acc}
\end{table*}

\begin{figure*}[t]
    \centering
    \begin{subfigure}[b]{0.195\linewidth}
        \centering
        \includegraphics[width=0.95\textwidth]{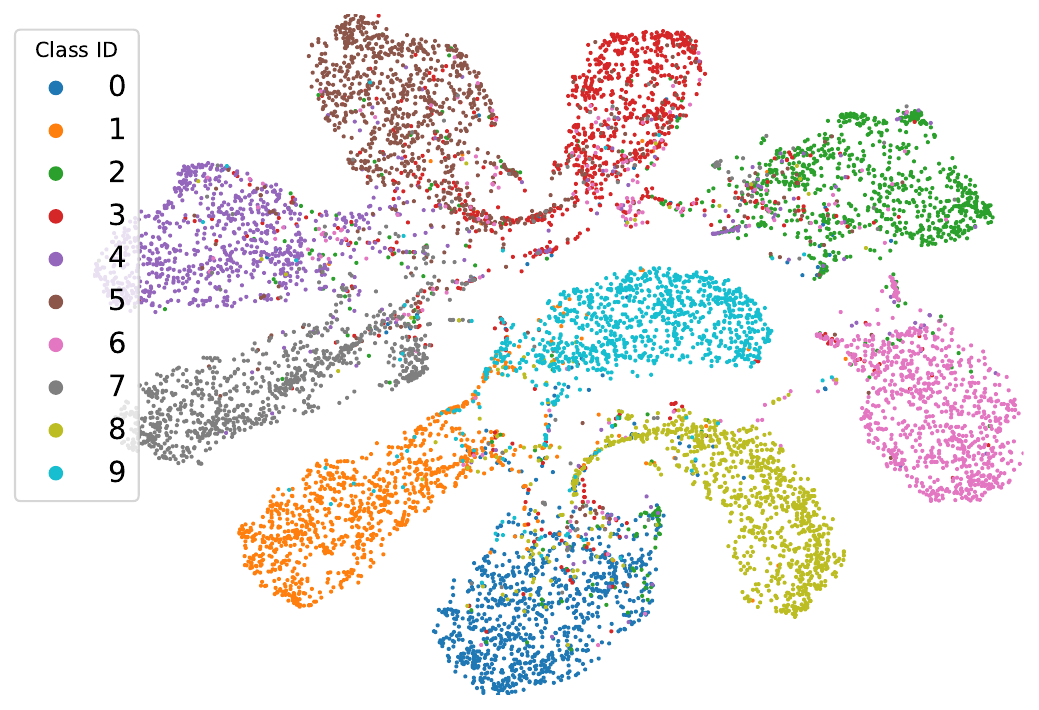}
        \caption[]%
        {{\small OnDev-LCT-1/1}}    
        \label{fig:cent_tsne_lct}
    \end{subfigure}
    \begin{subfigure}[b]{0.195\linewidth}  
        \centering 
        \includegraphics[width=0.95\textwidth]{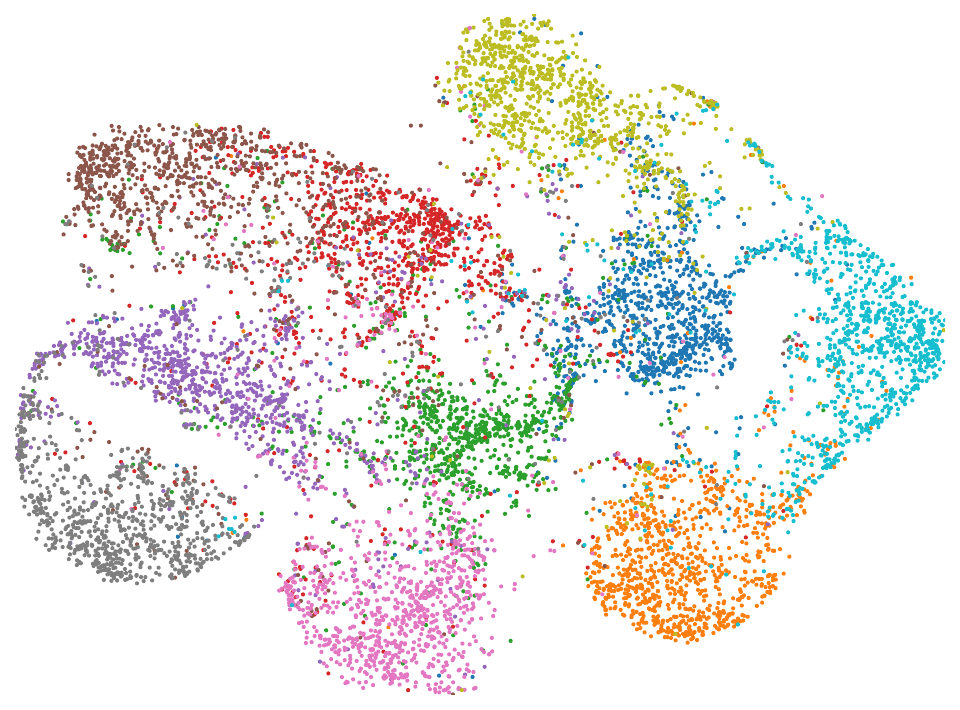}
        \caption[]%
        {{\small CCT-2/2}}    
        \label{fig:cent_tsne_cct}
    \end{subfigure}
    \begin{subfigure}[b]{0.195\linewidth}   
        \centering 
        \includegraphics[width=0.95\textwidth]{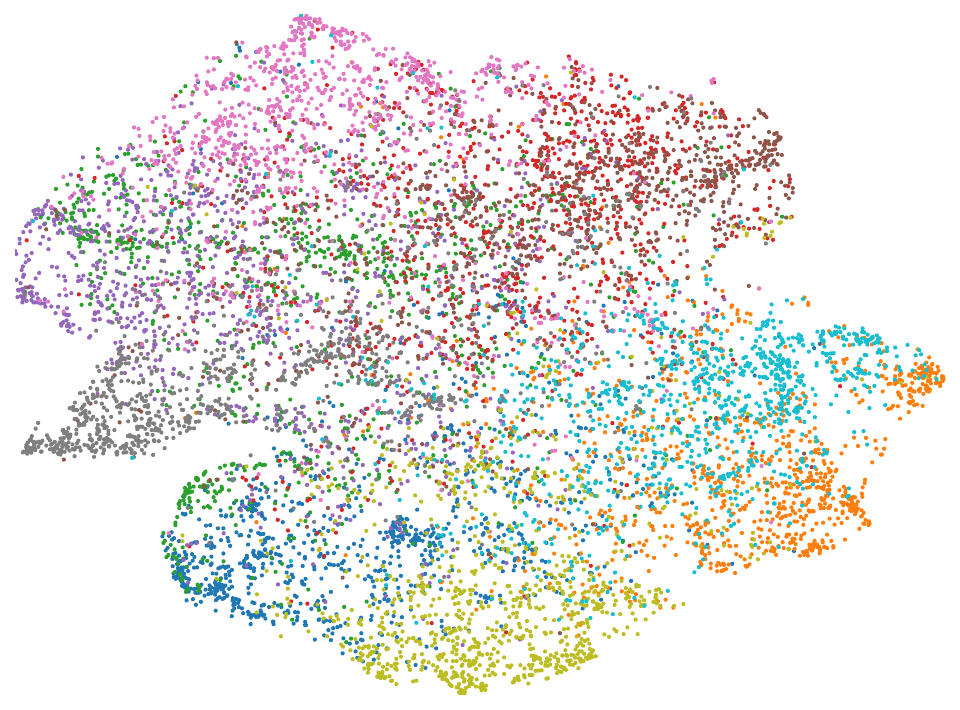}
        \caption[]%
        {{\small ViT-1/8}}    
        \label{fig:cent_tsne_vit}
    \end{subfigure}
    \begin{subfigure}[b]{0.195\linewidth}   
        \centering 
        \includegraphics[width=0.95\textwidth]{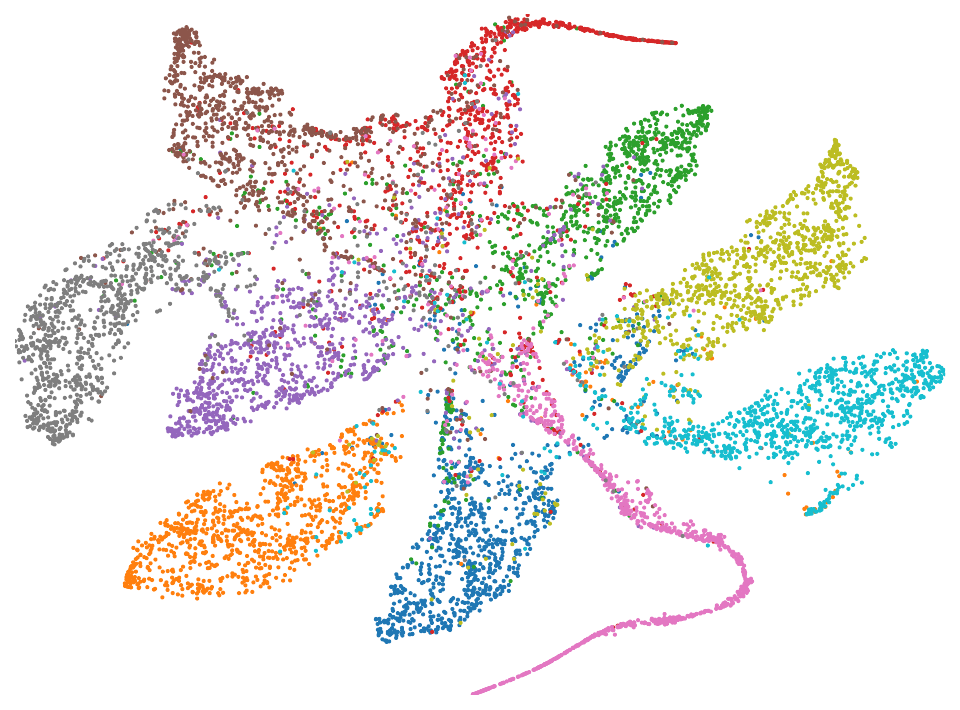}
        \caption[]%
        {{\small ResNet-20}}    
        \label{fig:cent_tsne_rn}
    \end{subfigure}
    \begin{subfigure}[b]{0.195\linewidth}   
        \centering 
        \includegraphics[width=0.95\textwidth]{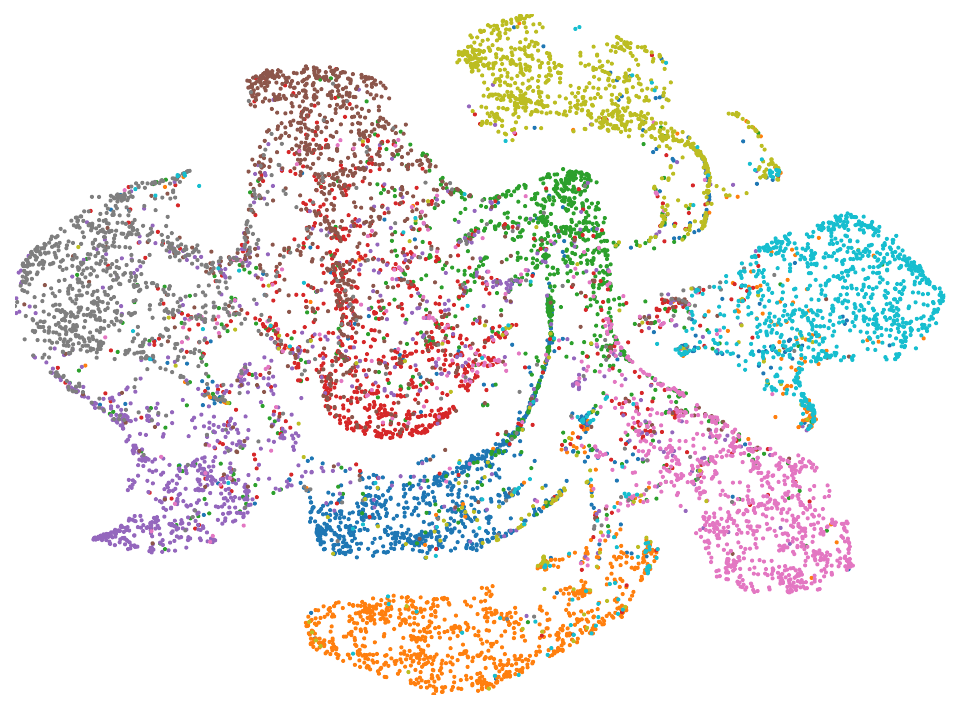}
        \caption[]%
        {{\small MobileNetv2/0.2}}    
        \label{fig:cent_tsne_mnv2}
    \end{subfigure}
    \caption[]
    {\small \textbf{t-SNE visualizations} of feature embeddings on CIFAR-10 test set learned by centralized models. Best viewed in color.}
    \label{fig:cent_tsne}
\end{figure*}

\begin{table}[t]
\centering
\resizebox{\columnwidth}{!}{%
\begin{tabular}{|l|cccccc|}
\hline
\multicolumn{1}{|c|}{\multirow{3}{*}{\textbf{Model}}} & \multicolumn{6}{c|}{\textbf{FedAvg with 10 clients}} \\
\cline{2-7} 
\multicolumn{1}{|c|}{}                                & \multicolumn{3}{c|}{\textbf{CIFAR-10}}                   & \multicolumn{3}{c|}{\textbf{CIFAR-100}} \\
\cline{2-7} 
\multicolumn{1}{|c|}{}                                & $\bm{\beta=0.1}$      & $\bm{\beta=0.5}$      & \multicolumn{1}{c|}{$\bm{\beta=5}$} & $\bm{\beta=0.1}$  & $\bm{\beta=0.5}$ & $\bm{\beta=5}$     \\ \hline
ResNet-20                         & 67.55                 & 75.43                 & \multicolumn{1}{c|}{77.72}          & 27.89             & 37.08          & 37.53                \\
ResNet-32                         & 71.55                 & 78.05                 & \multicolumn{1}{c|}{80.39}          & 31.34             & 41.33          & 42.68                \\
ResNet-44                         & 72.56                 & 77.73                 & \multicolumn{1}{c|}{80.55}          & 32.85             & 43.60          & 44.81                 \\ \hline
MobileNetv2/0.2         & 56.55                 & 67.07                 & \multicolumn{1}{c|}{69.44}          & 21.52             & 30.57          & 31.67                \\
MobileNetv2/0.5         & 60.50                 & 72.35                 & \multicolumn{1}{c|}{74.27}          & 23.40             & 33.28          & 32.58                \\ \hline
ViT-Lite-1/8              & 48.53                 & 58.05                 & \multicolumn{1}{c|}{59.70}          & 17.85             & 27.60          & 31.37                \\
ViT-Lite-2/8              & 49.03                 & 58.57                 & \multicolumn{1}{c|}{60.02}          & 17.99             & 28.20          & 32.32                \\ \hline
CCT-2/2                    & 66.45                 & 74.86                 & \multicolumn{1}{c|}{78.42}          & 35.21             & 47.99          & 50.50                \\
CCT-4/2                    & 66.38                 & 75.36                 & \multicolumn{1}{c|}{78.55}          & 39.44             & 50.43          & 52.76                \\ \hline
OnDev-LCT-1/1                                         & 76.02                 & 82.25                 & \multicolumn{1}{c|}{84.24}          & 42.64             & 55.23          & 58.07                \\
OnDev-LCT-2/1                                         & \textbf{76.85}        & \textbf{82.81}        & \multicolumn{1}{c|}{\textbf{84.95}} & \textbf{44.83}    & \textbf{56.74} & \textbf{59.53}       \\
OnDev-LCT-4/1                                         & \textbf{78.10}        & \textbf{84.01}        & \multicolumn{1}{c|}{\textbf{86.12}} & \textbf{47.51}    & \textbf{58.85} & \textbf{61.79}       \\
OnDev-LCT-8/1                                         & \textbf{79.70}         & \textbf{84.74}         & \multicolumn{1}{c|}{\textbf{86.77}}  & \textbf{49.75}     & \textbf{59.71}  & \textbf{62.42}        \\ \hline
\end{tabular}
}
\caption{\textbf{Comparison on image classification performance in FL with 10 clients} under different data heterogeneity settings.
All reported accuracy (\%) values are the best of 3 runs.
Three best accuracy values are marked in bold.}
\label{tab:fed1}
\end{table}

\begin{table}[t]
\centering
\resizebox{\columnwidth}{!}{%
\begin{tabular}{|l|cccccc|}
\hline
\multicolumn{1}{|c|}{\multirow{3}{*}{\textbf{Model}}} & \multicolumn{6}{c|}{\textbf{FedAvg with 50 clients}} \\
\cline{2-7} 
\multicolumn{1}{|c|}{}                                & \multicolumn{3}{c|}{\textbf{CIFAR-10}}                   & \multicolumn{3}{c|}{\textbf{CIFAR-100}} \\
\cline{2-7} 
\multicolumn{1}{|c|}{}                                & $\bm{\beta=0.1}$      & $\bm{\beta=0.5}$      & \multicolumn{1}{c|}{$\bm{\beta=5}$} & $\bm{\beta=0.1}$  & $\bm{\beta=0.5}$ & $\bm{\beta=5}$     \\ \hline
ResNet-20                         & 53.44                 & 69.62                 & \multicolumn{1}{c|}{74.84}          & 26.64             & 33.89          & 34.93                \\
ResNet-32                         & 55.79                 & 70.50                 & \multicolumn{1}{c|}{76.17}          & 26.27             & 36.89          & 38.02                \\
ResNet-44                         & 55.78                  & 71.08                  & \multicolumn{1}{c|}{76.10}           & 28.91              & 36.44           & 39.07                 \\ \hline
MobileNetv2/0.2         & 44.30                 & 55.97                 & \multicolumn{1}{c|}{60.84}          & 13.66             & 16.15          & 18.06                \\
MobileNetv2/0.5         & 46.59                 & 62.36                 & \multicolumn{1}{c|}{63.29}          & 12.19             & 18.52          & 18.83                \\ \hline
ViT-Lite-1/8              & 36.22                 & 52.94                 & \multicolumn{1}{c|}{57.05}          & 14.02             & 24.68          & 28.06                \\
ViT-Lite-2/8              & 36.85                 & 53.13                 & \multicolumn{1}{c|}{57.28}          & 14.44             & 25.39          & 28.22                \\ \hline
CCT-2/2                    & 55.67                 & 68.87                 & \multicolumn{1}{c|}{73.45}          & 33.51             & 42.17          & 45.59                \\
CCT-4/2                    & 56.05                 & 69.82                 & \multicolumn{1}{c|}{74.52}          & 35.26             & 43.68          & 46.95                \\ \hline
OnDev-LCT-1/1                                         & 66.18                 & 76.93                 & \multicolumn{1}{c|}{79.71}          & 32.69             & 49.34          & 53.83                \\
OnDev-LCT-2/1                                         & \textbf{68.25}        & \textbf{78.20}        & \multicolumn{1}{c|}{\textbf{80.55}} & \textbf{35.66}    & \textbf{52.10} & \textbf{54.71}       \\
OnDev-LCT-4/1                                         & \textbf{67.24}        & \textbf{77.68}        & \multicolumn{1}{c|}{\textbf{80.87}} & \textbf{38.82}    & \textbf{52.64} & \textbf{56.15}       \\
OnDev-LCT-8/1                                         & \textbf{68.44}         & \textbf{78.28}         & \multicolumn{1}{c|}{\textbf{82.42}}  & \textbf{42.23}     & \textbf{54.97}  & \textbf{58.18}        \\ \hline
\end{tabular}
}
\caption{\textbf{Comparison on image classification performance in FL with 50 clients} under different data heterogeneity settings.
All reported accuracy (\%) values are the best of 3 runs.
Three best accuracy values are marked in bold.}
\label{tab:fed2}
\end{table}

\paragraph{Hyperparameter selection}\label{para:hyperparameter}
We perform manual hyperparameter tuning for each model family and apply the best settings.
In the centralized scenario, the training batch size is set to 256 for EMNIST-Balanced~\citep{cohen2017emnist} and 128 for others.
In FL experiments, we reduce the batch size by half, i.e., 128 for EMNIST-Balanced~\citep{cohen2017emnist} and 64 for the rest.
It is important to note that, we set the training batch size to 16 for all experiments with the FEMNIST dataset.
We use Adam~\citep{kingma2014adam} optimizer for OnDev-LCTs and CNN baselines~\citep{he2016resnet, sandler2018mobilenetv2} while AdamW~\citep{loshchilov2017adamw} with weight decay $=1\mathrm{e}{-4}$ is used for CCT~\citep{hassani2021escaping} and ViT-Lite~\citep{dosovitskiy2020image} variants.
We manually tune the learning rate and use the best value for each model family.
The selected learning rate values and optimizers are shown in Table~\ref{tab:lr_optimizers}.
Since we are focusing on a fair comparison of the model performance, we train all the models from scratch without pre-training or applying data augmentation techniques and learning rate schedulers for all experiments by default.
These techniques could challenge the fairness of evaluation and provide sensitive results for tuning hyperparameters, especially in FL scenarios.
However, label smoothing with a probability of 0.1 is used during the training of all model variants.

\paragraph{Comparison in centralized scenario}\label{para:cen_exp}
Table~\ref{tab:centralized_top1_acc} reports image classification performance of our OnDev-LCTs compared to existing lightweight vision models in a centralized scenario.
The results show that our OnDev-LCT variants significantly outperform the other baselines while having lower computational demands.
Even our smallest OnDev-LCT-1/1 with 0.21M parameters can perform better in most cases compared to other model variants.
Our larger variant OnDev-LCT-8/3 further improves the performance, especially on CIFAR-10 and CIFAR-100~\citep{krizhevsky2009cifar}, achieving $4.62\%$ and $14.77\%$ better than ResNet-56~\citep{he2016resnet}, which has a comparable model size and computational demands.
Figure~\ref{fig:cent_tsne} shows t-SNE~\citep{JMLR:v9:vandermaaten08a} visualizations of feature embeddings learned by each model on the CIFAR-10~\citep{krizhevsky2009cifar} test data.
Visualizations show that features learned by our proposed OnDev-LCT model are more divergent and separable than those learned by other baselines.

\paragraph{Comparison in FL scenarios}\label{para:fl_exp}
We conduct experiments in FL scenarios with small variants of each model family, constraining the number of parameters to 1M (millions) and the computation (i.e., MACs) to 0.1G (billions).
For ViT-Lite~\citep{hassani2021escaping} models, instead of choosing the smallest variants, we choose larger variants with smaller patch sizes that can achieve better accuracy in centralized experiments. 
Table~\ref{tab:fed1} and Table~\ref{tab:fed2} provide the image classification performance of our OnDev-LCTs on CIFAR-10 and CIFAR-100~\citep{krizhevsky2009cifar} datasets compared to the other baselines under various data heterogeneity settings of FL.
We can observe that our OnDev-LCTs significantly outperform other baselines in all data heterogeneity settings.
From Table~\ref{tab:fed1} with a 10-client scenario, most of the transformer-based model variants can achieve comparable or only slightly lower results than the centralized performance under less data heterogeneity settings with $\beta$ values of 5 and 0.5.
For CNNs, we can observe a significant gap between their centralized and FL performance, especially for the highest data heterogeneity setting with $\beta=0.1$.
Remarkably, all the existing lightweight baseline models suffer from data heterogeneity in FL, especially the ViT-Lite model variants.
Meanwhile, our OnDev-LCT models achieve the best performance and are more robust to data heterogeneity across clients than the other baselines.
Similarly, in Table~\ref{tab:fed2} with a 50-client scenario, we can observe that our OnDev-LCTs outperform the other baselines in all degrees of data heterogeneity.

The reason behind this capability stems from the innovative combination of convolutions in our LCT tokenizer and the MHSA mechanism in our LCT encoder, which enables our OnDev-LCTs to strike a balance between local feature extraction and global context representation.
The strength of our LCT tokenizer lies in its efficient depthwise separable convolutions and residual linear bottleneck blocks, as these convolutions capture relevant spatial information efficiently while significantly reducing the computational burden and model size.
By effectively extracting local features, our LCT tokenizer can capture essential details in image representations, thus enhancing the overall accuracy of the model.
Moreover, the MHSA mechanism in our LCT encoder facilitates learning global representations of images, making OnDev-LCTs better at recognizing unseen patterns and more adaptive to new and diverse data.
This adaptability is crucial in practical applications where the model may encounter images from different sources or domains.

\begin{figure*}[t]
    \centering
    \begin{subfigure}[b]{0.195\linewidth}
        \centering
        \includegraphics[width=0.95\textwidth]{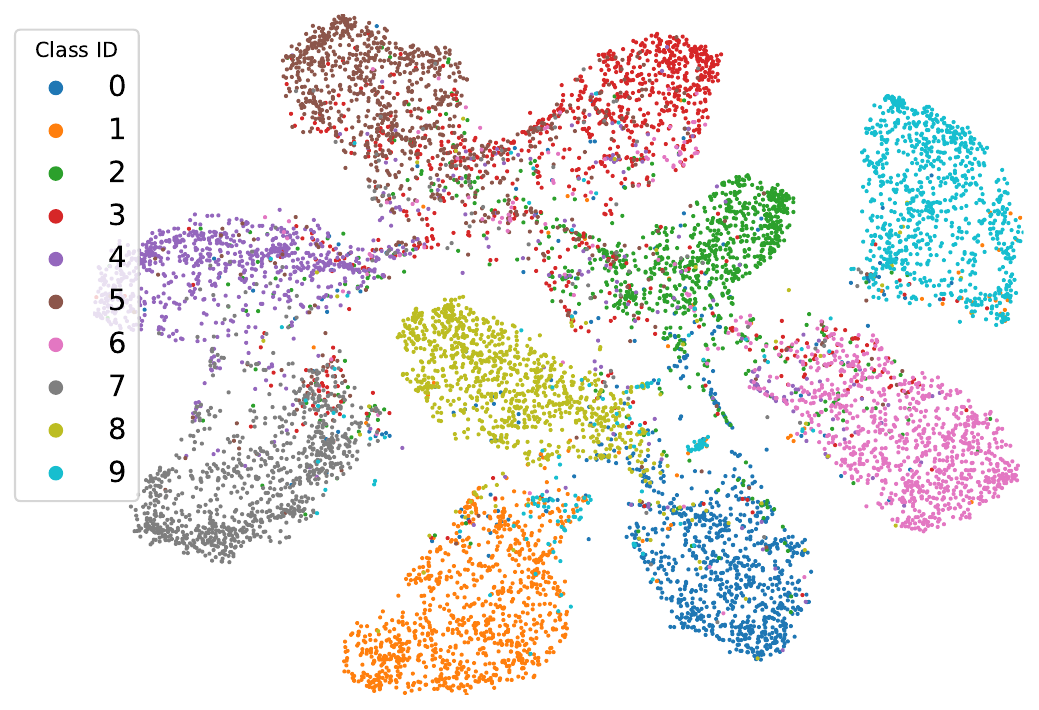}
        \caption[]%
        {{\small OnDev-LCT-1/1}}    
        \label{fig:global_tsne_lct}
    \end{subfigure}
    \begin{subfigure}[b]{0.195\linewidth}  
        \centering 
        \includegraphics[width=0.95\textwidth]{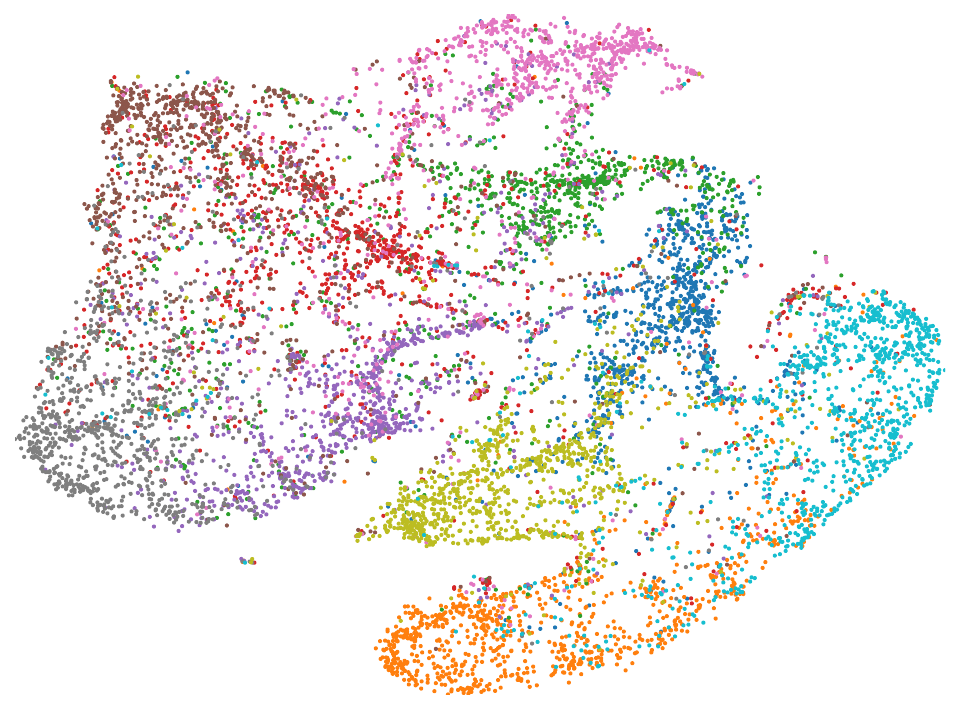}
        \caption[]%
        {{\small CCT-2/2}}    
        \label{fig:global_tsne_cct}
    \end{subfigure}
    \begin{subfigure}[b]{0.195\linewidth}   
        \centering 
        \includegraphics[width=0.95\textwidth]{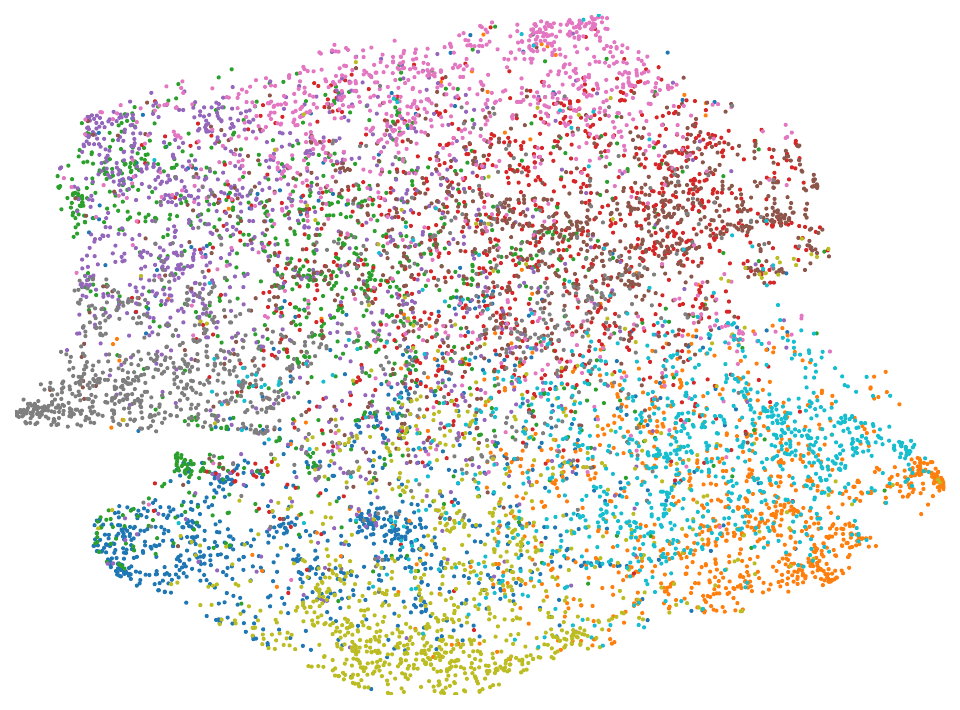}
        \caption[]%
        {{\small ViT-Lite-1/8}}    
        \label{fig:global_tsne_vit}
    \end{subfigure}
    \begin{subfigure}[b]{0.195\linewidth}   
        \centering 
        \includegraphics[width=0.95\textwidth]{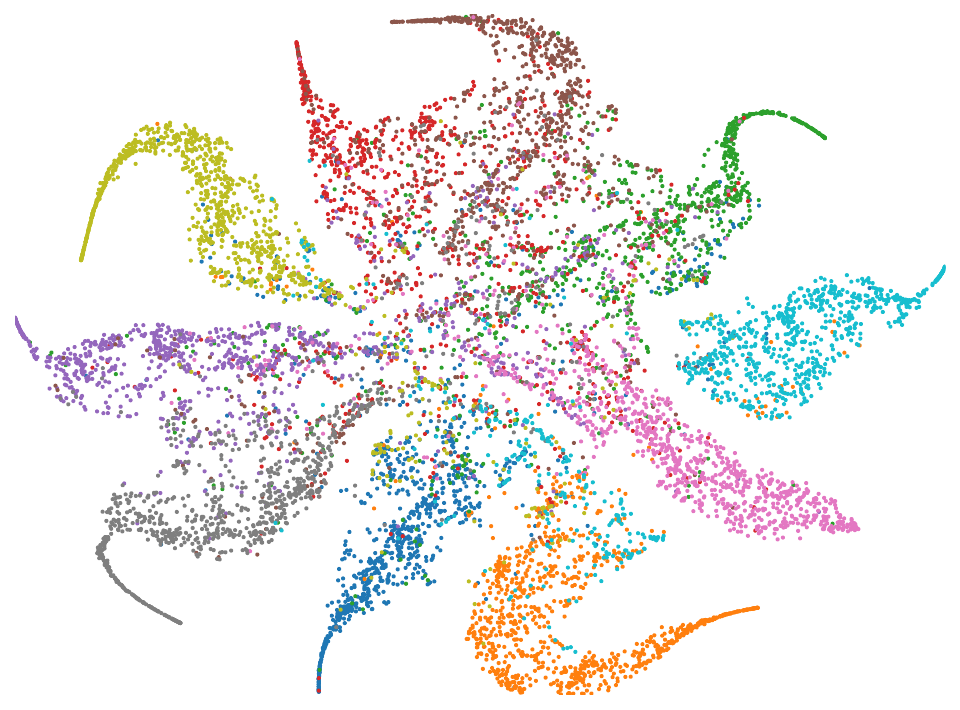}
        \caption[]%
        {{\small ResNet-20}}    
        \label{fig:global_tsne_rn}
    \end{subfigure}
    \begin{subfigure}[b]{0.195\linewidth}   
        \centering 
        \includegraphics[width=0.95\textwidth]{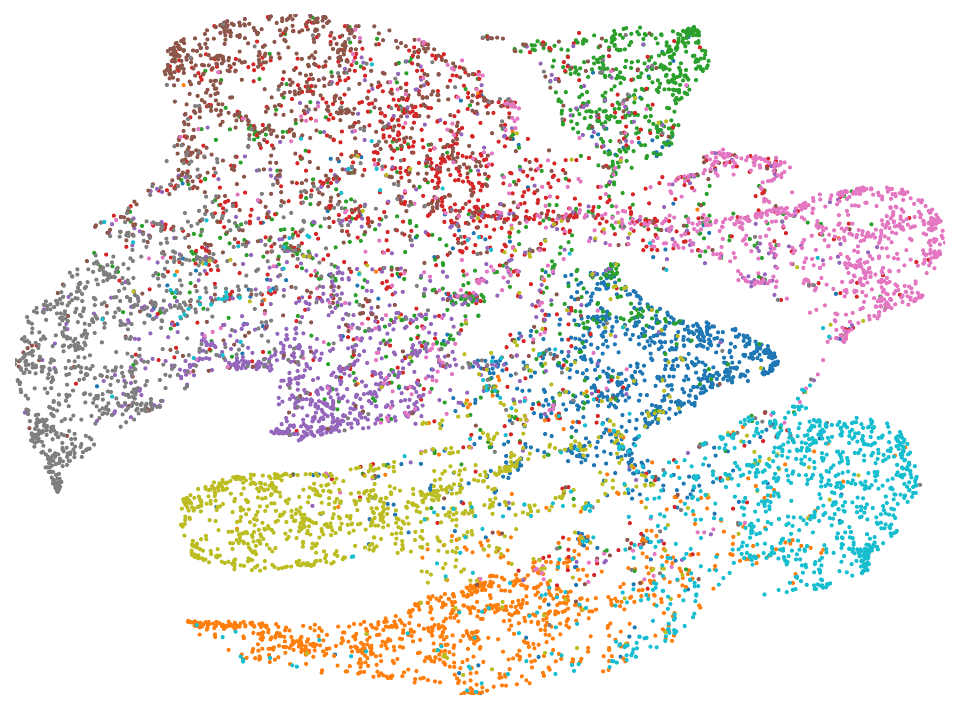}
        \caption[]%
        {{\small MobileNetv2/0.2}}    
        \label{fig:global_tsne_mnv2}
    \end{subfigure}
    \caption[]
    {\small \textbf{t-SNE visualizations} of feature embeddings on CIFAR-10 by FedAvg global models (10 clients with $\beta = 0.5$). Best viewed in color.}
    \label{fig:global_tsne}
\end{figure*}

\begin{figure*}[t]
    \centering
    \begin{subfigure}[b]{0.195\linewidth}
        \centering
        \includegraphics[width=0.95\textwidth]{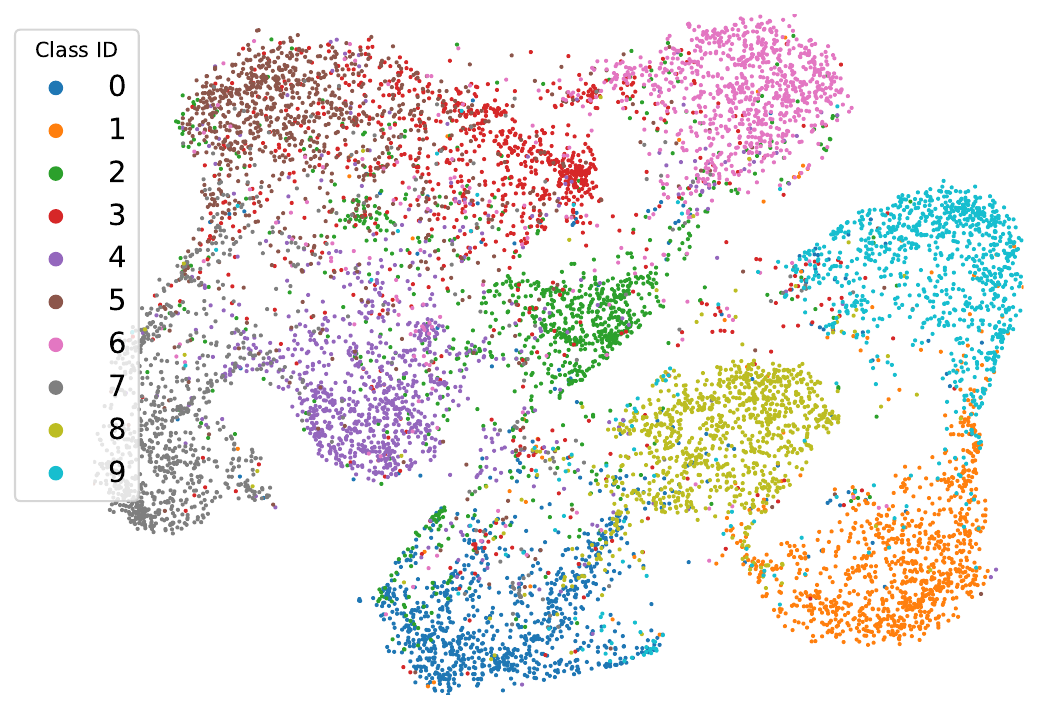}
        \caption[]%
        {{\small OnDev-LCT-1/1}}    
        \label{fig:global50_tsne_lct}
    \end{subfigure}
    \begin{subfigure}[b]{0.195\linewidth}  
        \centering 
        \includegraphics[width=0.95\textwidth]{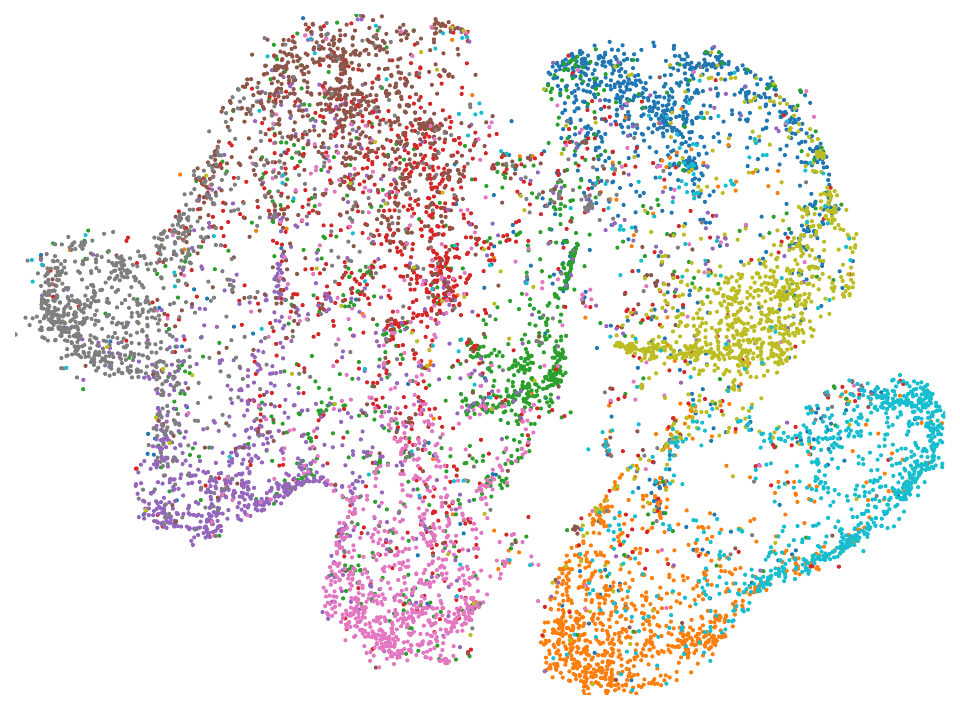}
        \caption[]%
        {{\small CCT-2/2}}    
        \label{fig:global50_tsne_cct}
    \end{subfigure}
    \begin{subfigure}[b]{0.195\linewidth}   
        \centering 
        \includegraphics[width=0.95\textwidth]{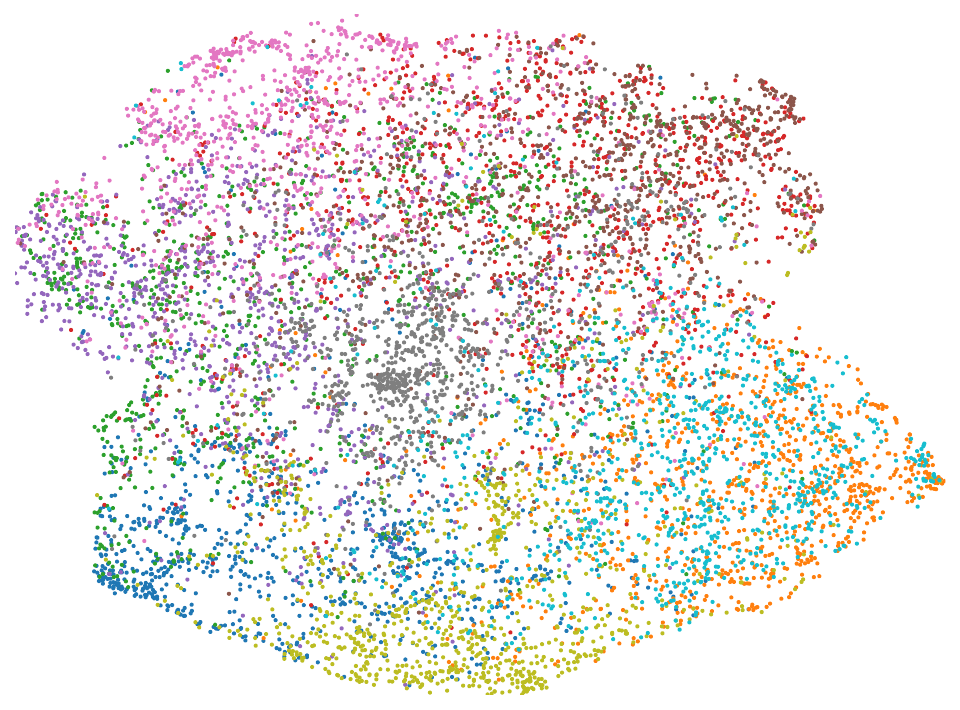}
        \caption[]%
        {{\small ViT-1/8}}    
        \label{fig:global50_tsne_vit}
    \end{subfigure}
    \begin{subfigure}[b]{0.195\linewidth}   
        \centering 
        \includegraphics[width=0.95\textwidth]{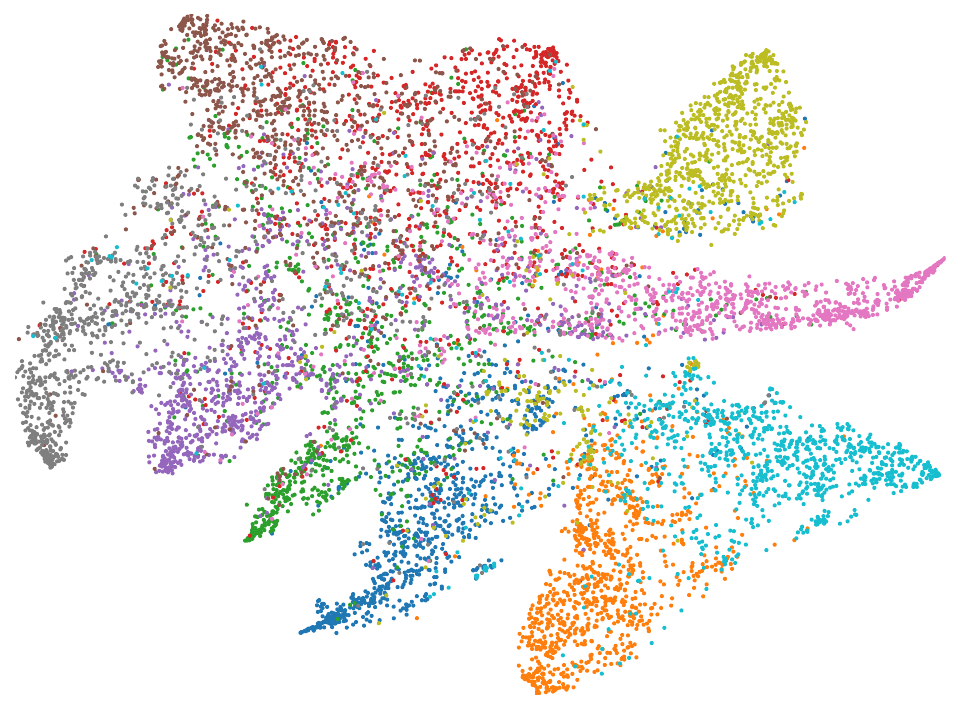}
        \caption[]%
        {{\small ResNet-20}}    
        \label{fig:global50_tsne_rn}
    \end{subfigure}
    \begin{subfigure}[b]{0.195\linewidth}   
        \centering 
        \includegraphics[width=0.95\textwidth]{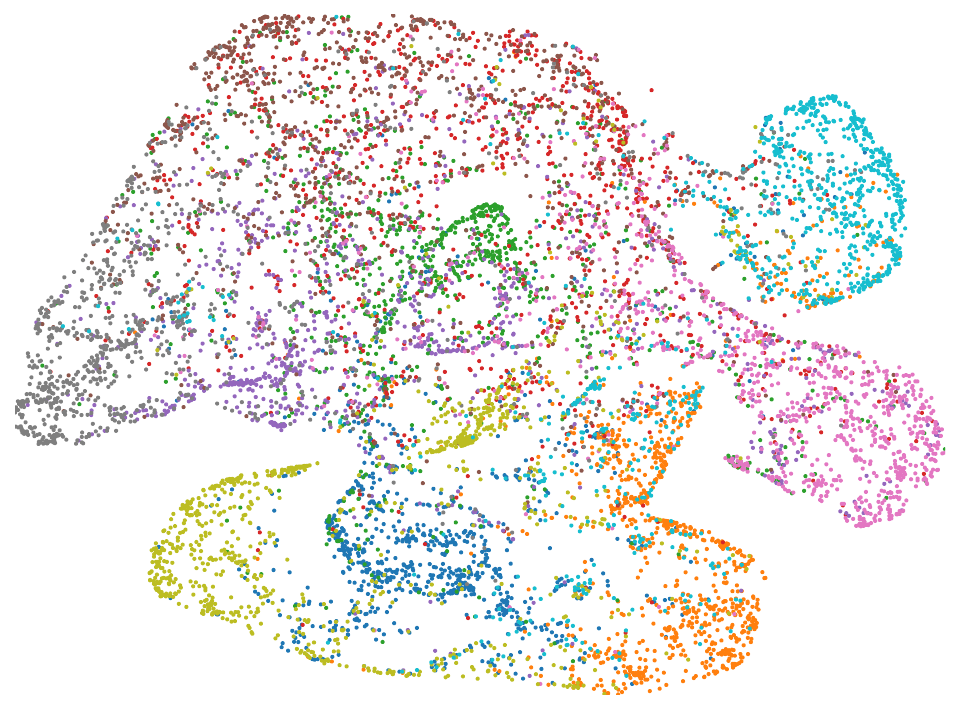}
        \caption[]%
        {{\small MobileNetv2/0.2}}    
        \label{fig:global50_tsne_mnv2}
    \end{subfigure}
    \caption[]
    {\small \textbf{t-SNE visualizations} of feature embeddings on CIFAR-10 by FedAvg global models (50 clients with $\beta = 0.5$). Best viewed in color.}
    \label{fig:global50_tsne}
\end{figure*}

In the context of FL, models are trained locally on individual devices with non-IID data distribution, and their updates are aggregated on the server.
This process often introduces challenges, such as data heterogeneity and communication bottlenecks.
However, the lightweight nature of our OnDev-LCT architecture allows for more efficient communication of model updates between client devices and the central server, mitigating the impact of communication constraints.
The advantage of OnDev-LCT can be attributed to its unique design and the integration of image-specific inductive biases through the convolutional tokenization step before the transformer encoder.
The incorporation of residual linear bottleneck blocks in the LCT tokenizer enhances the model's ability to adapt to local variations in data distribution, making it more resilient to non-IID data in the FL setting.
By leveraging these design choices, our OnDev-LCTs can learn more generalizable representations with fewer parameters, which is particularly beneficial in FL scenarios with limited data and communication constraints.
As a result, the proposed OnDev-LCT models demonstrate higher performance and robustness compared to other baselines under various data heterogeneity settings of FL.
However, the scope and applicability of OnDev-LCT are not confined to FL scenarios and, the flexibility and efficient nature of OnDev-LCT holds great potential in emerging fields like edge AI and on-device AI.
It is important to emphasize that the superior performance of the proposed OnDev-LCT highlights its potential value and broader applicability in FL and other resource-constrained on-device vision tasks across various domains.
Figure~\ref{fig:global_tsne} and Figure~\ref{fig:global50_tsne} show the t-SNE~\citep{JMLR:v9:vandermaaten08a} visualizations of feature embeddings learned by each global model under FedAvg setting ($\beta=0.5$) with 10 clients and 50 clients on the CIFAR-10~\citep{krizhevsky2009cifar} test data.

\begin{table}[t]
\centering
\resizebox{0.8\columnwidth}{!}{%
\begin{tabular}{|L{2.5cm}|C{1.5cm} C{1.5cm} C{1.5cm}|}
\hline 
\multicolumn{1}{|c|}{\multirow{2}{*}{\textbf{Model}}} & \multicolumn{3}{c|}{\textbf{No. of selected clients (C)}} \\ \cline{2-4} 
\multicolumn{1}{|c|}{}                       & \textbf{5}                      & \textbf{10}                      & \textbf{2$\sim$10}        \\ \hline
ResNet-20                                    & 78.94                  & 81.57                   & 78.29              \\
ResNet-32                                    & 79.82                  & \textbf{82.23}          & 79.27              \\
ResNet-44                                    & \textbf{80.01}         & \textbf{82.45}          & \textbf{80.20}     \\ \hline
MobileNetv2/0.2                              & 66.72                  & 68.68                   & 64.93              \\
MobileNetv2/0.5                              & 66.70                  & 74.60                   & 69.16              \\ \hline
ViT-Lite-1/8                                 & 59.75                  & 62.06                   & 59.57              \\
ViT-Lite-2/8                                 & 60.48                  & 63.48                   & 59.58              \\ \hline
CCT-2/2                                      & 78.58                  & 80.71                   & 78.78              \\
CCT-4/2                                      & 78.60                  & 81.28                   & \textbf{79.95}     \\ \hline
OnDev-LCT-1/1                                & 77.86                  & 80.68                   & 76.67              \\
OnDev-LCT-2/1                                & 79.57                  & 81.43                   & 76.56              \\
OnDev-LCT-4/1                                & \textbf{79.84}         & 81.84                   & 78.76              \\
OnDev-LCT-8/1                                & \textbf{80.33}         & \textbf{82.02}          & \textbf{80.06}     \\ \hline
\end{tabular}
}
\caption{\textbf{Comparison on image classification performance in realistic non-IID FL settings} using the FEMNIST dataset from LEAF benchmark.
All reported accuracy (\%) values are the best of 3 runs.
Three best accuracy values are marked in bold.
}
\label{tab:femnist}
\end{table}

\paragraph{Comparison in realistic FL scenarios}\label{para:femnist}
In this section, we conducted a performance comparison of our proposed OnDev-LCTs with other baseline models using the FEMNIST~\citep{caldas2018leaf} dataset. The results, as presented in Table~\ref{tab:femnist}, demonstrate that convolutional transformers, such as OnDev-LCTs and CCTs~\citep{hassani2021escaping}, achieve competitive performance on par with ResNet variants~\citep{he2016resnet}. ViT-Lite~\citep{hassani2021escaping} models, while being lightweight, show comparatively weaker performance, due to their limited ability to efficiently capture both local and global features, which are crucial for accurate image classification in realistic FL scenarios like FEMNIST. The accuracy of MobileNetv2 variants~\citep{sandler2018mobilenetv2} may not have reached its peak during the conducted training epochs, implying the potential for increased accuracy with further training. Nonetheless, our proposed OnDev-LCTs consistently show their impressive performance across various scenarios with different numbers of selected clients participating in each training round, thus demonstrating the effectiveness of our model design for on-device vision tasks in FL scenarios. The integration of efficient convolutions, self-attention mechanisms, and image-specific inductive biases empowers our models to strike a balance between local feature extraction and global context representation, contributing to their superior performance and efficiency. These compelling findings highlight the potential of our OnDev-LCT architecture as a practical solution for on-device vision tasks in FL with limited data and computational resources.

\begin{table*}[t]
\centering
\resizebox{0.9\linewidth}{!}{%
\begin{tabular}{|L{2.5cm}|c|C{2cm}|cccc|C{2cm}|C{2cm}|C{2cm}|}
\hline
\multicolumn{1}{|c|}{\multirow{3}{*}{\textbf{Model}}} & \multirow{3}{*}{\begin{tabular}[c]{@{}c@{}}\textbf{Quantization}\\ \textbf{method} \end{tabular}} & \multirow{3}{*}{\textbf{\#bit(W-A)}} & \multicolumn{4}{c|}{\textbf{CIFAR-10}}   & \multirow{3}{*}{\begin{tabular}[c]{@{}c@{}}\textbf{FEMNIST}\\ \textbf{($\bm{C=10}$)} \end{tabular}} & \multirow{3}{*}{\begin{tabular}[c]{@{}c@{}}\textbf{Inference}\\ \textbf{time (s)}\end{tabular}}    & \multirow{3}{*}{\textbf{\#Params}}    \\ \cline{4-7}
\multicolumn{1}{|c|}{}     &        &    & \multicolumn{1}{c|}{\multirow{2}{*}{\textbf{Centralized}}}    & \multicolumn{3}{c|}{\textbf{FedAvg with 50 clients}}      &                          &                    &  \\ \cline{5-7}
\multicolumn{1}{|c|}{}     &            & \multicolumn{1}{c|}{}          & \multicolumn{1}{c|}{}          & $\bm{\beta=0.1}$  & $\bm{\beta=0.5}$  & $\bm{\beta=5}$    &                          &                    &  \\ \hline
Q-ViT               & QAT                    & 4-4              & \multicolumn{1}{c|}{55.91}             & 37.89                 & 43.02             & 45.73             & 59.48                   & 3.50               & 22M      \\ \hline
MinMax              & PTQ                    & 8-8              & \multicolumn{1}{c|}{55.05}             & 31.04                 & 45.27             & 49.56             & 37.99                    & 2.42               & 22M \\
EMA                 & PTQ                    & 8-8              & \multicolumn{1}{c|}{52.86}             & 29.75                 & 45.08             & 51.19             & 50.73                    & 2.43               & 22M \\
Percentile          & PTQ                    & 8-8              & \multicolumn{1}{c|}{53.70}             & 29.82                 & 45.79             & 50.42             & 47.75                    & 2.46               & 22M \\
OMSE                & PTQ                    & 8-8              & \multicolumn{1}{c|}{52.69}             & 30.83                 & 45.66             & 50.24             & 49.50                    & 2.44               & 22M \\
FQ-ViT              & PTQ                    & 8-8              & \multicolumn{1}{c|}{51.72}             & 28.86                 & 44.52             & 49.19             & 49.37                    & 2.62               & 22M \\ \hline
OnDev-LCT-1/1       & None                   & 32-32            & \multicolumn{1}{c|}{84.55}             & 66.18                 & 76.93             & 79.71             & 80.68                    & 2.66               & 0.21M \\
OnDev-LCT-2/1       & None                   & 32-32            & \multicolumn{1}{c|}{\textbf{86.27}}    & \textbf{68.25}        & \textbf{78.20}    & \textbf{80.55}    & \textbf{81.43}           & 2.78               & 0.31M \\
OnDev-LCT-4/1       & None                   & 32-32            & \multicolumn{1}{c|}{\textbf{86.61}}    & \textbf{67.24}        & \textbf{77.68}    & \textbf{80.87}    & \textbf{81.84}           & 3.13               & 0.51M \\
OnDev-LCT-8/1       & None                   & 32-32            & \multicolumn{1}{c|}{\textbf{86.64}}    & \textbf{68.44}        & \textbf{78.28}    & \textbf{82.42}    & \textbf{82.02}           & 3.77               & 0.91M \\ \hline
\end{tabular}
}
\caption{\textbf{Comparative analysis with low-bit vision transformers} on CIFAR-10 and FEMNIST datasets.
Three best accuracy (\%) values are marked in bold.
Inference time represents the wall-clock time, measured on a single NVIDIA RTX 3080 GPU for the CIFAR-10 test set with a batch size of 1000.
}
\label{tab:low-bit_vit}
\end{table*}

\begin{table*}[t]
\centering
\resizebox{0.75\linewidth}{!}{%
\begin{tabular}{|l|C{1.7cm}|cccc|C{2cm}|C{2cm}|}
\hline
\multicolumn{1}{|c|}{\multirow{3}{*}{\textbf{Model}}}    & \multirow{3}{*}{\begin{tabular}[c]{@{}c@{}}\textbf{Pruning}\\ \textbf{Rate}\end{tabular}} & \multicolumn{4}{c|}{\textbf{CIFAR-10}}  & \multirow{3}{*}{\textbf{\#Params}} & \multirow{3}{*}{\begin{tabular}[c]{@{}c@{}}\textbf{Params}\\ \textbf{Ratio}\end{tabular}}  \\ \cline{3-6}
\multicolumn{1}{|c|}{}                                   &                                                                         & \multicolumn{1}{c|}{\multirow{2}{*}{\textbf{Centralized}}} & \multicolumn{3}{c|}{\textbf{FedAvg with 50 clients}} &                                                           &           \\ \cline{4-6}
\multicolumn{1}{|c|}{}                                   &                                                                         & \multicolumn{1}{c|}{}                             & $\bm{\beta=0.1}$  & $\bm{\beta=0.5}$  & $\bm{\beta=5}$    &                           &                                                                           \\ \hline
\begin{tabular}[c]{@{}l@{}}ViT-Tiny (Baseline)\end{tabular}     & 0.0                                                                       & \multicolumn{1}{c|}{55.86}                        & 36.25         & 47.85         & 50.76                                                                            & 5.49M                     & 1.000$\times$                                                                         \\ \hline
\begin{tabular}[c]{@{}l@{}}L1-norm Importance\end{tabular}    & 0.3                                                                    & \multicolumn{1}{c|}{52.91}                        & 33.23         & 44.94         & 46.88                                                                            & 3.01M                     & 0.548$\times$                                                                    \\ 
\begin{tabular}[c]{@{}l@{}}L2-norm Importance\end{tabular}    & 0.3                                                                    & \multicolumn{1}{c|}{53.21}                        & 33.59         & 44.86         & 46.71                                                                            & 3.01M
                     & 0.548$\times$                                                                    \\ 
\begin{tabular}[c]{@{}l@{}}Taylor Importance\end{tabular} & 0.3                                                                    & \multicolumn{1}{c|}{53.90}                        & 32.51         & 44.97         & 47.58                                                                            & 3.01M                     & 0.548$\times$                                                                    \\
Random Importance                                                           & 0.3                                                                    & \multicolumn{1}{c|}{39.88}                        & 27.66         & 37.59         & 39.08                                                                            & 3.01M                     & 0.548$\times$                                                                    \\\hline
OnDev-LCT-1/1                                                     & 0.0                                                                       & \multicolumn{1}{c|}{84.55}             & 66.18                 & 76.93             & 79.71                                                                            & 0.21M                     & 0.038$\times$                                                                         \\
OnDev-LCT-2/1                                                     & 0.0                                                                       & \multicolumn{1}{c|}{\textbf{86.27}}    & \textbf{68.25}        & \textbf{78.20}    & \textbf{80.55}                                                                            & 0.31M                     & 0.056$\times$                                                                         \\
OnDev-LCT-4/1                                                     & 0.0                                                                       & \multicolumn{1}{c|}{\textbf{86.61}}    & \textbf{67.24}        & \textbf{77.68}    & \textbf{80.87}                                                                            & 0.51M                     & 0.093$\times$                                                                         \\
OnDev-LCT-8/1                                                     & 0.0                                                                       & \multicolumn{1}{c|}{\textbf{86.64}}    & \textbf{68.44}        & \textbf{78.28}    & \textbf{82.42}                                                                            & 0.91M                     & 0.166$\times$                                                                         \\ \hline
\end{tabular}
}
\caption{\textbf{Comparative analysis with pruned vision transformers} on CIFAR-10 dataset.
Three best accuracy (\%) values are marked in bold.
}
\label{tab:pruned_vit}
\end{table*}

\begin{table*}[t]
\centering
\resizebox{0.75\linewidth}{!}{%
\begin{tabular}{|L{5cm}|cccc|C{2cm}|C{2cm}|}
\hline
\multicolumn{1}{|c|}{\multirow{3}{*}{\textbf{Model}}} & \multicolumn{4}{c|}{\textbf{CIFAR-10}}                                                                   & \multirow{3}{*}{\begin{tabular}[c]{@{}c@{}}\textbf{FEMNIST}\\ \textbf{($\bm{C=10}$)} \end{tabular}} & \multirow{3}{*}{\textbf{\#Params}} \\ \cline{2-5}
\multicolumn{1}{|c|}{}                       & \multicolumn{1}{c|}{\multirow{2}{*}{\textbf{Centralized}}} & \multicolumn{3}{c|}{\textbf{FedAvg with 50 clients}} &                          &                           \\ \cline{3-5}
\multicolumn{1}{|c|}{}                       & \multicolumn{1}{c|}{}                             & $\bm{\beta=0.1}$      & $\bm{\beta=0.5}$      & $\bm{\beta=5}$      &                          &                           \\ \hline
EdgeNeXt-XXS                                 & \multicolumn{1}{c|}{58.17}                        & 36.22                & 45.24             & 53.16             & 63.65                         & 1.17M                         \\
MobileViT-XXS                               & \multicolumn{1}{c|}{52.91}                        & 28.30                & 45.91             & 49.37             & 75.37                         & 0.94M                       \\
MobileViTv2-0.5                              & \multicolumn{1}{c|}{52.84}                        & 28.28                & 48.95             & 51.80             & 75.64                         & 1.13M                        \\
CCT-2/2                      & \multicolumn{1}{c|}{79.71}            & 55.67         & 68.87             & 73.45         & 80.71                 & 0.28M             \\
CCT-4/2                      & \multicolumn{1}{c|}{80.92}            & 56.05         & 69.82             & 74.52         & 81.28                 & 0.48M             \\
Hybrid Vision Transformer (ViT)              & \multicolumn{1}{c|}{74.64}                        & 45.22                & 65.70             & 70.93             & 80.25                         & 0.75M                         \\
Hybrid Vision Performer (ViP)                & \multicolumn{1}{c|}{77.52}                        & 43.87                & 65.53             & 70.89             & 80.37                         & 0.76M                        \\
Hybrid Vision Linformer (ViL)                & \multicolumn{1}{c|}{50.68}                        & 31.83                & 49.01             & 57.91             & 76.65                         & 0.95M                         \\
Hybrid Vision Nystr\"oformer (ViN)            & \multicolumn{1}{c|}{77.35}                        & 51.25                & 63.81             & 68.56             & 81.31                         & 0.76M                        \\ \hline
OnDev-LCT-1/1                                & \multicolumn{1}{c|}{84.55}                        & 66.18                & 76.93             & 79.71             & 80.68                         & 0.21M \\
OnDev-LCT-2/1                                & \multicolumn{1}{c|}{\textbf{86.27}}               & \textbf{68.25}       & \textbf{78.20}    & \textbf{80.55}    & \textbf{81.43}                & 0.31M \\
OnDev-LCT-4/1                                & \multicolumn{1}{c|}{\textbf{86.61}}               & \textbf{67.24}       & \textbf{77.68}    & \textbf{80.87}    & \textbf{81.84}                & 0.51M \\
OnDev-LCT-8/1                                & \multicolumn{1}{c|}{\textbf{86.64}}               & \textbf{68.44}       & \textbf{78.28}    & \textbf{82.42}    & \textbf{82.02}                & 0.91M                     \\
\hline
\end{tabular}
}
\caption{\textbf{Comparative analysis with convolutional transformers} on CIFAR-10 and FEMNIST datasets.
Each model variant is trained from scratch without applying data augmentation or learning rate schedulers. 
Three best accuracy (\%) values are marked in bold.
}
\label{tab:conv_token_schemes}
\end{table*}

\begin{table*}[t]
\centering
\resizebox{0.75\linewidth}{!}{%
\begin{tabular}{|L{2.5cm}|C{2cm}|cccc|C{2cm}|C{2cm}|}
\hline
\multicolumn{1}{|c|}{\multirow{3}{*}{\textbf{Model}}}             & \multirow{3}{*}{\textbf{Pre-trained}} & \multicolumn{4}{c|}{\textbf{CIFAR-10}}             & \multirow{3}{*}{\begin{tabular}[c]{@{}c@{}}\textbf{FEMNIST}\\ \textbf{($\bm{C=10}$)} \end{tabular}} & \multirow{3}{*}{\textbf{\#Params}} \\ \cline{3-6}
\multicolumn{1}{|c|}{}                                                            &                             & \multicolumn{1}{c|}{\multirow{2}{*}{\textbf{Centralized}}} & \multicolumn{3}{c|}{\textbf{FedAvg with 50 clients}} &                          &                           \\ \cline{4-6}
\multicolumn{1}{|c|}{}                                                            &                             & \multicolumn{1}{c|}{}                             & $\bm{\beta=0.1}$      & $\bm{\beta=0.5}$      & $\bm{\beta=5}$      &                          &                           \\ \hline
MobileNetv2/1.0 & \cmark                        & \multicolumn{1}{c|}{82.22}                        & 49.85         & \textbf{80.74}         & \textbf{84.00}       &    79.53                      &   2.30M                        \\
MobileNetv2/1.0 & \xmark                       & \multicolumn{1}{c|}{36.76}                        & 31.34         & 47.92         & 50.72       &   71.81                       &    2.30M                       \\ \hline
ViT-Tiny                                                                          & \cmark                        & \multicolumn{1}{c|}{67.51}                        & 48.82         & 64.09         & 68.20       &        61.77          &  5.49M                         \\
ViT-Tiny                                                                          & \xmark                       & \multicolumn{1}{c|}{55.86}                        & 36.25         & 47.85         & 50.76       &         57.10           &  5.49M                         \\ \hline
MobileViT-XXS                                                                     & \cmark                        & \multicolumn{1}{c|}{81.27}                        & \textbf{56.75}         & \textbf{79.12}         & \textbf{82.77}       &       \textbf{81.04}             &  0.94M                         \\
MobileViT-XXS                                                                     & \xmark                       & \multicolumn{1}{c|}{52.91}                        & 28.30         & 45.91         & 49.37       &         75.37           &  0.94M                         \\ \hline
MobileViTv2-0.5                                                                   & \cmark                        & \multicolumn{1}{c|}{\textbf{82.69}}                        & 44.06         & \textbf{80.52}         & \textbf{83.13}       &        80.65            &  1.13M                         \\
MobileViTv2-0.5                                                                   & \xmark                       & \multicolumn{1}{c|}{52.84}                        & 28.28         & 48.95         & 51.80       &         75.64           &  1.13M                         \\ \hline
OnDev-LCT-8/1                                                                     & \xmark                       & \multicolumn{1}{c|}{\textbf{86.64}}                        & \textbf{68.44}         & 78.28         & 82.42       & \textbf{82.02}                    & 0.91M                     \\
OnDev-LCT-8/3                                                                     & \xmark                       & \multicolumn{1}{c|}{\textbf{87.65}}                        & \textbf{68.56}         & 78.56         & 82.70       & \textbf{82.75}                    & 0.95M                     \\ \hline
\end{tabular}
}
\caption{\textbf{Comparative analysis with ImageNet pre-trained models} on CIFAR-10 and FEMNIST datasets.
Three best accuracy (\%) values are marked in bold.
}
\label{tab:pretrained}
\end{table*}

\section{Additional Results}\label{sec:additional_results}
We present additional results for the performance analysis of our proposed architecture in this section.

\paragraph{Comparative analysis with low-bit vision transformers}\label{para:low_bit}
As an orthogonal approach to our work, we explore the existing literature on popular quantized, aka low-bit vision transformers, emphasizing their contributions to the domain of on-device image classification tasks.  
\citet{liu2021vtptq} present VT-PTQ, a mixed-precision post-training quantization scheme, which introduces a ranking loss to the quantization objective to keep the relative order of the self-attention results after quantization.
\citet{yuan2021ptq4vit} develop the efficient PTQ4ViT framework, employing twin uniform quantization to handle the special distributions of post-softmax and post-GELU activations.
FQ-ViT~\citep{lin2021fqvit} is then implemented under the PTQ, a training-free paradigm, by introducing the power-of-two factor (PTF) and log-int-softmax (LIS) quantization methods for LayerNorm and Softmax modules, which are not quantized in PTQ4ViT.
In comparison to the adaptation of vision transformers to prior PTQ-based quantization schemes, including MinMax, EMA~\citep{jacob2018ema}, Percentile~\citep{li2019percentile}, OMSE~\citep{choukroun2019omse}, and PTQ4ViT~\citep{yuan2021ptq4vit}, FQ-ViT achieves a nearly lossless quantization performance with full-precision models.
On the other hand, \citet{li2022qvit} analyze the quantization robustness of each component in the vision transformer to implement a fully quantized ViT baseline under the straightforward QAT pipeline.
The Q-ViT is then proposed by introducing an information rectification module (IRM) and a distribution-guided distillation (DGD) scheme towards accurate and fully quantized vision transformers.
Experimental results on the ImageNet~\citep{deng2009imagenet} dataset show that Q-ViTs outperform the baselines and achieve competitive performance with the full-precision counterparts while significantly reducing memory and computational resources.

In Table~\ref{tab:low-bit_vit}, we present a comprehensive comparison of our proposed OnDev-LCTs with various low-bit transformers (i.e., ViT-S backbone) in terms of accuracy, inference time, and number of parameters for image classification tasks on CIFAR-10~\citep{krizhevsky2009cifar} and FEMNIST~\citep{caldas2018leaf} datasets.
To ensure a fair comparison, we implement Q-ViT without the DGD scheme, which relies on the performance of a powerful teacher model.
All models are trained from scratch without pre-training or applying data augmentation techniques and learning rate schedulers.
The results clearly show that our proposed OnDev-LCTs outperform the low-bit vision transformers in all cases.
Nonetheless, we maintain the perspective that the performance of low-bit vision transformers can also be significantly improved by leveraging a powerful backbone model along with a proper configuration of data augmentation and learning rate scheduler while being smaller and faster than traditional vision transformers.
However, it is worth noting that low-bit vision transformers can be more complex to train and may not be as effective for certain tasks.
As a result, the choice of whether to use our OnDev-LCT or a low-bit vision transformer may depend on the specific task and application.

\paragraph{Comparative analysis with vision transformer pruning}\label{para:pruning}
In this section, we delve into prior research on vision transformer pruning, which is an emerging study within the broader domain of model compression and optimization.
This process systematically identifies and removes unnecessary parameters, such as unimportant weights or neurons, from a pre-trained vision transformer, guided by various criteria, including weight magnitude~\citep{lecun1989obd, han2015deep_compression, liu2017learning}, gradient magnitude~\citep{molchanov2019importance, molchanov2016pruning}, activation values~\citep{chen2021chasing}, or importance scores~\citep{zhu2021vtp}.
The primary goal is to create a more compact vision transformer that maintains reasonable performance, making it suitable for deployment on resource-constrained devices where computational efficiency is crucial.
One notable approach, VTP~\citep{zhu2021vtp}, extends the principles of network slimming~\citep{liu2017learning} to reduce the number of embedding dimensions through the introduction of learnable coefficients that evaluate importance scores. Neurons with small coefficients are removed based on a predefined threshold. However, it's worth noting that VTP requires manual tuning of thresholds for all layers and subsequent model fine-tuning.
Another method, WDPruning~\citep{yu2022width}, is a structured pruning technique for vision transformers, which employs a 0/1 mask to differentiate unimportant and important parameters based on their magnitudes. While it uses a differentiable threshold, the non-differentiable mask may introduce gradient bias, potentially resulting in suboptimal weight retention.
X-Pruner~\citep{yu2023x} focuses on removing less contributing units, where each prunable unit is assigned an explainability-aware mask to quantify its contribution to predicting each class in terms of explainability. 
Finally, DepGraph~\citep{fang2023depgraph} serves as a grouping algorithm used to analyze dependencies in networks for enabling any structural pruning over various network architectures.
Nonetheless, while vision transformer pruning can yield smaller and faster models, the challenge remains in finding the right balance between reducing model size and preserving task-specific features.

Table~\ref{tab:pruned_vit} provides a performance comparison between our proposed OnDev-LCTs and pruned vision transformers on the CIFAR-10~\citep{krizhevsky2009cifar} dataset.
We focus on key metrics such as accuracy and the reduction in parameters compared to the ViT-Tiny~\citep{dosovitskiy2020image} backbone.
We followed the Torch-Pruning (TP) implementation for structural pruning, which employs the DepGraph algorithm to physically remove parameters~\citep{fang2023depgraph}.
Several important criteria are considered, including L-p Norm, Taylor, and Random.
The results demonstrate that our proposed OnDev-LCTs consistently outperform the pruned ViT models in all scenarios.
From one point of view, developing effective pruning strategies can be a complex task, often involving iterative trial and error to identify the optimal model components for pruning.
Aggressive pruning carries the risk of significant model generalization loss, as removing too many parameters may lead to underfitting and reduced performance.
In summary, vision transformer pruning is a technique applied to existing large models to reduce their size.
On the other hand, OnDev-LCT provides an efficient and lightweight neural network architecture, offering an alternative solution for resource-constrained on-device vision applications.

\begin{table*}[t]
\centering
\resizebox{0.95\linewidth}{!}{%
\begin{tabular}{|L{3cm}|C{2cm}|C{2cm}||ccc|ccc|C{2cm}|C{2cm}|}
\hline
\multicolumn{1}{|c|}{\multirow{2}{*}{\textbf{Model}}} & \multirow{2}{*}{\textbf{\begin{tabular}[c]{@{}c@{}}Centralized\\ Top-1 Acc\end{tabular}}} & \multirow{2}{*}{\textbf{\begin{tabular}[c]{@{}c@{}}Centralized\\ Top-5 Acc\end{tabular}}} & \multicolumn{3}{c|}{\textbf{FedAvg with 10 clients}} & \multicolumn{3}{c|}{\textbf{FedAvg with 50 clients}} & \multirow{2}{*}{\textbf{\#Params}}    & \multirow{2}{*}{\textbf{MACs}} \\ \cline{4-9}
\multicolumn{1}{|c|}{}                                &                                         &                                                  & $\bm{\beta=0.1}$   & $\bm{\beta=0.5}$   & $\bm{\beta=5}$  &  $\bm{\beta=0.1}$   & $\bm{\beta=0.5}$   & $\bm{\beta=5}$   &                                 &                                  \\ \hline
ResNet-20              & 26.99     & 50.38         & 13.91           & 22.18           & 26.13           & 9.08            & 16.64         & 22.09       & 0.34M       & 0.04G            \\
ResNet-32              & 27.92     & 51.98         & 14.78           & 22.82           & 27.03           & 10.23           & 17.89         & 21.84       & 0.53M       & 0.07G            \\
ResNet-44              & 28.12     & 52.35         & 15.16           & 23.45           & 27.67           & 11.32           & 17.25         & 23.04       & 0.73M       & 0.10G            \\ 
ResNet-56              & 28.18     & 52.57         & 15.03           & 23.56           & 27.93           & 12.45           & 18.42         & 23.06       & 0.93M       & 0.13G            \\
\hline
MobileNetv2/0.5        & 18.44     & 39.22         & 11.16           & 18.87           & 20.28           & 8.90            & 14.09         & 16.19       & 1.99M        & \textless{}0.01G \\ 
MobileNetv2/0.75       & 20.66     & 42.72         & 16.27           & 23.60           & 25.58           & 10.63           & 15.85         & 18.04       & 2.66M        & \textless{}0.01G \\
MobileNetv2/1.0        & 29.84     & 54.60         & 16.58           & 24.78           & 27.76           & 11.00           & 16.92         & 20.14       & 3.54M        & 0.01G \\
\hline
ViT-Lite-1/4           & 11.76     & 27.39         & 4.85            & 8.53            & 10.61           & 3.42            & 7.26          & 10.18       & 1.46M        & 0.04G            \\
ViT-Lite-2/4           & 12.94     & 30.31         & 5.55            & 9.35            & 11.73           & 3.81            & 8.00          & 11.08       & 1.48M        & 0.07G            \\ 
\hline
CCT-2/2                & 26.97     & 50.04         & 14.07           & 22.71           & 27.43           & 10.26           & 19.16         & 24.14       & 0.40M        & 0.03G            \\
CCT-4/2                & 30.27     & 53.89         & 15.82           & 25.21           & 29.99           & 11.20           & 20.92         & 26.46       & 0.60M        & 0.05G            \\ 
\hline
OnDev-LCT-1/1          & 38.46     & 62.95         & 20.91           & 32.01           & 36.93           & 13.87           & 24.68         & 31.07       & 0.34M        & 0.03G    \\    
OnDev-LCT-2/1          & \textbf{40.59}                  & \textbf{65.53}              & \textbf{21.78}           & \textbf{32.93}           & \textbf{38.59}           & \textbf{14.84}           & \textbf{26.23}         & \textbf{32.39}       & 0.44M        & 0.04G            \\
OnDev-LCT-4/1          & \textbf{42.34}                  & \textbf{67.33}              & \textbf{23.27}           & \textbf{34.66}           & \textbf{39.98}           & \textbf{15.54}           & \textbf{27.28}         & \textbf{33.77}       & 0.64M        & 0.05G            \\
OnDev-LCT-8/1          & \textbf{43.37}                  & \textbf{68.30}              & \textbf{24.87}           & \textbf{35.23}           & \textbf{41.03}           & \textbf{16.32}           & \textbf{28.10}         & \textbf{35.84}       & 1.04M        & 0.08G            \\
\hline
\end{tabular}
}%
\caption{\textbf{Performance analysis on ImageNet-32 dataset.}
Three best accuracy (\%) values are marked in bold.}
\label{tab:imagenet}
\end{table*}

\paragraph{Comparative analysis with convolutional transformers}\label{para:conv_trans}
In this analysis, we conduct a comparative study between our OnDev-LCTs and other convolutional transformers that do not rely on heavy data augmentation techniques.
All models are trained from scratch without utilizing data augmentation or learning rate schedulers.
It is important to note that while CCTs~\citep{hassani2021escaping} require high-level data augmentation techniques to achieve their best performance, we include them in our comparison to provide a complete evaluation of existing lightweight convolutional transformers.
Based on the reported values in Table~\ref{tab:conv_token_schemes}, our models outperform the other benchmarks, demonstrating a significant performance gap compared to variants with similar sizes.
This highlights the superior performance and efficiency of our proposed architecture in terms of parameters and computation cost. 
In contrast to other methods that combine convolutions and transformers, the distinguishing factor of our OnDev-LCT lies in its unique and efficient LCT tokenizer design, effectively leveraging depthwise separable convolutions and residual linear bottleneck blocks.
By integrating the LCT tokenizer before the transformer encoder, we introduce image-specific inductive biases to our OnDev-LCT, enabling it to capture essential local details from image representations. 
Furthermore, the LCT encoder's MHSA mechanism aids in learning global representations of images.
This combination of complementary components allows our model to achieve high accuracy while maintaining its lightweight nature, making it well-suited for on-device vision tasks with limited resources.

\paragraph{Comparative analysis with pre-trained models}\label{para:pretrained_models}
In this analysis, we evaluate the performance of our OnDev-LCTs in comparison to variants of existing state-of-the-art models that are pre-trained on the ImageNet~\citep{deng2009imagenet} dataset.
We implement all the pre-trained models using the PyTorch~\citep{paszke2019pytorch} framework, following the official implementation provided by the Py\textbf{T}orch \textbf{Im}age \textbf{M}odels (timm)~\citep{rw2019timm} collection. 
According to the results shown in Tabel~\ref{tab:pretrained}, our OnDev-LCTs achieve competitive performance even with limited training data, illustrating their robustness and effectiveness for on-device vision tasks.
While pre-trained models have shown promising performance across various vision tasks, their practicality in resource-constrained on-device scenarios is often hindered by challenges related to accessing and utilizing large pre-training datasets.
Such datasets may not be readily available in many domains~\citep{varoquaux2022machine, parekh2022review, mutis2020challenges, ghassemi2020review} due to privacy concerns or data distribution heterogeneity across devices.
Since pre-trained models are designed and optimized for specific datasets and objectives, which may not closely align with the target domain, fine-tuning a pre-trained model on downstream tasks may not always be straightforward for on-device scenarios with limited training data and resources.
Despite not relying on extensive pre-training datasets, our OnDev-LCTs leverage depthwise separable convolutions and MHSA mechanism to efficiently capture both local and global features from images, making them well-suited for on-device vision tasks with limited data.
Thus, the unique design of our OnDev-LCT offers a practical and effective alternative to pre-train-and-fine-tune schemes, providing more robust and efficient solutions for on-device vision tasks in FL scenarios, where pre-training datasets may not be readily available.

\begin{table}[t]
\centering
\resizebox{\columnwidth}{!}{%
\begin{tabular}{|l|C{1.5cm}|c|C{1.5cm}|C{1.5cm}|} 
\hline
\textbf{Model}          & \textbf{Augment}   & \begin{tabular}[c]{@{}c@{}}\textbf{Augment +} \\\textbf{LR Scheduler}\end{tabular} & \textbf{Baseline}    & \textbf{\#Params}  \\ 
\hline
OnDev-LCT-1/1     & 90.04 & 91.05       & \textbf{84.55}      & 0.21M      \\
OnDev-LCT-8/1     & 90.36 & 91.14       & \textbf{86.64}      & 0.91M      \\ 
\hline
OnDev-LCT-1/3     & 90.25 & 91.41       & \textbf{85.73}      & 0.25M      \\
OnDev-LCT-8/3     & 90.50 & 91.99       & \textbf{87.65}      & 0.95M      \\
\hline
\end{tabular}
}%
\caption{\textbf{Performance analysis} of our OnDev-LCT architecture on CIFAR-10 dataset for the impact of simple data augmentation and learning rate scheduler.}
\label{tab:aug_scheduler}
\end{table}

\paragraph{Analysis on ImageNet-32 dataset}\label{para:imagenet}
Table~\ref{tab:imagenet} compares the centralized and FL performance of our proposed models for image classification on ImageNet-32, the down-sampled version of the original ImageNet-1K~\citep{deng2009imagenet} dataset, which contains 1,281,167 training samples and 50,000 test samples of $32\times32$ RGB images in 1,000 classes.
We apply the same hyperparameter settings and training strategy used with the CIFAR-100 dataset and train all models from scratch without data augmentation and learning rate schedulers.
According to the reported values, our models outperform the other baselines, and there is a significant gap between their performance and that of variants with comparable sizes.
Thus, we can summarize that our proposed architecture is better in performance and also efficient in terms of parameters and computation cost.

\begin{table*}[t]
\centering
\resizebox{\linewidth}{!}{%
\begin{tabular}{|l|ccccccccc||ccccccccc|}
\hline
\multicolumn{1}{|c|}{\multirow{3}{*}{\textbf{Model}}} & \multicolumn{9}{c||}{\textbf{FedAvg with 10 clients}}                                                                                                                                                                                                  & \multicolumn{9}{c|}{\textbf{FedAvg with 50 clients}}                                                                                                                                                                                                  \\ \cline{2-19} 
\multicolumn{1}{|c|}{}                                & \multicolumn{3}{c|}{\textbf{MNIST}}                                 & \multicolumn{3}{c|}{\textbf{\begin{tabular}[c]{@{}c@{}}Fashion\\ MNIST\end{tabular}}} & \multicolumn{3}{c||}{\textbf{\begin{tabular}[c]{@{}c@{}}EMNIST\\ Balanced\end{tabular}}} & \multicolumn{3}{c|}{\textbf{MNIST}}                                 & \multicolumn{3}{c|}{\textbf{\begin{tabular}[c]{@{}c@{}}Fashion\\ MNIST\end{tabular}}} & \multicolumn{3}{c|}{\textbf{\begin{tabular}[c]{@{}c@{}}EMNIST\\ Balanced\end{tabular}}} \\ \cline{2-19} 
\multicolumn{1}{|c|}{}                                & $\bm{\beta=0.1}$   & $\bm{\beta=0.5}$ & \multicolumn{1}{c|}{$\bm{\beta=5}$} & $\bm{\beta=0.1}$   & $\bm{\beta=0.5}$       & \multicolumn{1}{c|}{$\bm{\beta=5}$}       & $\bm{\beta=0.1}$   & $\bm{\beta=0.5}$   & $\bm{\beta=5}$              & $\bm{\beta=0.1}$   & $\bm{\beta=0.5}$ & \multicolumn{1}{c|}{$\bm{\beta=5}$} & $\bm{\beta=0.1}$   & $\bm{\beta=0.5}$       & \multicolumn{1}{c|}{$\bm{\beta=5}$}       & $\bm{\beta=0.1}$   & $\bm{\beta=0.5}$               & $\bm{\beta=5}$              \\ \hline
ResNet-20            & 98.99                & \textbf{99.50}        & \multicolumn{1}{c|}{\textbf{99.58}}        & 81.33                & 91.83                & \multicolumn{1}{c|}{92.75}                     & 81.47                & 88.46                   & 89.68     & 97.65                & \textbf{99.46}        & \multicolumn{1}{c|}{\textbf{99.54}}        & 70.63                & 90.71                & \multicolumn{1}{c|}{\textbf{92.06}}            & 82.98                & 88.64                   & 89.50           \\
ResNet-32            & \textbf{99.13}       & 99.49                 & \multicolumn{1}{c|}{\textbf{99.59}}        & 82.23                & 92.04                & \multicolumn{1}{c|}{92.77}                     & 83.18                & \textbf{89.23}          & 89.64     & 97.38                & \textbf{99.42}        & \multicolumn{1}{c|}{\textbf{99.51}}        & 72.24                & 90.75                & \multicolumn{1}{c|}{\textbf{92.12}}            & 83.70                & \textbf{88.78}          & 89.51           \\
ResNet-44            & 98.99                & \textbf{99.51}        & \multicolumn{1}{c|}{99.57}                 & 82.27                & \textbf{92.15}       & \multicolumn{1}{c|}{92.73}                     & 83.86                & 88.99                   & 89.88     & 97.85                & \textbf{99.44}        & \multicolumn{1}{c|}{\textbf{99.52}}        & 71.51                & \textbf{91.01}       & \multicolumn{1}{c|}{92.04}                     & \textbf{83.90}       & \textbf{88.74}          & \textbf{89.61}  \\ \hline
MobileNetv2/0.2      & 98.31                & 99.02                 & \multicolumn{1}{c|}{99.26}                 & 72.01                & 89.49                & \multicolumn{1}{c|}{89.87}                     & 67.78                & 84.97                   & 83.94     & 94.16                & 98.64                 & \multicolumn{1}{c|}{98.77}                 & 46.82                & 86.62                & \multicolumn{1}{c|}{87.59}                     & 69.66                & 80.64                   & 80.69           \\
MobileNetv2/0.5      & 98.71                & 99.24                 & \multicolumn{1}{c|}{99.32}                 & 75.03                & 90.15                & \multicolumn{1}{c|}{90.62}                     & 73.37                & 85.70                   & 86.90     & 96.85                & 98.97                 & \multicolumn{1}{c|}{98.93}                 & 50.20                & 87.41                & \multicolumn{1}{c|}{89.58}                     & 75.46                & 83.87                   & 80.07           \\ \hline
ViT-Lite-1/8         & 93.58                & 97.95                 & \multicolumn{1}{c|}{98.10}                 & 73.90                & 87.82                & \multicolumn{1}{c|}{88.79}                     & 57.96                & 81.02                   & 84.00     & 83.33                & 96.89                 & \multicolumn{1}{c|}{97.23}                 & 57.54                & 82.33                & \multicolumn{1}{c|}{85.17}                     & 58.55                & 77.72                   & 80.24           \\
ViT-Lite-2/8         & 93.71                & 97.99                 & \multicolumn{1}{c|}{98.11}                 & 75.74                & 87.89                & \multicolumn{1}{c|}{88.82}                     & 58.92                & 81.10                   & 84.10     & 84.19                & 97.01                 & \multicolumn{1}{c|}{97.31}                 & 58.97                & 82.17                & \multicolumn{1}{c|}{85.42}                     & 58.86                & 77.85                   & 80.26           \\ \hline
CCT-2/2              & 98.45                & 99.29                 & \multicolumn{1}{c|}{99.41}                 & 81.90                & 90.42                & \multicolumn{1}{c|}{91.33}                     & 82.74                & 87.94                   & 89.39     & 95.56                & 99.03                 & \multicolumn{1}{c|}{99.18}                 & 72.44                & 88.07                & \multicolumn{1}{c|}{89.75}                     & 81.77                & 86.87                   & 88.45           \\
CCT-4/2              & 98.47                & 99.34                 & \multicolumn{1}{c|}{99.45}                 & 82.80                & 90.74                & \multicolumn{1}{c|}{91.53}                     & 83.64                & 88.32                   & 89.53     & 96.49                & 99.07                 & \multicolumn{1}{c|}{99.34}                 & 73.82                & 87.30                & \multicolumn{1}{c|}{89.98}                     & 82.77                & 87.47                   & 88.69           \\ \hline
OnDev-LCT-1/1        & 98.97                & 99.43                 & \multicolumn{1}{c|}{99.53}                 & 82.75                & 92.03                & \multicolumn{1}{c|}{92.85}                     & 82.32                & 88.96                   & 89.95     & \textbf{97.92}       & 99.29                 & \multicolumn{1}{c|}{99.38}                 & 75.30                & \textbf{90.92}       & \multicolumn{1}{c|}{91.88}                     & 82.13                & 87.92                   & 89.34           \\
OnDev-LCT-2/1        & \textbf{99.18}       & 99.46                 & \multicolumn{1}{c|}{99.52}                 & \textbf{83.12}       & 92.07                & \multicolumn{1}{c|}{\textbf{92.97}}            & \textbf{84.43}       & 89.06                   & \textbf{90.06}     & 97.85                & 99.37                 & \multicolumn{1}{c|}{99.37}                 & \textbf{78.56}       & 90.78                & \multicolumn{1}{c|}{\textbf{92.21}}            & 82.98                & 88.04                   & 89.58           \\
OnDev-LCT-4/1        & 99.07                & \textbf{99.50}        & \multicolumn{1}{c|}{99.54}                 & \textbf{83.71}       & \textbf{92.26}       & \multicolumn{1}{c|}{\textbf{93.13}}            & \textbf{85.96}       & \textbf{89.25}          & \textbf{90.32}     & \textbf{98.21}       & 99.38                 & \multicolumn{1}{c|}{99.36}                 & \textbf{78.24}       & \textbf{90.92}       & \multicolumn{1}{c|}{92.03}                     & \textbf{84.15}       & 88.32                   & \textbf{89.64}  \\
OnDev-LCT-8/1        & \textbf{99.10}       & 99.43                 & \multicolumn{1}{c|}{\textbf{99.59}}        & \textbf{83.42}       & \textbf{92.37}       & \multicolumn{1}{c|}{\textbf{93.15}}            & \textbf{86.99}       & \textbf{89.48}          & \textbf{90.35}     & \textbf{97.87}       & 99.28                 & \multicolumn{1}{c|}{99.43}                 & \textbf{78.82}       & 90.84                & \multicolumn{1}{c|}{91.94}                     & \textbf{84.56}       & \textbf{88.84}          & \textbf{89.88}  \\ \hline
\end{tabular}
}
\caption{\textbf{Comparison on image classification performance in FL} under different degrees of data heterogeneity.
All reported accuracy (\%) values are the best of 3 runs.
All experiments were conducted without applying any data augmentation or learning rate schedulers.
Three best accuracy values are marked in bold.}
\label{tab:fed3}
\end{table*}

\paragraph{Analysis on the impact of data augmentation and learning rate scheduler}\label{para:data_aug_lr}
In this experiment, we analyze the improved performance of our proposed OnDev-LCTs on the CIFAR-10~\citep{krizhevsky2009cifar} dataset by applying simple data augmentation and learning rate schedulers.
We conduct experiments in the centralized scenario by selecting the smallest and largest model variants of our OnDev-LCT with different numbers of standard convolutions in the LCT tokenizer.
First, we apply simple data augmentation techniques, including horizontal flip, random rotation, and zoom.
As shown in Table~\ref{tab:aug_scheduler}, our models significantly improve accuracy even with simple augmentations applied to the training data.
Then, we set a linear learning rate warm-up for the initial 10 epochs, followed by a gradual reduction of the learning rate per epoch using cosine annealing~\citep{loshchilov2016sgdr}.
The performance is further enhanced, demonstrating that a proper configuration of data augmentation and learning rate scheduler can effectively improve the efficiency of our OnDev-LCT when implementing it in real-world applications.

\begin{figure}[t]
	\centering
	\includegraphics[width=0.86\columnwidth]{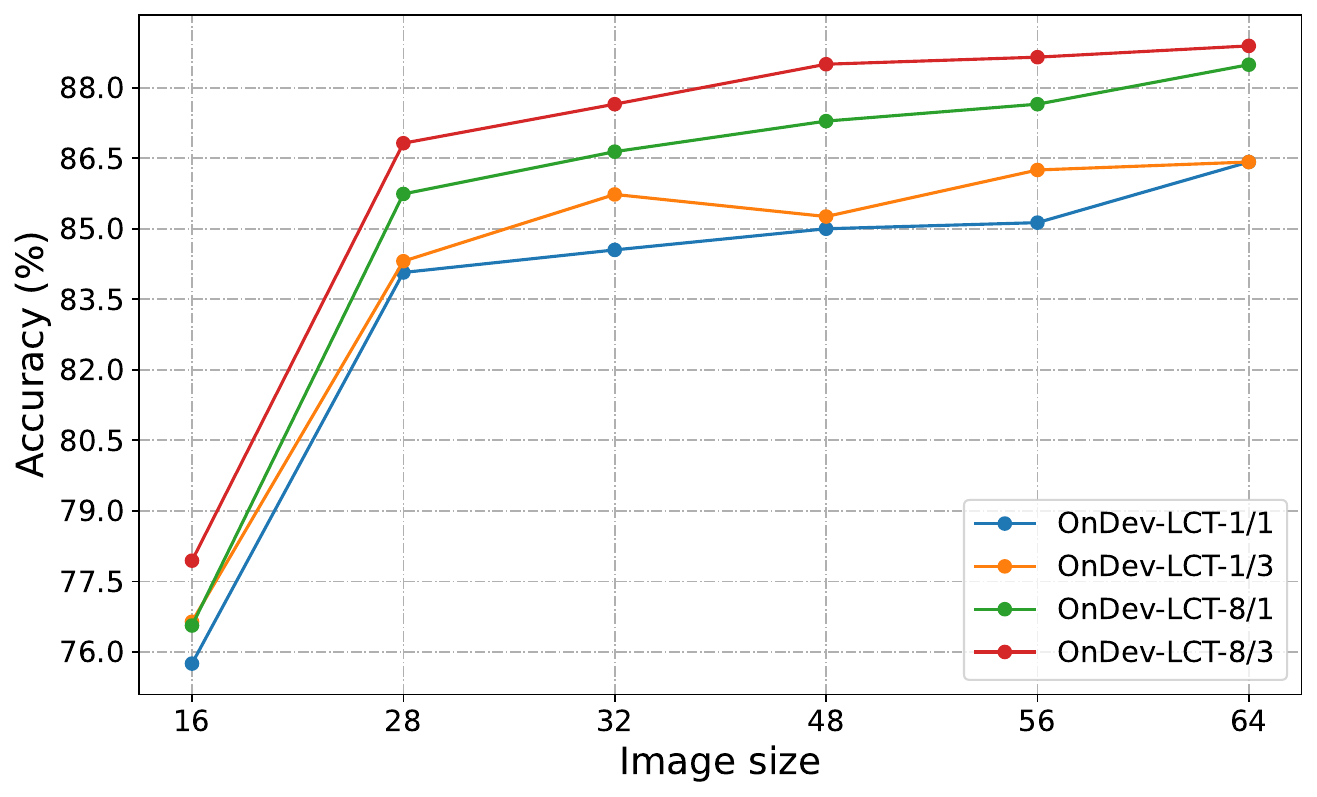}
	\caption{\textbf{Accuracy (\%) versus image size} on CIFAR-10.}
	\label{fig:img_size}
\end{figure}

\paragraph{Analysis on the impact of input image resolution}\label{para:img_resol}
Here, we analyze the performance of our OnDev-LCTs on various image sizes using the CIFAR-10~\citep{krizhevsky2009cifar} dataset.
We conduct this analysis by downsampling the images to two smaller sizes, i.e., $16\times16$, $28\times28$ or upsampling the images to three larger sizes, i.e., $48\times48$, $56\times56$, and $64\times64$.
From Figure~\ref{fig:img_size}, we can observe that our models still perform well with smaller images, even without data augmentation.
Thus, we can infer that by incorporating inductive biases via the LCT tokenizer, our OnDev-LCTs can maintain the spatial information of images, which helps in learning better image representations. 

\begin{figure}[t]
	\centering
	\includegraphics[width=0.86\columnwidth]{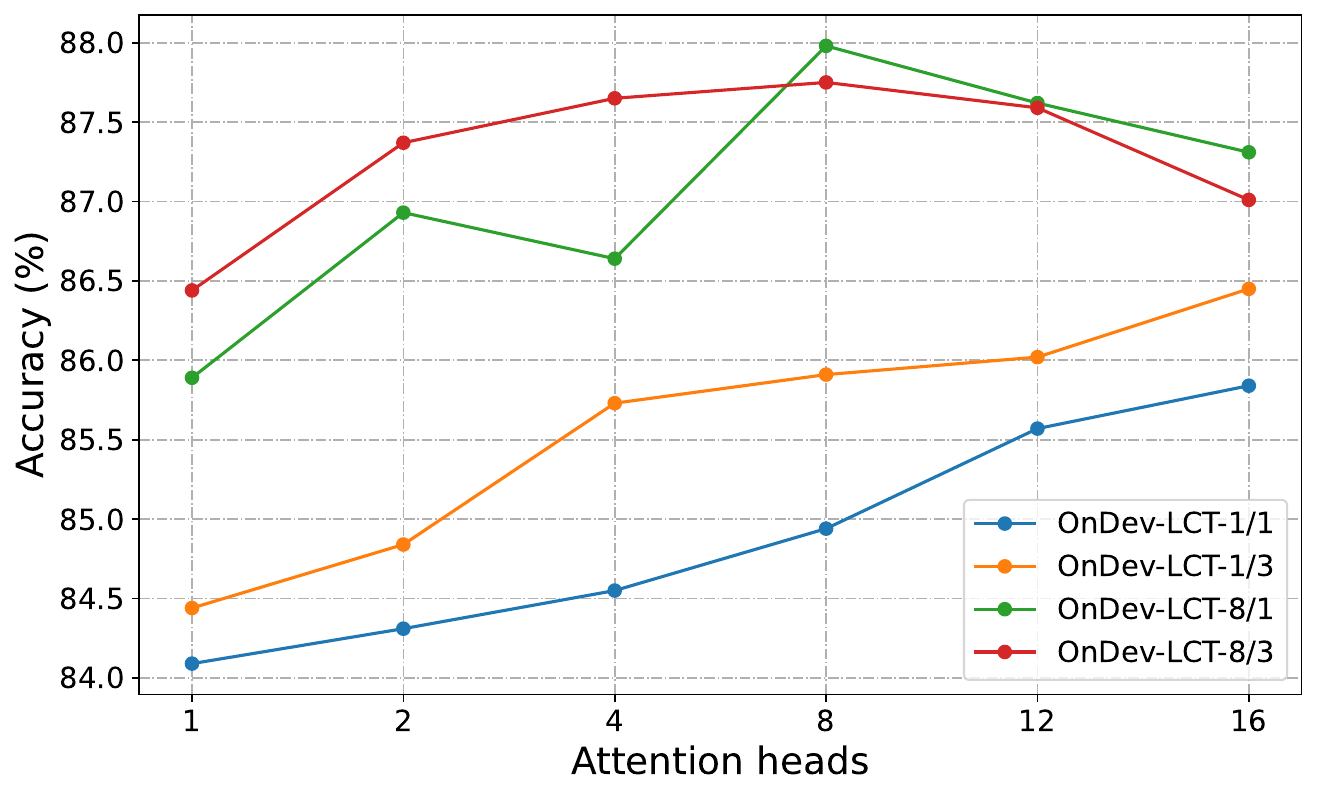}
	\caption{\textbf{Accuracy (\%) versus number of attention heads} on CIFAR-10.}
	\label{fig:heads}
\end{figure}

\paragraph{Analysis on the impact of multi-head self-attention}\label{para:mhsa}
In this experiment, we investigate the impact of varying the number of attention heads in the LCT encoder on the performance of our OnDev-LCTs.
Multiple attention heads in the MHSA block of an LCT encoder help in parallel processing, giving our OnDev-LCT greater power to encode multiple relationships between image patches.
Figure~\ref{fig:heads} demonstrates the performance of our OnDev-LCTs with different numbers of attention heads.
Obviously, small variants of our models perform better with more attention heads, but larger variants gradually drop in accuracy when using 12 and 16 attention heads.
Still, all our models preserve their performance even with a single attention head.

\begin{figure*}[t]
     \centering
     \begin{subfigure}[b]{0.33\textwidth}
         \centering
         \includegraphics[width=\textwidth]{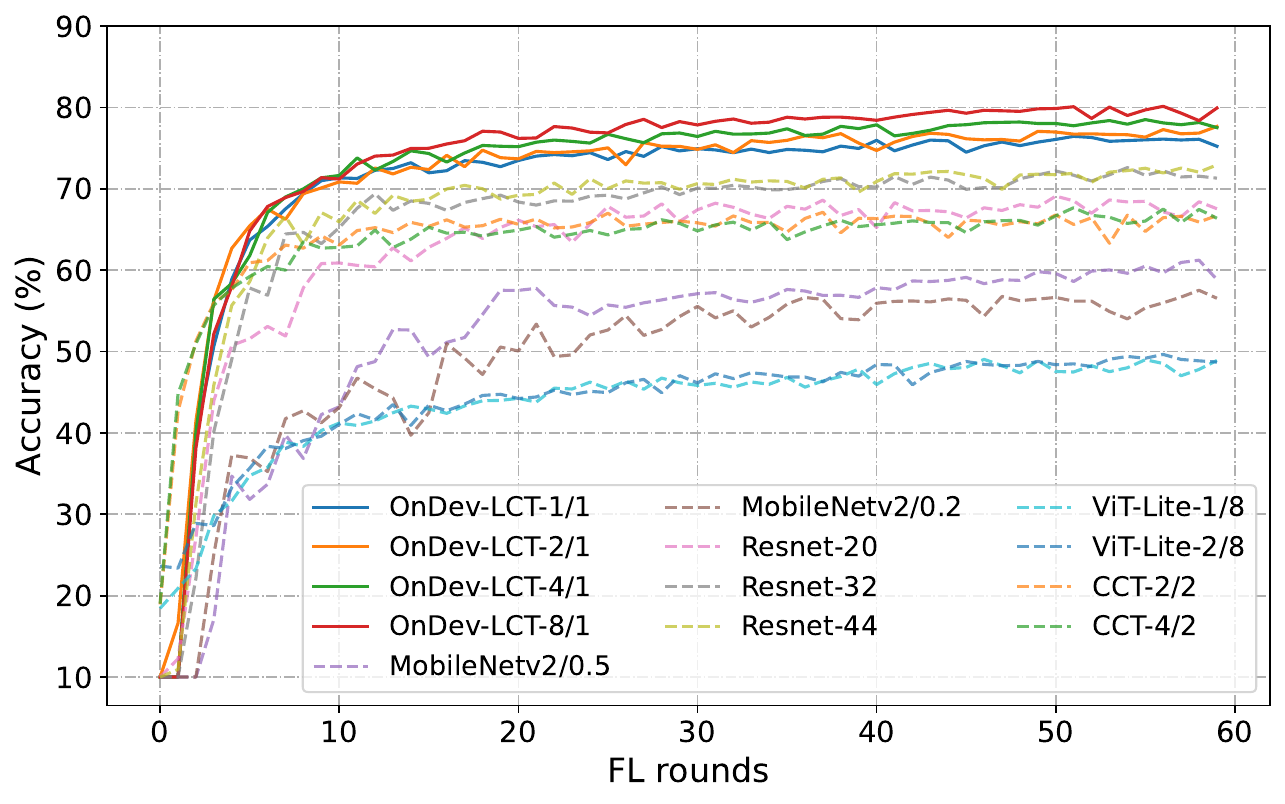}
         \caption{For 10 clients ($\beta=0.1$)}
         \label{fig:10clients_beta01}
     \end{subfigure}
     \hfill
     \begin{subfigure}[b]{0.33\textwidth}
         \centering
         \includegraphics[width=\textwidth]{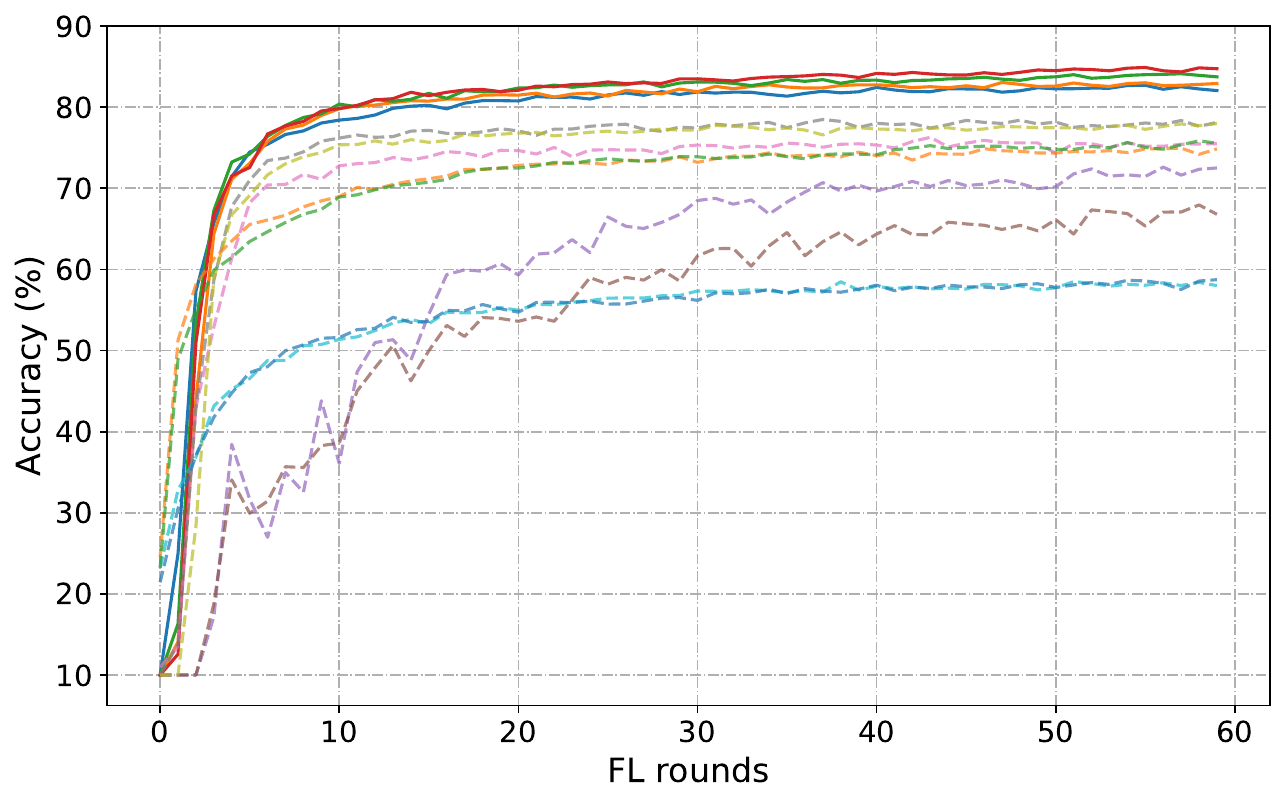}
         \caption{For 10 clients ($\beta=0.5$)}
         \label{fig:10clients_beta05}
     \end{subfigure}
     \hfill
     \begin{subfigure}[b]{0.33\textwidth}
         \centering
         \includegraphics[width=\textwidth]{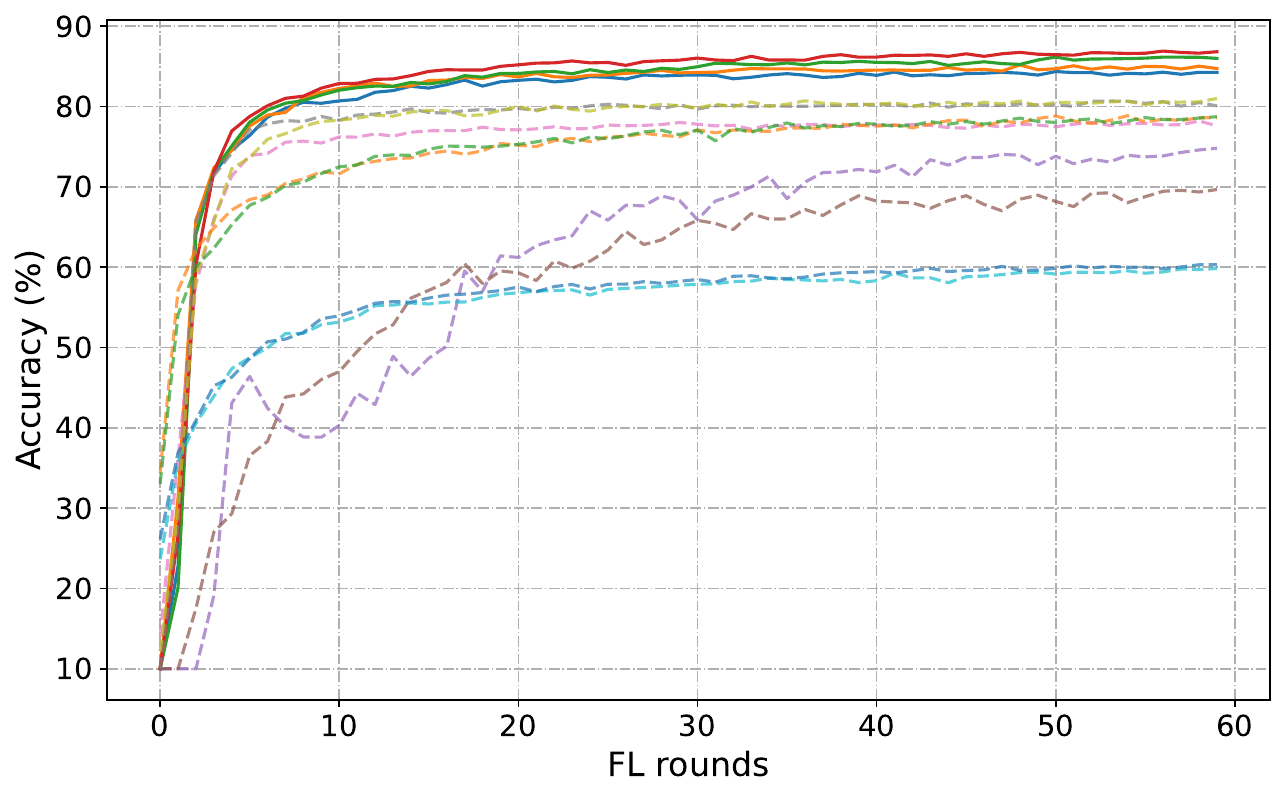}
         \caption{For 10 clients ($\beta=5$)}
         \label{fig:10clients_beta5}
     \end{subfigure}
     \begin{subfigure}[b]{0.33\textwidth}
         \centering
         \includegraphics[width=\textwidth]{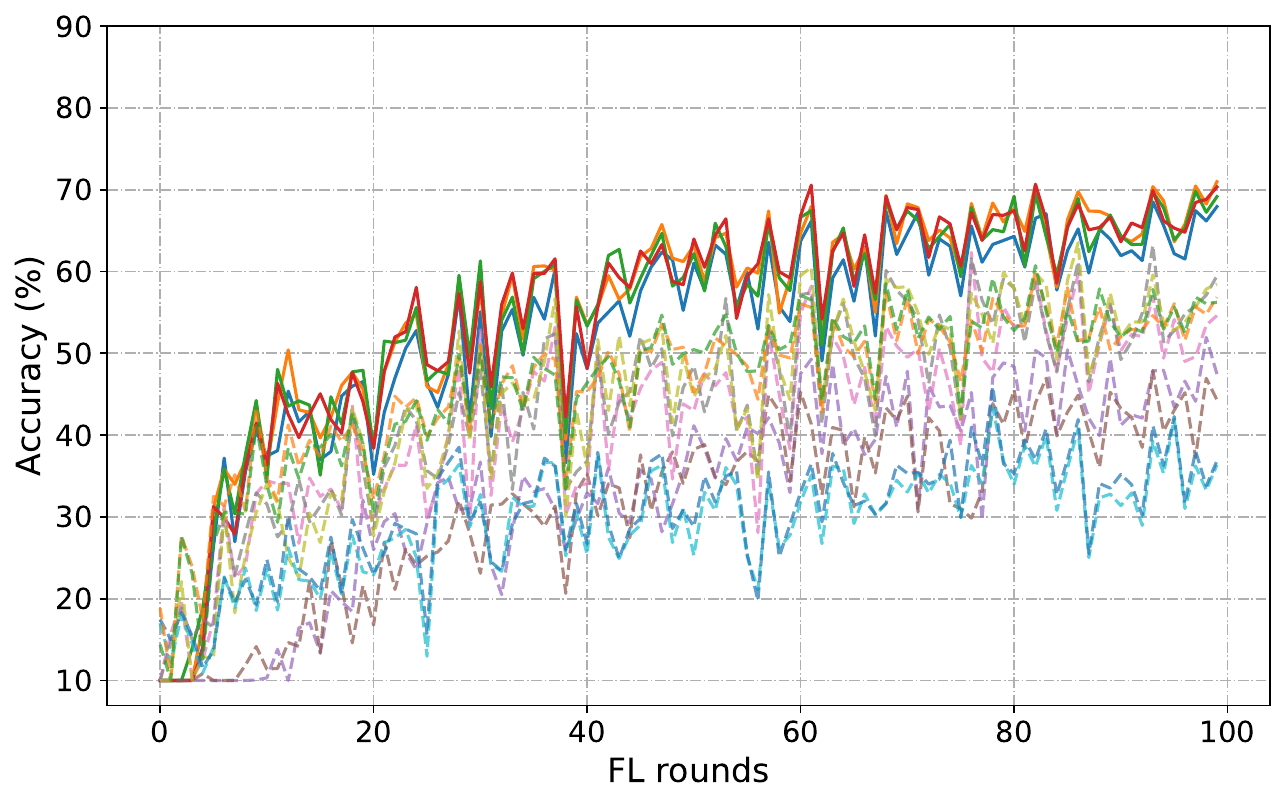}
         \caption{For 50 clients ($\beta=0.1$)}
         \label{fig:50clients_beta01}
     \end{subfigure}
     \hfill
     \begin{subfigure}[b]{0.33\textwidth}
         \centering
         \includegraphics[width=\textwidth]{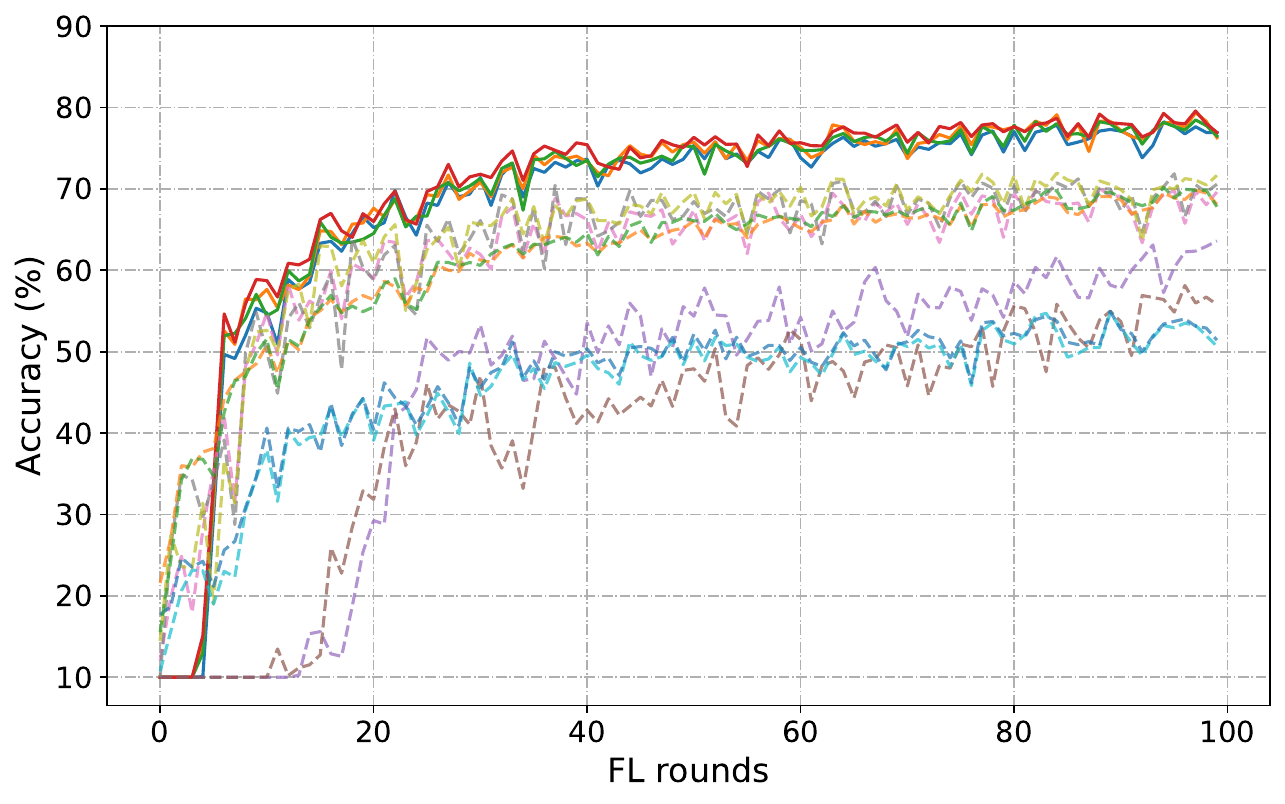}
         \caption{For 50 clients ($\beta=0.5$)}
         \label{fig:50clients_beta05}
     \end{subfigure}
     \hfill
     \begin{subfigure}[b]{0.33\textwidth}
         \centering
         \includegraphics[width=\textwidth]{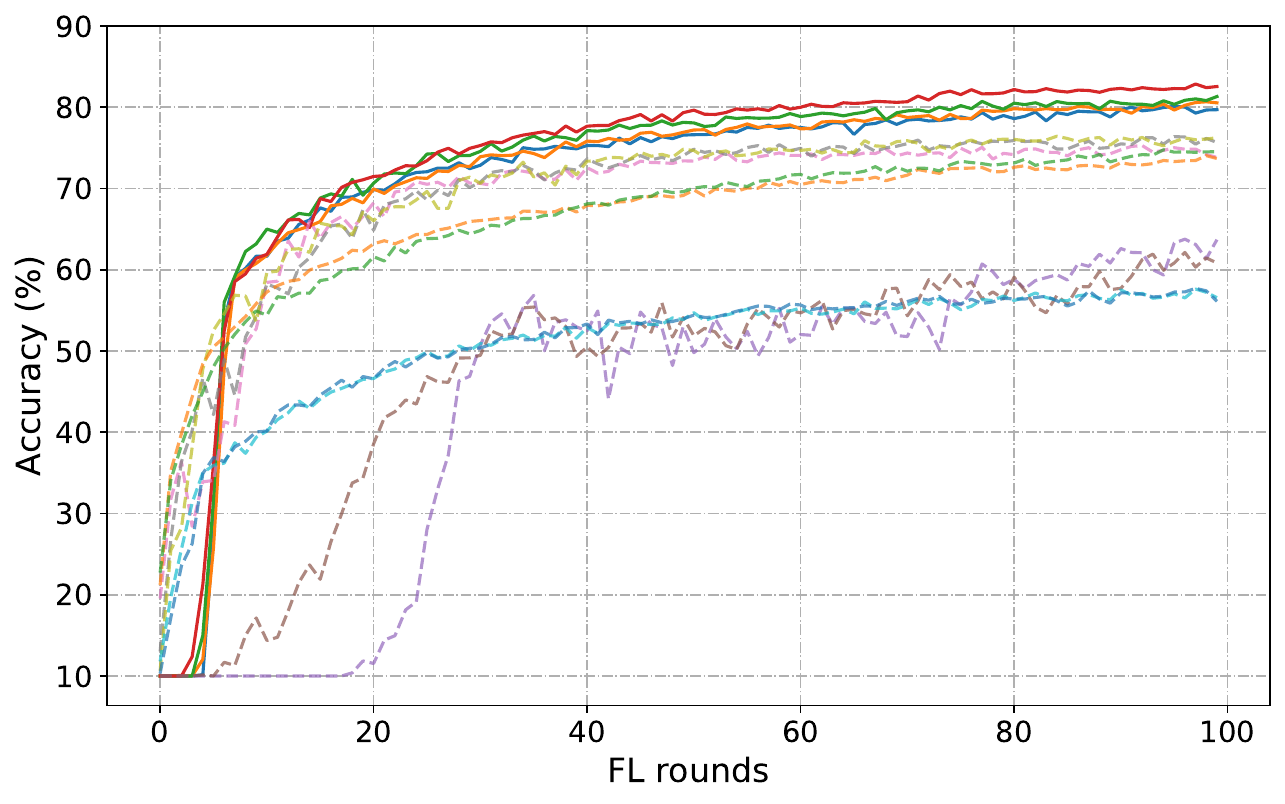}
         \caption{For 50 clients ($\beta=5$)}
         \label{fig:50clients_beta5}
     \end{subfigure}
        \caption{\textbf{Accuracy (\%) versus FL rounds} on CIFAR-10 test set under different degrees of data heterogeneity. Best viewed in color.}
        \label{fig:acc_vs_rnd}
\end{figure*}

\begin{figure*}[t]
     \centering
     \begin{subfigure}[b]{0.33\textwidth}
         \centering
         \includegraphics[width=\textwidth]{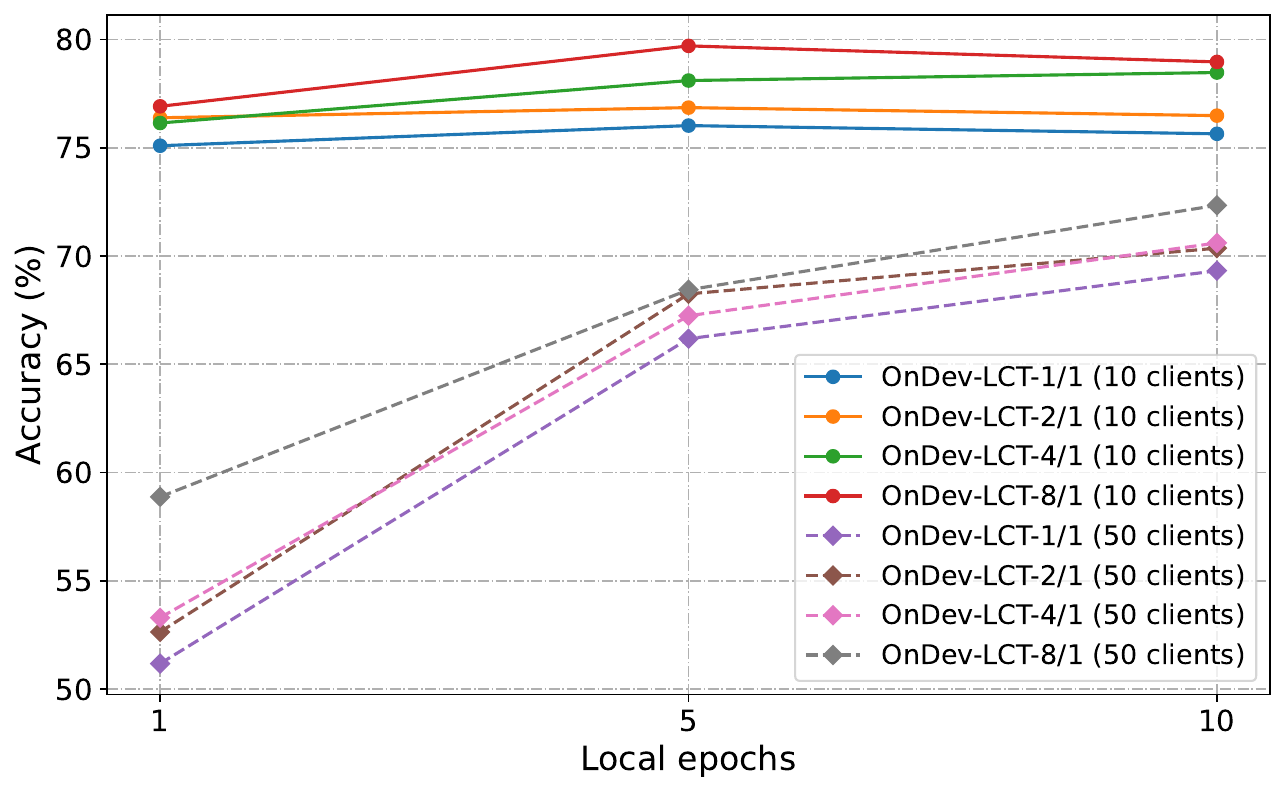}
         \caption{$\beta=0.1$}
         \label{fig:beta01}
     \end{subfigure}
     \hfill
     \begin{subfigure}[b]{0.33\textwidth}
         \centering
         \includegraphics[width=\textwidth]{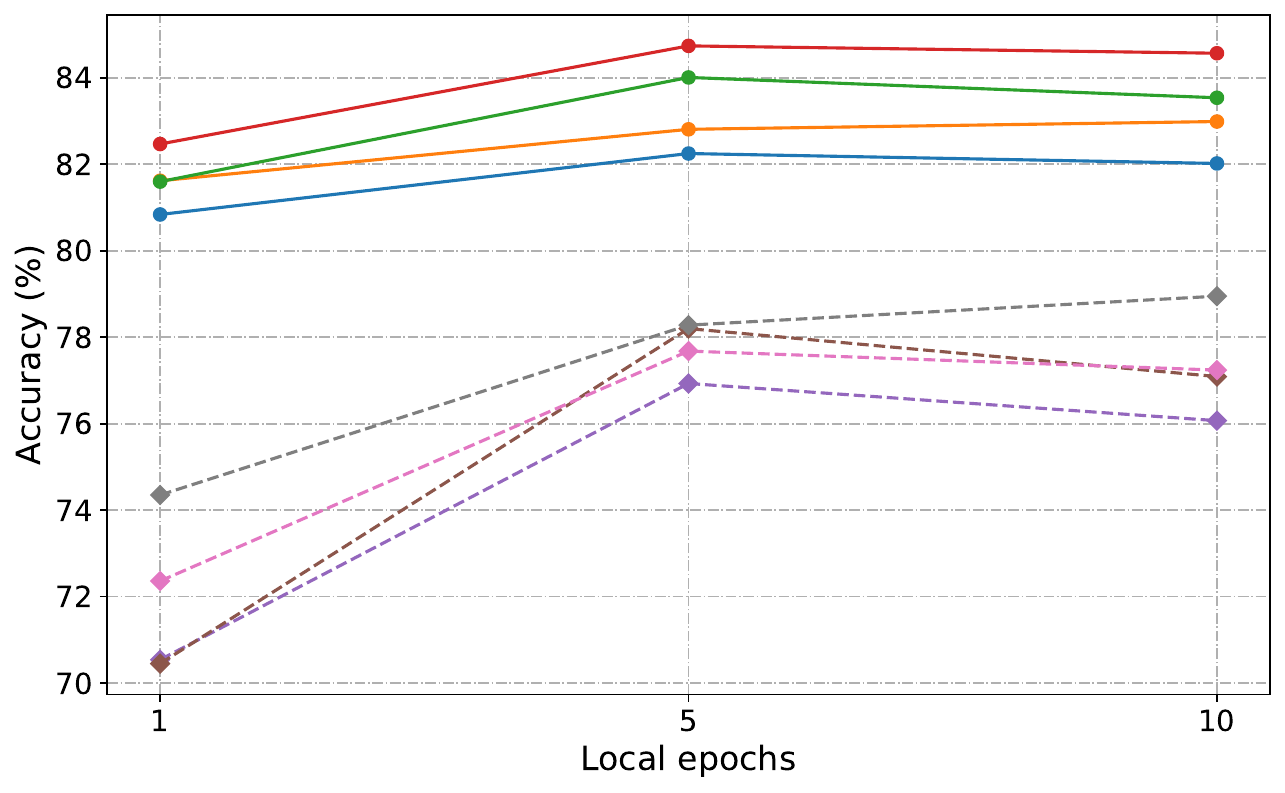}
         \caption{$\beta=0.5$}
         \label{fig:beta05}
     \end{subfigure}
     \hfill
     \begin{subfigure}[b]{0.33\textwidth}
         \centering
         \includegraphics[width=\textwidth]{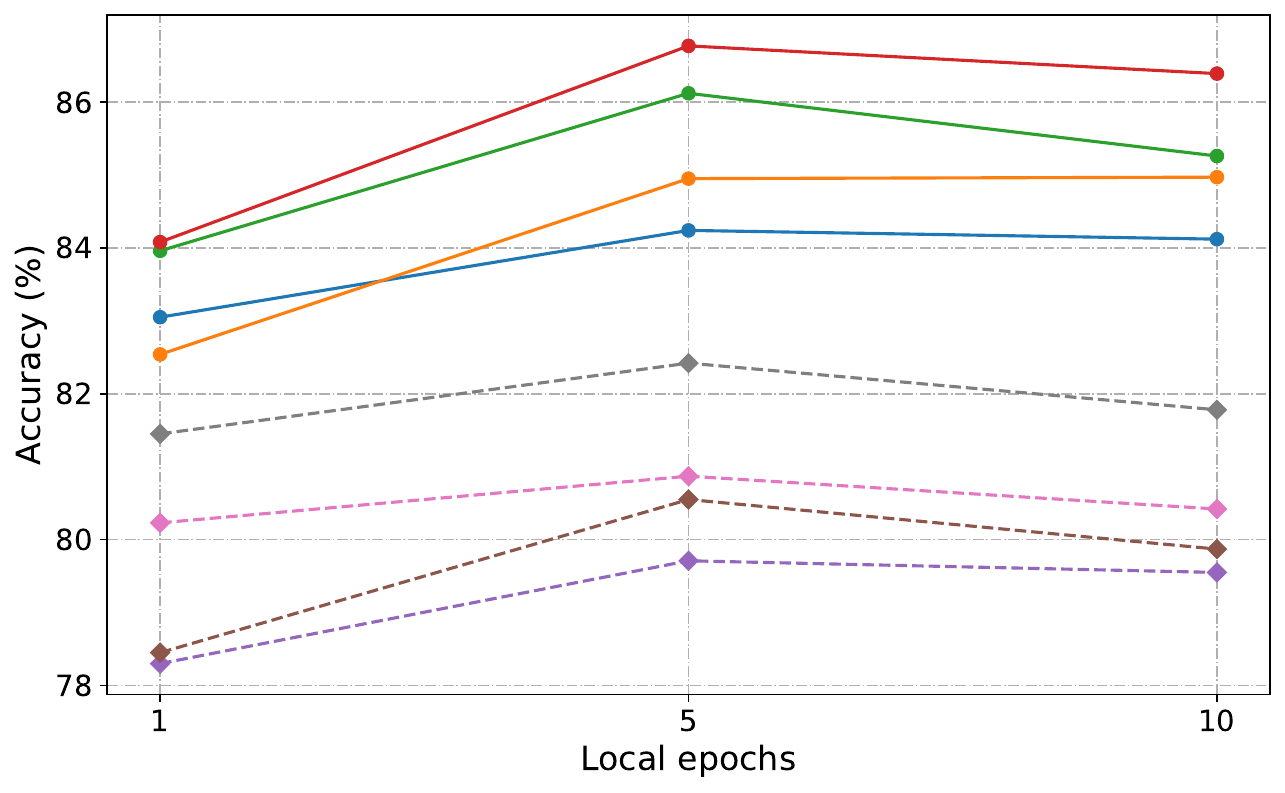}
         \caption{$\beta=5$}
         \label{fig:beta5}
     \end{subfigure}
        \caption{\textbf{Accuracy (\%) versus the number of local epochs} on CIFAR-10 test set under different degrees of data heterogeneity. Best viewed in color.}
        \label{fig:acc_vs_eps}
\end{figure*}

\paragraph{Analysis on various FL settings}\label{para:various_fl}
Image classification performance of our proposed OnDev-LCTs on the MNIST, Fashion-MNIST, and EMNIST-Balanced datasets, compared to the other baselines under various data heterogeneity settings, is shown in Table~\ref{tab:fed3}.
The reported values are the best median values from the last 5 FL rounds of 3 separate runs.
Our findings indicate that OnDev-LCTs outperform the competition in most situations.
Figure~\ref{fig:acc_vs_rnd} shows the accuracy versus the number of FL rounds on the CIFAR-10 dataset with varying degrees of data heterogeneity for both 10-client and 50-client scenarios; thus, our proposed models outperform the other baselines in all cases.
Especially for higher $\beta$ values, OnDev-LCT variants converge faster than the other baselines with a significant performance gap in between.
Figure~\ref{fig:acc_vs_eps} depicts the impact of local epochs in each FL round for the CIFAR-10 dataset.
When we increase the local epoch from 1 to 5, the performance of our models dramatically improves, but just a minor gain when increasing to 10.
As a result, we set the default number of local epochs to 5.

\section{Conclusion}\label{sec:conclusion}
In this study, we propose a novel design, namely OnDev-LCT, which introduces inductive biases of CNNs to the vision transformer by incorporating an early convolutional stem before the transformer encoder.
We leverage efficient depthwise separable convolutions to build residual linear bottleneck blocks of the LCT tokenizer for extracting local features from images. 
The LCT encoder enables our models to learn global representations of images via multi-head self-attention.
Combining the strengths of convolutions and attention mechanisms, we design our OnDev-LCTs for on-device vision tasks with limited training data and resources.
We conduct extensive experiments to analyze the efficiency of our proposed OnDev-LCTs on various image classification tasks compared to the popular lightweight vision models.
Centralized experiments on five benchmark image datasets show the superiority of our OnDev-LCTs over the other baselines.
Furthermore, extensive FL experiments under various settings indicate the efficacy of our OnDev-LCTs in dealing with data heterogeneity and communication bottlenecks since our models significantly outperform the other baselines while having fewer parameters and lower computational demands.
We advocate that the proper configuration of data augmentations and learning rate schedulers could further improve the performance of our OnDev-LCT when implementing it in real-world low-data scenarios.



\bibliographystyle{elsarticle-harv} 
\bibliography{references}







\end{document}